\documentclass[12pt,draftcls,onecolumn]{IEEEtran}
\usepackage{amssymb,amsmath,amsopn,amsfonts,graphicx,color,amsthm}
\usepackage{mathrsfs}
\usepackage{algorithmic}
\usepackage{graphicx}
\usepackage{mathtools}
\usepackage{subfigure}
\usepackage{float}
\usepackage{multirow}
\usepackage{textcomp}
\usepackage{graphicx}
\usepackage{epstopdf}
 \usepackage{caption}

\newtheorem{remark}{Remark}[section]
\newtheorem{theorem}{Theorem}[section]
\newtheorem{lemma}{Lemma}[section]
\newtheorem{proposition}{Proposition}[section]
\newtheorem{condition}{Condition}[section]
\newtheorem{assumption}{Assumption}[section]
\newtheorem{definition}{Definition}[section]
\newtheorem{corollary}{Corollary}[section]

\renewcommand\thesection{\Roman{section}}
\renewcommand\thesubsection{\thesection.\arabic{subsection}}

\makeatletter

\renewcommand{\thelemma}{\arabic{section}.\arabic{lemma}}
\renewcommand{\theproposition}{\arabic{section}.\arabic{proposition}}

\renewcommand{\thedefinition}{\arabic{section}.\arabic{definition}}

\renewcommand{\theremark}{\arabic{section}.\arabic{remark}}
\allowdisplaybreaks

\def\({\Big(}
\def\){\Big)}

\def\X{\mathscr X}
\def\E{\mathbb{E}}
\def\P{\mathbb{P}}
\def\F{\mathcal{F}}

\def\b{\beta}

\def\ba{\begin{array}}
\def\ea{\end{array}}
\def\ban{\begin{eqnarray*}}
\def\ean{\end{eqnarray*}}
\def\bann{\begin{eqnarray*}}
\def\eann{\end{eqnarray*}}
\def\bd{\begin{description}}
\def\ed{\end{description}}
\def\be{\begin{equation}}
\def\ee{\end{equation}}
\def\bna{\begin{eqnarray}}
\def\ena{\end{eqnarray}}

\def\d{\delta}

\def\F{{\mathcal F}}

\def\ln{\mbox{ln}}

\def\HH{\mathscr H}

\def\ban{\begin{eqnarray*}}
\def\ean{\end{eqnarray*}}
\def\bna{\begin{eqnarray}}
\def\ena{\end{eqnarray}}
\def\bnaa{\begin{eqnarray}}
\def\enaa{\end{eqnarray}}
\def\bann{\begin{eqnarray*}}
\def\eann{\end{eqnarray*}}

\ifCLASSINFOpdf
\else
\fi
\begin{document}

\title{Online Regularized Statistical Learning in Reproducing Kernel Hilbert Space With Non-Stationary Data}
%
%
%

\author{Yan Chen, Tao Li,~\IEEEmembership{Senior Member,~IEEE},  and Xiwei Zhang
\thanks{This work was supported by the National Natural Science Foundation
of China under Grant 62261136550. The authors are listed in
alphabetical order. \emph{(Corresponding author: Tao Li.)} }
\thanks{Yan Chen is   with the State Key Laboratory of Mathematical Sciences, Academy of Mathematics and Systems Science, Chinese Academy of Sciences, Beijing 100190, China (e-mail:yanchen2026@amss.ac.cn).}
\thanks{Tao Li is with the Key Laboratory of Management, Decision and Information Systems, Institute of Systems Science, Academy of Mathematics and Systems Science, Chinese Academy of Sciences,  Beijing 100190, China, and also with School of
Mathematical Sciences, University of Chinese Academy of Sciences, Beijing 100149, China (email: litao@amss.ac.cn).}
\thanks{Xiwei Zhang is with the No.2 High School of East China Normal University, Shanghai, 201203, China (e-mail: xwzhangmath@sina.com
).}
}

%
%

\markboth{Journal of \LaTeX\ Class Files, December~2023}%
{Shell \MakeLowercase{\textit{et al.}}: Bare Demo of IEEEtran.cls for IEEE Journals}
%



\maketitle

\begin{abstract}
We study recursive regularized learning algorithms in the reproducing kernel Hilbert space (RKHS) with non-stationary online data streams. We introduce the concept of a \textit{random Tikhonov regularization path} and decompose the  tracking error of the algorithm's output for  the regularization path into random difference equations in RKHS. We show that  the  tracking error vanishes in mean square and almost surely if the regularization path is slowly time-varying. Then, leveraging the monotonicity of inverse operators and the spectral decomposition of compact operators, and introducing the \textit{RKHS persistence of excitation} condition, we  develop a dominated convergence method to prove the mean square  and almost sure consistency between  the regularization path and the unknown function to be learned. Especially, for independent and non-identically distributed data streams, the mean square and almost sure  consistency between the algorithm's output and the  unknown function is achieved if the input data's marginal probability measures are slowly time-varying and the average measure over each fixed-length time period is uniformly above a strictly positive finite Borel measure.
\end{abstract}


\section{Introduction}
\label{sec:introduction}
\IEEEPARstart{S}{upervised}
statistical learning aims to effectively approximate the mapping relationship between inputs and outputs by training  datasets. A crucial aspect of this endeavor is to control the complexity of the hypothesis space. The reproducing kernel Hilbert space (RKHS), a prevalent hypothesis space in the nonparametric regression, offers a unified framework for generalized smooth spline function spaces as well as finite bandwidth real-analytic function spaces (\cite{Wahba}). The consistency and the optimal rate of the offline batch learning algorithms in RKHS with independent and identically distributed (i.i.d.) datasets have been systematically investigated (\cite{Smale3}-\cite{Bousselmi}).

\subsection{Online learning with non-i.i.d. data}

In fact, i.i.d. datasets are difficult to   obtain  in many application scenarios. For instance, for speech recognition and system diagnosis, data  usually exhibits intrinsically temporal correlations, leading to dependent and non-stationary properties (\cite{Steinwart}). Many scholars have long been dedicated to weakening the stringent assumption of i.i.d. data in statistical learning (\cite{Steinwart}-\cite{Ziemann}).
The above works concentrated on offline batch learning algorithms, and relied on the mixing and ergodic nature of the datasets.
In the past two decades, online statistical learning has been widely studied. Compared with offline batch learning, which processes the entire dataset at once, online learning processes one data sample at a time and updates the output in real time, which can reduce  the computational complexity as well as the storage of data. Studies of online learning with non-i.i.d. data have achieved promising results in specific applications (\cite{Agarwal}-\cite{Godichon}). Agarwal and Duchi (\cite{Agarwal}) extended the results on the generalization ability of online algorithms with i.i.d. samples to the cases of stationary $\beta$-mixing and $\phi$-mixing ones. Xu et al. (\cite{Xutangzou}) established the bound on the misclassification error of an online support vector machine (SVM) classification algorithm with uniformly ergodic Markov chain samples. Kuznetsov and Mohri (\cite{Kuznetsov1}) provided generalization bounds for finite-dimensional time series predictions with  non-stationary data. Godichon-Baggioni and Werge (\cite{Godichon}) analyzed the stochastic streaming descent algorithms with weakly time-dependent data for finite-dimensional stochastic optimization problems.

\subsection{{Online learning in RKHS with i.i.d. data}}
The theoretical understanding of convergence properties of online learning algorithms in RKHS is not yet well-established. Fruitful results on convergence of online learning algorithms based on i.i.d. data streams have been obtained (\cite{Smale4}-\cite{gcs}). Smale and Yao (\cite{Smale4}) provided the rate at which the output of the online regularized algorithm is consistent with the deterministic Tikhonov regularization path, by appropriately choosing a fixed regularization parameter. Yao (\cite{Yao}) later proposed the bound of the probability that the output of the algorithm is consistent with the regression function, where decaying regularization parameters were considered. Ying and Pontil (\cite{Ying}) analyzed the mean square error between the output of the online regularized algorithm and the regression function in finite horizons. Tarr$\grave{\text{e}}$s and Yao (\cite{Tarres}) proved that if the regression function satisfies certain regularity conditions (priori information), then the online regularized learning algorithm achieves the same optimal consistency rate as the offline batch learning. Dieuleveut and Bach (\cite{Dieuleveut}) considered the random-design LS regression problem within the RKHS framework, and showed that the averaged non-regularized  algorithm with a given sufficient large step-size can attain optimal rates of consistency for a variety of regimes for the smoothness of the optimal prediction function in RKHS. More results on non-regularized online algorithms can be found in \cite{YYing}-\cite{gcs}. It is worth noting that all of the above works on online learning require i.i.d. data. Smale and  Zhou (\cite{Smale5}) and Hu and Zhou (\cite{Hu}) further investigated online regularized statistical learning algorithms in RKHS with  independent and non-identically distributed online data streams. Smale and Zhou (\cite{Smale5}) obtained the convergence rate of the online regularized learning algorithm if the marginal probability measures of the observation data converge exponentially in the dual of the H\"{o}lder space and the regression function satisfies the regularity condition associated with the limiting probability measure. Subsequently, Hu and Zhou (\cite{Hu}) gave the convergence rates of the LS regression and SVM algorithms with general loss functions, respectively, under the condition that the marginal probability measures of the observation data satisfy the polynomial-level convergence condition.

Motivated by the non-stationary online data in practical real-time scenarios of information processing,
we study the convergence of recursive regularized learning algorithm in RKHS with dependent and non-stationary online data streams. Removing the assumption of time-independent data  inherently complicates the consistency analysis of online algorithms,  and the existing methods which typically rely on independence-based properties are no longer applicable. For non-regularized online learning algorithms, Smale and Yao (\cite{Smale4}), Yao (\cite{Yao}), Ying and Pontil (\cite{Ying}), Dieuleveut and Bach (\cite{Dieuleveut}), and Guo and Shi (\cite{gzc}) utilized the properties of i.i.d. data to equivalently transform the estimation error equations to a special class of  random difference equations, where the homogeneous term is deterministic and time-invariant and the non-homogeneous term is a martingale difference sequence with values in the Hilbert space. Using the spectral decomposition properties of compact operators, they derived mean square consistency results for the algorithms. For regularized online learning algorithms, Smale and Yao (\cite{Smale4}), Yao (\cite{Yao}), Ying and Pontil (\cite{Ying}), and Tarr$\grave{\text{e}}$s and Yao (\cite{Tarres}) initially studied the error between the output of the regularized algorithm and the Tikhonov regularization path of the regression function. They proved the convergence of the homogeneous part of the random difference equation with the help of regularization parameters, and further decomposed the non-homogeneous part into martingales according to the independence of online data streams. Especially, Yao (\cite{Yao}), and Tarr$\grave{\text{e}}$s and Yao (\cite{Tarres}) transformed the online statistical learning in RKHS with i.i.d. data streams into an inverse problem with a deterministic time-invariant Hilbert-Schmidt operator. Then they employed the singular value decomposition (SVD) for linear compact operators in the Hilbert space to derive the consistency results. All the methodologies mentioned above require that the estimation error equation is a random difference equation whose non-homogeneous term is a sequence of martingale differences or reverse martingale differences with values in the Hilbert space by data independence, and rely on the spectral properties of deterministic and time-invariant compact operators. Therefore, all these methods are not applicable for the online statistical learning in RKHS with non-stationary data, which comes down to an inverse problem with randomly time-varying forward operators without independency.
Notably,   the techniques of using blocks of dependent random variables with martingale concentration inequality used in \cite{Agarwal}-\cite{Xutangzou} all rely on the stationary distribution of data, which are also not applicable for non-stationary data.
\subsection{Finite-dimensional parameter estimation with non-i.i.d. data}

From a historical side, aiming to solve the problems of finite-dimensional parameter estimation and signal tracking, many scholars have proposed the persistence of excitation (PE) conditions based on the positivity conditions on the observation/regression matrices (\cite{Green}). Guo (\cite{guo1}) was the first to propose the stochastic PE condition in the analysis of Kalman filtering algorithms  with non-stationary and dependent data. Later, Zhang et al. (\cite{ZGC1991}), Guo (\cite{guo2}), Guo and Ljung (\cite{guo3}) and Guo et al. (\cite{guo4}) generalized the PE condition, and proved that if the regression vectors are $\phi$-mixing, then the PE condition is necessary and sufficient for the exponential stability of the algorithm. The above finite-dimensional PE conditions in \cite{guo1}-\cite{guo4} all require, to some extent, that the auto-covariance matrix of the regression vectors is positive definite, i.e. all its eigenvalues  have a common strictly positive lower bound. Obviously, this does not hold for the statistical learning problems in an infinite-dimensional RKHS.  It is known that even if the data-induced auto-covariance operator in an RKHS is strictly positive, the infimum of its eigenvalues is still zero. To this end,   Zhang et al. (\cite{LZ}) proposed the infinite-dimensional spatio-temporal PE condition for the convergence of  decentralized non-regularized online algorithms in an RKHS, i.e. the  data-induced conditional auto-covariance operators converge  to a strictly positive deterministic time-invariant compact operator in mean square. Note that this condition imposes a convergence requirement
 on the sequence of  data-induced conditional auto-covariance operators even for independent and non-identically distributed data streams.

\subsection{Main contributions}
To address the challenges posed by the removal of independence and stationarity assumptions on the data,
we introduce the concept of a \textit{random Tikhonov regularization path}, which is the optimal solution of the randomly time-varying Tikhonov regularized mean square error (MSE) minimization problem in RKHS.
It is shown that the statistical learning problem in RKHS with  online data streams is an ill-posed inverse problem involving a sequence of randomly time-varying forward operators. We show that the forward operator at each time instant is just the \textit{conditional auto-covariance operator induced by the input data}, and clarify that the process of approximating the unknown function by the \textit{random Tikhonov regularization path} is essentially a regularization method for solving the above random inverse problem.

We investigate the relationship between the output of the algorithm and the random Tikhonov regularization path. By choosing the appropriate algorithm gains and regularization parameters, we obtain a structural decomposition of the tracking error of the algorithm's output for  the regularization path, which shows that the tracking error is jointly determined by the multiplicative noise depending on the random input data, the sampling error of the  regularization path with respect to the input data, and the drift of the regularization path. Tarr$\grave{\text{e}}$s and Yao (\cite{Tarres}) showed that for the case with i.i.d. data streams, the tracking error vanishes in mean square if the drift of the regularization path is slowly time-varying in some sense. To remove the reliance on the independence and stationarity of the data, we equivalently decompose the tracking error equation into two types of random difference equations in RKHS, where the non-homogeneous terms are the martingale difference sequence and the drifts of the regularization paths  respectively, and further investigate the mean square asymptotic stabilities of these two types of difference equations. On this basis, we show that if the random Tikhonov regularization path is slowly time-varying in some sense, then the tracking error vanishes in mean square and almost surely.

The time-varying \textit{conditional auto-covariance operator induced by the input data} in the random Tikhonov regularization path brings the difficulty in the  consistency between the  regularization path and the unknown function to be learned.
To this end,  based on  operator theory, particularly the monotonicity of the inverses of operators and the spectral decomposition  of compact operators, we introduce the \textit{RKHS persistence of excitation} condition, i.e. there exists  a fixed-length time period, such that the accumulated \textit{conditional auto-covariance operator induced by the input data}  over every time period is uniformly greater than a strictly positive compact random operator in the sense of operator order,  and  develop a dominated convergence method to  show  the consistency.
%
%
%
%
Consequently, we show that if the  regularization path is slowly time-varying, and the data stream satisfies the \textit{RKHS persistence of excitation} condition, then the random Tikhonov regularization path is consistent with the unknown function to be learned in mean square and almost surely as the regularization parameter vanishes. This together with the convergence of the tracking error of the algorithm's output for  the random Tikhonov regularization path gives the consistency between  the algorithm's output and  the unknown function. As a special case, for independent and non-identically distributed online data streams, we show that the algorithm achieves mean square and almost sure consistency if the data-induced marginal probability measures are slowly time-varying  and the average measure of the marginal probability measure series over each fixed-length time period is uniformly above a strictly positive finite Borel measure.

The rest of this paper is organized as follows.  Section II gives the statistical learning model in RKHS. Section III defines the random Tikhonov regularization path of the regression function and proposes an online regularized iterative learning algorithm in RKHS. Section IV gives the main results. Section V  gives the special case  with independent and non-identically distributed online data streams. Section VI gives the numerical examples. Section VII concludes the paper. The proofs of all propositions, lemmas, corollaries and theorems in Sections III-V can be found in appendix.

The following notations will be used throughout the paper.
Denote $\mathbb R^n$ as the $n$-dimensional real vector space, $\mathbb N$ as the set of nonnegative integers, and $(\Omega,\F,\mathbb P)$ as a complete probability space.
Let $(\mathscr V,\|\cdot\|_{\mathscr V})$ be a Banach space.
Denote $\mathscr B(\mathscr V)$ as the Borel $\sigma$-algebra of the Banach space $(\mathscr V,\|\cdot\|_{\mathscr V})$. For any given sets $\{A_i,i\in \mathscr I\}$,
where $\mathscr I$ is a set of indices, denote $\sigma\left(\bigcup_{i\in \mathscr I}A_i\right)$ by $\bigvee_{i\in \mathscr I}A_i$.  Let $L^0(\Omega;\mathscr V)$ be a linear space composed of all mappings which take values in $\mathscr V$ and are strongly $\P$-measurable  with reference to $(\Omega,\F,\mathbb P)$. In particular, for  a sub-$\sigma$-algebra $\mathscr G$ of $\F$, $L^0(\Omega,\mathscr G;\mathscr V)$ is defined with reference to  $(\Omega,\mathscr G,\mathbb{P}|_{\mathscr G})$.
For $f\in L^0(\Omega;\mathscr V)$,
denote $\|f\|_{L^p(\Omega;\mathscr V)}:=(\int_{\Omega}\|f\|^p_{\mathscr V}\,\mathrm{d}\mathbb P)^{\frac{1}{p}}$, $1\leq p<\infty$,
and
denote the $\sigma$-algebra generated by $f$ as $\sigma(f):=\{f^{-1}(B):B\in \mathscr B(\mathscr V)\}$.
Denote $L^2(\Omega,\mathscr G;\mathscr V)=\{f\in L^0(\Omega,\mathscr G;\mathscr V): \|f\|_{L^2(\Omega;\mathscr V)}<\infty \}.$
Let $\{\F_k,k\in\mathbb N\}$ be a filtration in the probability space $(\Omega,\F,\mathbb P)$, where $\F_{-1}=\{\emptyset,\Omega\}$.
If $\{f_k,\F_k,k\in\mathbb N\}$ is an adapted sequence, $f_k$ is Bochner integrable over $\F_{k-1}$ and satisfies $\E[f_k|\F_{k-1}]=0$, $\forall~ k\in\mathbb N$, then $\{f_k,\F_k,k\in\mathbb N\}$ is called a martingale difference sequence.
Denote $\mathscr L(\mathscr Y,\mathscr Z)$ as the linear space consisting of all bounded linear operators mapping from the Banach space $\mathscr Y$ to the Banach space $\mathscr Z$, $\mathscr L(\mathscr Z):=\mathscr L(\mathscr Z,\mathscr Z)$.
For any given Hilbert space $ (\mathscr V, \langle \cdot,\cdot \rangle_{\mathscr V})$ and self-adjoint operator $A \in \mathscr L(\mathscr V)$, if $\langle Ax,x \rangle_{\mathscr V} \geq 0, \ \forall \ x \in \mathscr V$, then $A$ is positive.
For any given bounded linear self-adjoint operators $A,B$, if $A-B$ is positive, then we denote $A\succeq B$.
Denote the smallest eigenvalue of the real symmetric matrix $A$ as $\Lambda_{\min}(A)$.
Let the set of eigenvalues of the compact operator $T$ be $\{\Lambda_i(T),i=1,2,\cdots\}$, where $\Lambda_i(T)$ is the $i$-th largest eigenvalue of $T$.
Let $\X$ be a   subset of $\mathbb R^n$.
Denote $\mathcal M(\X)$ as the space of finite Borel signed measures on $\X$  and $\mathcal M_{+}(\X)$ as the subspace consisting of all positive finite measures in $\mathcal M(\X)$.
Denote $C(\X)$ as the whole continuous functions defined on $\X$.
For any $\alpha,\beta\in \mathcal M(\X)$, if $\alpha-\beta\in \mathcal M_{+}(\X)$, then we denote $\alpha\ge \beta$.
Given $\gamma\in \mathcal M_{+}(\X)$, we say that $\gamma$ is strictly positive if for any nonempty open set $U$ in $\X$,  $\gamma(U) > 0$.
Given a sequence of real numbers $\{a_k,k\in\mathbb N\}$ and a sequence of positive real numbers $\{b_k,k\in\mathbb N\}$, if $\lim_{k\to\infty}\sup\frac{|a_k|}{b_k}<\infty$, then we write $a_k=O(b_k)$.
Let $a_k=o(b_k)$ if $\lim_{k\to\infty}\frac{a_k}{b_k}=0$.
Denote $\lceil x \rceil$ as the smallest integer not less than $x$.

\section{Statistical learning model in RKHS}
We study online statistical learning in an RKHS, focusing on approximating an unknown function in RKHS using online data streams. First, we provide the definition of RKHS.
\begin{definition}[\cite{Theodoridis}]\label{definitionofrkhs}
  \rm{Let  $\HH$ be a real Hilbert space consisting of real-valued functions defined on an input space $\X \subseteq \mathbb R^n$ and equipped with the inner product  $\langle \cdot,\cdot\rangle_{\HH}$. The
   space $\HH$ is called an RKHS, if there exists a function
$
K: \X \times \X \to \mathbb{R}
$
with the following properties.}
\begin{itemize}
  \item For every $x \in \X, K_x$ belongs to $\HH$;
  \item $K(\cdot, \cdot)$ has the so-called reproducing property, that is,
$
f(x)=\langle f, K_x\rangle_{\HH},\ \forall \ f \in \HH, \ \forall \ x \in \X,
$
\end{itemize}
where $K_x=K(\cdot,x).$
\end{definition}

In  Definition \ref{definitionofrkhs}, $K$ is called a reproducing kernel of $\HH$.  If $K(\cdot,\cdot):\X\times\X\to\mathbb R$ is a symmetric function, and for   any given $m=1,2,\ldots$, $\alpha_{1},\ldots,\alpha_{m}\in \mathbb{R}$ and   $x_{1},\ldots,x_{m}\in \X$, we always have $\sum_{i=1}^{m}\sum_{j=1}^{m}\alpha_{i}\alpha_{j} K(x_{j},x_{i})\geq 0,$
  then $K$ is called a  positive definite kernel   (\cite{Theodoridis}). The positive definite kernel $K$ ensures that there exists a unique  RKHS,
 denoted by $(\HH_K, \langle \cdot, \cdot \rangle_{\HH_K})$,
for which $K$ is the reproducing kernel. If $K$ is also continuous, then $(\HH_K, \langle \cdot, \cdot \rangle_{\HH_K})$ is separable (\cite{Steinwart1}).


We consider the measurement equation at instant $k$  given by
\setlength{\abovedisplayskip}{2pt}   
\setlength{\belowdisplayskip}{2pt}   
\bna\label{model}
y_k=f^{\star}(x_k)+v_k,~k\in \mathbb N,
\ena
where the random vector $x_k: (\Omega,\mathcal{F})\to (\mathscr X,\mathscr B(\mathscr X))$, the random variables $y_k:(\Omega,\mathcal{F})\to (\mathbb R,\mathscr B(\mathbb R))$ and   $v_k:(\Omega,\mathcal{F})\to (\mathbb R,\mathscr B(\mathbb R))$  are the input data,  the output data  and   the observation noise at instant $k$, respectively. Online statistical learning  aims to recursively construct an estimate $f_{k}$ of the unknown function $f^{\star}$  in  a hypothetical RKHS  at each instant, using the current observation data $(x_k,y_k)$ and the estimate $f_{k-1}$ at the last instant.

For the statistical learning model (\ref{model}), we have the following assumptions.

\begin{assumption}\label{ass2}
\rm{ The unknown function $f^{\star} \in  \HH_K$, where $K$ is a uniformly continuous positive definite kernel and $\sup_{x \in \X}K(x,x) <\infty$.
}
\end{assumption}
\vskip 0.2cm

\begin{assumption}\label{ass1}
\rm{(i)} \rm{ There exists a filtration $\{\F_k,\ k\in \mathbb N \}$ such that both $\{v_k,\F_k,k\in \mathbb N\}$ and $\{v_kK_{x_k},\F_k,k\in \mathbb N\}$ are martingale difference sequences; \rm{(ii)} there exists a constant $\b>0$, such that $\sup_{k\in \mathbb N}\E\left[v_k^2|\F_{k-1}\right]\leq \b$\ a.s. }
\end{assumption}
\begin{remark}\label{fangzhenzhu}
\rm{
Bousselmi et al. (\cite{Bousselmi}) assumed that the data stream $\{(x_k,y_k),\ k\in \mathbb N\}$ and the observation noise $\{v_k,\ k\in \mathbb N\}$ in the  model (\ref{model}) are both i.i.d., whereas Assumption \ref{ass1} (i) holds if $\{v_k,\ k\in \mathbb N\}$ is a martingale difference sequence, $v_k$ and $K_{x_k}$ are conditionally uncorrelated with respect to $\F_{k-1}$. In particular, if  $\{v_k,\ k\in \mathbb N\}$ is a martingale difference sequence independent of $\{x_k,k\in \mathbb N\}$, then by Proposition B.5 in \cite{LZ},  one obtains   $\E[v_kK_{x_k}|\F_{k-1}] = \E[v_k|\F_{k-1}] \E[K_{x_k}|\F_{k-1}]=0$,  that is,  Assumption \ref{ass1} (i) holds. \hfill \hfill  \hfill $\blacksquare$

}
\end{remark}
\vskip 0.2cm

\begin{remark}
\rm{
The statistical learning model based on i.i.d. sampling   in \cite{Smale4}-\cite{Lin2} can be regarded as a special case of the statistical learning based on the measurement model (\ref{model}),  and both Assumptions \ref{ass2} and \ref{ass1} hold. They focus on a fixed joint probability distribution $\rho$ with a sample space $\X\times \mathscr Y$, $\mathscr Y\subseteq \mathbb R$, that is, that is, the random vector $Z=(X,Y)\sim\rho$, from which the data stream $\{(x_k,y_k),k\in \mathbb N\}$ is generated by independently sampling. The regression function
\bna\label{regression}
f_{\rho}(x):=\int_{\mathscr Y}y\,\mathrm{d}\rho_{\mathscr Y|x},~\forall~ x\in \X,
\ena
where $\rho_{\mathscr Y|x}$ is the conditional probability distribution on $\mathscr Y$ given $x\in \X$, is the optimal solution of the following MSE problem
$\arg\min_{f\in \mathscr L^2_{\rho_{\X}}}\int_{\X\times \mathscr Y}(f(x)-y)^2\,\mathrm{d}\rho,$
where $\rho_{\X}$ is the marginal probability distribution induced by $\rho$ over $\X$ and $\mathscr L^2_{\rho_{\X}}$ is the Hilbert space formed by all measurable functions which are square integrable with respect to $\rho_{\X}$. The regression function $f_{\rho}$ can be  approximated by the online learning algorithms in RKHS  (\cite{Smale4}-\cite{Lin2}).
Define $L_K:\mathscr L^2_{\rho_{\X}}\to\mathscr L^2_{\rho_{\X}}$ as the integral operator defined by the positive definite kernel $K$ and the marginal probability distribution $\rho_{\X}$, i.e.
\bna\label{jifensuanzi}
L_Kf(t):=\int_{\X}K(t,x)f(x)\,\mathrm{d}\rho_{\X}(x),~\forall~ f\in \mathscr L^2_{\rho_{\X}}.
\ena
The compactness of $L_K$ guarantees the existence of the orthonormal eigensystem $(\mu_k,\varphi_k, k\in\mathbb N)$ in $\mathscr L^2_{\rho_{\X}}$(\cite{Smale4}, \cite{Tarres}). For any $r>0$, define $L^r_K:\mathscr L^2_{\rho_{\X}}\to\mathscr L^2_{\rho_{\X}}$ as
$
L^r_K\left(\sum_{k=0}^{\infty}c_k\varphi_k\right)=\sum_{k=0}^{\infty}c_k\mu^r_k\varphi_k,~\forall~c_k\in\mathbb R,~\forall~k\in\mathbb N.
$
It is worth noting that, the regression function is required to satisfy a certain regularity condition (priori information) in \cite{Smale4}-\cite{Lin2}, that is, there exists a constant $r>0$ such that $f_{\rho}\in L_K^r(\mathscr L^2_{\rho_{\X}})$.
By the isometrical isomorphism of Hilbert spaces: $L_K^{1/2}(\mathscr L^2_{\rho_{\X}})=\HH_K$ and $L_K^{s}(\mathscr L^2_{\rho_{\X}})\subseteq L_K^{t}(\mathscr L^2_{\rho_{\X}})$, $\forall~s\ge t>0$ (\cite{Smale4}, \cite{Tarres}), the above regularity condition implies that if  $r\ge 1$, then $f_{\rho}\in\HH_K$.

To verify Assumption \ref{ass1}, we take  the filtration  $\F_k = \bigvee_{i=0}^k (\bigvee_{x\in \X}   \sigma  (K(x,x_i)  ) \bigvee  \sigma \left(y_i\right)  )$,    $\forall ~k\in\mathbb N$,
where $\F_{-1}=\{\emptyset,\Omega\}$.
Let $v_k=y_k-f_{\rho}(x_k)$. Then
$
y_k=f_{\rho}(x_k)+v_k.
$
Since $(x_k,y_k)\sim\rho$, then it follows from Fubini  theorem and (\ref{regression}) that
$\E[v_k|\F_{k-1}]=\int_{\X\times \mathscr Y}\left(y-f_{\rho}(x)\right)\,\mathrm{d}\rho=\int_{\X}\left(\int_{\mathscr Y}y-f_{\rho}\left(x\right)\,\mathrm{d}\rho_{\rho_{\mathscr Y|x}}\right)\,\mathrm{d}\rho_{\X}(x)=0,~\forall ~k\in\mathbb N.$
Similarly, we have
$\E[v_kK_{x_k}|\F_{k-1}]=\int_{\X\times \mathscr Y} (y -f_{\rho}\left(x\right) )K_x\,\mathrm{d}\rho  =0,~\forall ~k\in\mathbb N.$
Additionally, in \cite{Smale4}-\cite{Lin2}, it was assumed that $\E[Y^2]<\infty$ and $\sup_{x\in\X}K(x,x)<\infty$, which means that there exists a constant $\b>0$, such that $\sup_{k\in \mathbb N}\E[v_k^2]\leq \b$.
\hfill  $\blacksquare$
}
\end{remark}

\section{Online learning algorithm in RKHS}\label{dierjie}

The statistical learning problem can be viewed as a random inverse problem, which is typically ill-posed. To this end, we introduce the concept of a random Tikhonov regularization path for approximating the unknown function.
Then, we propose an online algorithm, which can be interpreted as a tracking scheme of the regularization path. This interpretation naturally leads to a decomposition of the overall  estimation error of the algorithm into an approximation error  and a tracking error, which forms the basis of the subsequent analysis.

\subsection{Random Tikhonov regularization path of the regression function}

For the statistical learning model (\ref{model}) in RKHS, consider the following randomly time-varying Tikhonov regularized MSE problem
\bna\label{youhua}
\arg\min \limits_{\widehat f_k\in L^2(\Omega,\F_{k-1};\HH_K)}J_k(\widehat f_k)~\mathrm{a.s.},~\forall~k\in \mathbb N,
\ena
where $J_k(\widehat f_k)=\frac{1}{2}\E\big[ \big(y_k-\widehat f_k(x_k)\big)^2 +\lambda_k\big\|\widehat f_k\big\|^2_{\HH_K} \big|\F_{k-1}\big]$, $\lambda_k$ is the Tikhonov regularization parameter, $\|f\|_{\HH_K}=\sqrt{\langle f,f \rangle_{\HH_K} }$, $\forall~ f\in \HH_K$.

For any  $x\in \X$, define the operator $(K_x\otimes K_x): \HH_K \to \HH_K$ by $ (K_x\otimes K_x) f=f(x)K_x$.
Assumption \ref{ass2} guarantees the existence and uniqueness of the operator-valued random element $\E[K_{x_k}\otimes K_{x_k}|\F_{k-1}]$.
Denote $H_k=K_{x_k}\otimes K_{x_k}$, $T_k=\E[H_k|\F_{k-1}],\ k\geq 0$. Regarding the optimal solution of (\ref{youhua}), we have the following proposition.

 \vskip 0.2cm
\begin{proposition}\label{mingti6}
\rm{For the statistical learning model (\ref{model}), if Assumptions \ref{ass2}-\ref{ass1} hold, then
 \bna\label{gradientoperator}
\mathrm{grad}\,J_k(f)=\E[(f(x_k)-y_k)K_{x_k}+\lambda_kf|\F_{k-1}]~\mathrm{a.s.},
\ena
where $\mathrm{grad}\,J_k:\HH_K\to\HH_K$ is the gradient operator. The optimal solution $f_{\lambda,k}$ of (\ref{youhua}) satisfies
\bna\label{tidufcdddd}
(T_{k}+\lambda_k I)f_{\lambda,k}=\E\left[y_kK_{x_k}|\F_{k-1}\right]~\mathrm{a.s.},~\forall~k\in \mathbb N,
\ena
where $I:\HH_K\to\HH_K$ is the identity operator. Especially, if $\lambda_k=0$, then  $f_{\lambda,k}=f^{\star}$,
and if $\lambda_k>0$, then
\bna\label{tidufcddddcc}
f_{\lambda,k}=\left( T_{k}+\lambda_k I \right)^{-1}T_{k}f^{\star}~\mathrm{a.s.},~\forall~k\in \mathbb N.
\ena
}
\end{proposition}
\vskip 0.2cm
\begin{proof}
  See Appendix \ref{xiaosiAAA} for the proof.
\end{proof}

\begin{definition}\label{zhengzehua}
\rm{For the model (\ref{model}), if the regularization parameter $\lambda_k>0$, then the solution (\ref{tidufcddddcc}) of (\ref{youhua}) is called the random Tikhonov regularization path of  $f^{\star}$.}
\end{definition}

\begin{remark}\label{pinglun}
\rm{
Regularization paths have been extensively studied in the statistical learning theory (\cite{Tarres}, \cite{Rosset}). LASSO regularization paths are piecewise linear so that the entire regularization paths can be tracked by locating a finite number of change points. Rosset and Zhu (\cite{Rosset}) generalized this property to the case where the loss function and the regularized term are piecewise quadratic and piecewise linear, respectively. Different from this, Tikhonov regularization does not possess piecewise linear paths (\cite{Tarres}). It is worth noting that Proposition \ref{mingti6} shows that the random Tikhonov regularization path of the unknown function $f^{\star}$ uniquely exists with probability $1$, and the explicit form of  $f_{\lambda,k}$ is given by  (\ref{tidufcddddcc}). Especially, if the online data stream $\{(x_k,y_k),k\in \mathbb N\}$ is independently sampled with an identical probability measure $\rho$, i.e. $(x_k,y_k)\sim\rho$, then the randomly time-varying Tikhonov regularized MSE problem (\ref{youhua}) degenerates into the optimization problem based on i.i.d. sampling in \cite{Smale4}-\cite{Tarres}, that is,
$\arg\min\limits_{f\in \HH_K}\E_{(x,y)\sim\rho}\frac{1}{2}\left[(y-f(x))^2+\lambda_k\|f\|^2_{\HH_K}\right],~\lambda_k\ge 0.$
Meanwhile, the random Tikhonov regularization path degenerates into the regularization paths in \cite{Smale4}-\cite{Tarres}, that is,
$
f_{\lambda,k}=\left(\E_{x\sim\rho_{\X}}\left[K_x\otimes K_x\right]+\lambda_kI\right)^{-1}\E_{x\sim\rho_{\X}}\left[K_x\otimes K_x\right]f^{\star}
= (L_K +\lambda_kI )^{-1}L_Kf^{\star},~\forall~k\in\mathbb N,
$
where the integral operator $L_K$ is given by (\ref{jifensuanzi}).

The statistical learning problems in RKHS are essentially the random inverse problems in the Hilbert spaces (\cite{LZ}), and the regularization paths are inextricably linked to resolving the inverse problems (\cite{Smale4}-\cite{Yao}, \cite{Tarres}).
To make this connection explicit, we next show how the statistical learning model can be rewritten as a random inverse problem. By the reproducing property of RKHS, multiplying both sides of (\ref{model}) by $K_{x_k}$ yields $y_kK_{x_k}=f^{\star}(x_k)K_{x_k}+v_kK_{x_k}=H_k f^{\star}+v_kK_{x_k}$. Suppose that Assumptions \ref{ass2}-\ref{ass1} hold. Taking the conditional expectations on the both sides   with respect to $\F_{k-1}$ leads to
\bna\label{fanwenti}
T_kf^{\star}=z_k,~\forall~k\in \mathbb N,
\ena
where  $z_k=\E[y_kK_{x_k}|\F_{k-1}]$. In Definition 1 of \cite{ZXWLT}, $\E[H_k]$ is called an auto-covariance  operator. Here,   we call $T_k$ the \textit{conditional auto-covariance operator induced by the input data}. It follows from Proposition D.3  in \cite{Zhang} that $T_k$ is a self-adjoint operator which is almost surely compact, and by the spectral decomposition of the compact operator, the condition number of the forward operator $T_k$ satisfies $\kappa(T_k)=\|T_k^{-1}\|\|T_k\|=\infty$ a.s. Therefore, resolving $f^{\star}$ from (\ref{fanwenti}) is a randomly time-varying  ill-posed inverse problem. Notably, it can be seen that $T_k=\E[H_k]=L_K$ if the data stream $\{(x_k,y_k),k\in\mathbb N\}$ is sampled independently from a common joint distribution $\rho$, and then (\ref{fanwenti}) degenerates into the inverse problem with the deterministic time-invariant forward operator studied in \cite{Smale4}-\cite{Yao} and \cite{Tarres}, i.e.,
\bna\label{fwt}
L_Kf^{\star}=z.
\ena

Using  the Tikhonov regularization strategy, the corresponding well-posed equations for the ill-posed equations (\ref{fanwenti}) are
\bna\label{shidingwenti}
(T_k+\lambda_kI)u(k)=z_k,~\forall~k\in \mathbb N.
\ena
If Assumptions \ref{ass2}-\ref{ass1} hold, then by Proposition \ref{mingti6}, the solution of the well-posed equation (\ref{shidingwenti}) is $u(k)=f_{\lambda,k}$ a.s. This means that $f_{\lambda,k}$ is the Tikhonov regularization path of the solution of the ill-posed equation (\ref{fanwenti}).
\hfill $\blacksquare$
}
\end{remark}

\subsection{Online regularized learning algorithm in RKHS}

By (\ref{gradientoperator}) in Proposition \ref{mingti6}, we have $ \mathrm{grad}\,J_k(f)=\E[(f(x_k)-y_k)K_{x_k}+\lambda_kf|\F_{k-1}]$  a.s. Hence, we have
$$\E[(f(x_k)-y_k)K_{x_k}+\lambda_kf-\mathrm{grad}\,J_k(f)|\F_{k-1}]=0~\mathrm{a.s.},$$
which shows that $(f(x_k)-y_k)K_{x_k}+\lambda_kf$ is an unbiased estimate of the gradient $\mathrm{grad}\,J_k(f)$ with respect to $\F_{k-1}$.
Based on (\ref{youhua}) and the stochastic gradient descent method, the online regularized statistical learning algorithm in RKHS is given by
\bna\label{AL}
f_{k+1}=f_k-a_k\left((f_k(x_k)-y_k)K_{x_k}+\lambda_k f_k\right),~\forall~k\in \mathbb N,
\ena
where $f_0\in\HH_K$, $a_k$ is the algorithm gain and $\lambda_k$ is the regularization parameter.
\vskip 0.2cm

\begin{remark}
\rm{Within the realm of results on RKHS online learning with independent data streams, (\ref{AL}) is referred to as the online regularized algorithm (\cite{Smale4}-\cite{Yao}, \cite{Tarres}, \cite{Smale5}-\cite{Hu}) if the regularization parameter $\lambda_k>0$. For   $\lambda_k=0$, it is called the non-regularized online algorithm (\cite{Ying}, \cite{Dieuleveut}-\cite{gzc}, \cite{LZ}).}
\end{remark}
\vskip 0.2cm

For the algorithm gains and the regularization parameter in the algorithm (\ref{AL}), we need the following condition.

\begin{condition}\label{con1}
\rm{
The sequences of gains $\{a_k,k\in\mathbb N\}$ and regularization parameters $\{\lambda_k,k\in \mathbb N\}$ satisfy \[a_k=\frac{\alpha_{1}}{(k+1)^{\tau_1}},\quad \lambda_k=\frac{\alpha_{2}}{(k+1)^{\tau_2}},~\forall~ k\in \mathbb N,\]
where $\alpha_{1},\ \alpha_{2}$,  $\tau_1,\ \tau_{2}>0$,  $\tau_1+\tau_2<1,~3\tau_2<\tau_1$.
}
\end{condition}

\section{Convergence analysis}\label{disanjie}

In this section, we will investigate the  consistency of the algorithm (\ref{AL}) in RKHS.

Proposition \ref{mingti6} indicates that the optimal solution to the optimization problem (\ref{youhua}) is the random Tikhonov regularization path $f_{\lambda,k}$ of $f^{\star}$. Therefore, we first consider the relationship between the algorithm's output $f_k$ and $f_{\lambda,k}$.

Denote the tracking error of the algorithm (\ref{AL}) with respect to  $f_{\lambda,k}$ by $\d_k=f_k-f_{\lambda,k}$. Subtracting $f_{\lambda,k+1}$ from both sides of  (\ref{AL}) and by (\ref{tidufcddddcc}), we obtain
\begin{align}\label{EERR}
   \d_{k+1}
=& \left(I-a_k\left(H_{k}+\lambda_kI\right)\right)\d_k+a_kv_kK_{x_k}-a_k ( (H_{k} +\lambda_kI )f_{\lambda,k}
- H_{k} f^{\star} ) -(f_{\lambda,k+1}-f_{\lambda,k}).
\end{align}
Thereby, it is shown that the tracking error $\d_{k+1}$ at instant $k+1$ consists of four terms including (i) tracking error $\d_k$ at instant $k$; (ii) multiplicative noise $v_kK_{x_k}$ depending on the random input data at instant $k$; (iii) the sampling error $\left(H_{k}+\lambda_kI\right)f_{\lambda,k}- H_{k} f^{\star}$ of the random Tikhonov regularization path with respect to the input data $x_k$ at instant $k$; (iv) drift error $f_{\lambda,k+1}-f_{\lambda,k}$ generated by the random Tikhonov regularization path.

 \vskip 1mm
  By Lemmas \ref{LLEEMM6}-\ref{LLEEMM7}, we prove that the tracking error $f_k-f_{\lambda,k}$ converges to zero as shown in the following lemma.

\vskip 0.1cm

\begin{lemma}\label{THMM}
\rm{For the algorithm (\ref{AL}), if Assumptions \ref{ass2}-\ref{ass1} and Condition \ref{con1} hold, and
\begin{itemize}
  \item[(i)]
  \bna\label{hmp}
\lim_{k\to\infty}\hspace{-2pt}\sum_{i=0}^k\hspace{-3pt}\left\|f_{\lambda,i+1}-f_{\lambda,i}\right\|_{L^2\left(\Omega;\HH_K\right)}\hspace{-7pt} \prod_{j=i+1}^k\hspace{-7pt}\left(1-a_j\lambda_j\right)=0,
\ena
then
\begin{align}\label{dkjunf}
 \lim_{k\to\infty}\left\|f_k-f_{\lambda,k}\right\|_{L^2\left(\Omega;\HH_K\right)}=0;
\end{align}
\item[(ii)]
\bna\label{hmp00}
\lim_{k\to\infty}\sum_{i=0}^k\left\|f_{\lambda,i+1}-f_{\lambda,i}\right\|_{\HH_K}\prod_{j=i+1}^k\left(1-a_j\lambda_j\right)=0,
\ena
then
\begin{align}\label{dkjihu}
\lim_{k\to\infty}\left\|f_k-f_{\lambda,k}\right\|_{\HH_K}=0\ \text{a.s.}
\end{align}
\end{itemize}
}
\end{lemma}

\begin{proof}
  See Appendix \ref{fuluBBBB} for the proof.
\end{proof}

\begin{remark}
\rm{
Specifically, the condition (\ref{hmp}) of Lemma \ref{THMM} holds if $\|f_{\lambda,k+1}-f_{\lambda,k}\|_{L^2(\Omega;\HH_K)}=o(a_k\lambda_k)$ (see Lemma III.6 in \cite{Tarres}). From Lemma D.3  in \cite{Zhang}, we can see that the drift of the regularization path is influenced by the drift of the conditional expectation of the operator induced by the input data as well as the regularization parameter, i.e.
\begin{align}\label{xnck}
 \left\|f_{\lambda,k+1}-f_{\lambda,k}\right\|_{L^2\left(\Omega;
\HH_K\right)}
=& O\left(\big(\big(\E\big[ \|\widetilde{\Delta}_k \|^2_{\mathscr L(\HH_K)}\big]\big)^{\frac{1}{2}}+\lambda_k-\lambda_{k+1}\big)
 \lambda_k^{-1}\right),
\end{align}
where $\widetilde{\Delta}_k:=T_{k+1}-T_k$.  As shown in Remark \ref{pinglun}, for the case with i.i.d. data stream $\{(x_k,y_k),k\in \mathbb N\}$, $f_{\lambda,k}$ degenerates into the regularization paths presented in \cite{Smale4} and \cite{Ying}-\cite{Tarres}, and (\ref{xnck}) degenerates to $\|f_{\lambda,k+1}-f_{\lambda,k}\|_{L^2\left(\Omega;\HH_K\right)}=O((\lambda_k-\lambda_{k+1})/\lambda_k)$, which is exactly the bound of the drift error of the regularization path given by Tarr$\grave{\text{e}}$s and Yao (\cite{Tarres}).

Smale and Yao (\cite{Smale4}) gave a convergence rate of the output of the online regularized algorithm with a fixed  regularization parameter. Similar to the offline batch learning, Ying and Pontil (\cite{Ying}) performed the mean square error analysis of online regularized algorithms in finite horizons by selecting the regularization parameter as a function of the sample size up to a given time. As the sample size increases with time in the online learning, the regularization parameter needs to be updated over time to ensure that the output of the algorithm can track the regularization path. For this purpose, Tarr$\grave{\text{e}}$s and  Yao (\cite{Tarres}) proved that if the drift of the regularization path satisfies the slowly time-varying condition (\ref{hmp}), the tracking error of the output of the online regularized algorithm with respect to the regularization path converges to zero. Compared with above works, Lemma \ref{THMM} shows that, with no restrictions on the independence and stationarity of the data, the mean square error between the output of the algorithm (\ref{AL}) and the regularization path converges to zero if the drift of the regularization path is slowly time-varying as in (\ref{hmp}). \hfill  $\blacksquare$
}
\end{remark}

\vskip 0.1cm
Next, we will investigate the approximation error $f_{\lambda,k}-f^{\star}$. We introduce the following definition.
 \vskip 0.1cm

\begin{definition}\label{pe}
\rm{ We say that $\{(x_k,y_k),k\in \mathbb N\}$ satisfies the RKHS persistence of excitation condition, 
if there exists an integer $h>0$ and a strictly positive compact random operator $R\in L^2(\Omega;\mathscr L(\HH_K))$, such that 
\bna\label{suanzi}
\sum_{i=k}^{k+h-1}\E\left[\left. H_{i}\right|\F_{k-1}\right]\succeq R~\mathrm{a.s.},~\forall~k\in\mathbb N.
\ena
}
\end{definition}
\vskip 0.1cm

The following proposition provides a sufficient condition for the above condition under   $\phi$-mixing input data, including deterministic processes, $M$-dependent sequences, and processes generated from bounded white noise filtered through a  stable finite-dimensional linear filter.
\vskip 0.1cm
\begin{proposition}\label{mixing}
\rm{Suppose that the input data stream  $\{(x_k,\F_k),\ k\in \mathbb N\}$ is an $n$-dimensional $\phi$-mixing process, where $\F_k=\sigma(x_i,\ 0\leq i\leq k)$.
If (a) there exists  an integer $h_0>0$ and a strictly positive compact  operator $R_{0}\in \mathscr{L}(\HH_K)$  such that $ \sum_{i=k}^{k+h_0-1} \E\left[ H_{i}\right] \succeq R_{0},~\forall~k\in\mathbb N,$
then $\{(x_k,y_k),k\in \mathbb N\}$ satisfies the RKHS persistence of excitation condition. Especially,  if $\{(x_k,\F_k),\ k\in \mathbb N\}$  is an  $M$-dependent sequence and condition (a) holds, then  $\{(x_k,y_k),k\in \mathbb N\}$ satisfies the RKHS persistence of excitation condition with $h=h_0+M$ and $R=\frac{R_{0}}{2}$.
}
\end{proposition}

 \begin{proof}
  See Appendix \ref{fuluBBBB} for the proof.
\end{proof}

\vskip 1mm
To analyse the difference between $f_{\lambda,k}$ and $f^{\star}$, we develop a dominated convergence method in Lemma \ref{lemma333} based on  operator theory and the RKHS persistence of excitation condition. In this method, we  use  the monotonicity of the inverses of operators and the spectral decomposition of compact operators to give an upper bound of the difference, which together with the dominated convergence theorem shows the decaying of the difference over time. Based upon this, we have the following lemma, whose proof is directly from Lemmas \ref{lemma222}-\ref{lemma333} and is  therefore  omitted.
\vskip 1mm
\begin{lemma}\label{THMM2}
\rm{For the algorithm (\ref{AL}), if Assumptions \ref{ass2}-\ref{ass1} and Condition \ref{con1} hold, the online data stream  $\{(x_k,y_k),k\ge 0\}$  satisfies the RKHS persistence of excitation condition, and
\begin{itemize}
  \item[(i)] the random Tikhonov regularization path satisfies
\bna\label{rsrs}
\left\|f_{\lambda,k+1}-f_{\lambda,k}\right\|_{L^2\left(\Omega:\HH_K\right)}=o\left(\lambda_k\right),
\ena
then
$$
\lim_{k\to\infty}\left\|f_{\lambda,k}-f^{\star}\right\|_{L^2\left(\Omega;\HH_K\right)}=0;
$$
 \item[(ii)]
the random Tikhonov regularization path satisfies, for  $ j=k,\cdots, k+h-2,$
\bna\label{rsrs00}
\mathbb{E}\left[\left\|f_{\lambda,j+1}-f_{\lambda,j}\right\|_{\HH_K}^2|\mathcal{F}_{k-1}\right]^{\frac{1}{2}}=o(\lambda_k)~\text{a.s.}, \
\ena
  then
$$\lim_{k\to\infty}\|f_{\lambda,k}-f^{\star}\|_{\HH_K}=0 \ \text{a.s.}
$$

\end{itemize}

}
\end{lemma}
\vskip 0.2cm
 \begin{proof}
  See Appendix \ref{fuluBBBB} for the proof.
\end{proof}
\vskip 0.2cm

Based on Lemma \ref{THMM}-\ref{THMM2}, the following theorem provides the conditions for the mean square and almost sure consistency of the algorithm.

\vskip 1mm

\begin{theorem}\label{wuxian}
\rm{For the algorithm (\ref{AL}), if Assumptions \ref{ass2}-\ref{ass1} and Condition \ref{con1} hold, the online data stream $\{(x_k,y_k),k\in \mathbb N\}$ satisfies the \textit{RKHS persistence of excitation} condition, and

(i) the random Tikhonov regularization path is slowly time-varying in the sense that
\bna\label{cnm}
\left\|f_{\lambda,k+1}-f_{\lambda,k}\right\|_{L^2\left(\Omega;\HH_K\right)}=o\left(a_k\lambda_k\right),
\ena
then $$\lim_{k\to\infty}\|f_k-f^{\star}\|_{L^2(\Omega;\HH_K)}=0;$$

(ii)  the random Tikhonov regularization path is  slowly time-varying in the sense that
\begin{align}\label{cn11cy}
\left\|f_{\lambda,k+1}-f_{\lambda,k}\right\|_{\HH_K}=o\left(a_k\lambda_k\right),\ \text{a.s.},
\end{align}
\begin{align}
\mathbb{E}\left[\left(\sup_{k\in\mathbb{N}}\frac{\|f_{\lambda,k+1}-f_{\lambda,k}
\|_{\mathscr{H}_K}}{\lambda_k}\right)^2\right]
<\infty,\label{finite}
\end{align}
then $$\lim_{k\to\infty}\|f_k-f^{\star}\|_{\HH_K}=0 \ \text{a.s.}$$
}
\end{theorem}
 \begin{proof}
  See Appendix \ref{fuluBBBB} for the proof.
\end{proof}
\vskip 0.2cm

The following corollary shows the mean square convergence rate for the special case.

\begin{corollary}\label{wuxiannsuanfasud}
\rm{For the algorithm (\ref{AL}), if Assumptions \ref{ass2}-\ref{ass1} and Condition \ref{con1} hold, the online data stream $\{(x_k,y_k),k\in \mathbb N\}$ satisfies the \textit{RKHS persistence of excitation} condition,
 and  there exists $\hat{C}_5>0$, such that
\bna\label{cnmcy}
\left\|f_{\lambda,k+1}-f_{\lambda,k}\right\|_{L^2\left(\Omega;\HH_K\right)}\leq \hat{C}_5 a_k^2,  \ \forall \ k\geq  0,
\ena
then, we have, for any $k\geq k_0$,
\begin{align}
& \left\|f_{k}-f^{*}\right\|_{L^2\left(\Omega;\HH_K\right)}\notag\\
\leq    &    C_7  \frac{\ln^{\frac{3}{2}}(k+1)}{k^{\frac{\tau_1-3\tau_2}{2}}}  \hspace{-3pt}+\hspace{-3pt}\left(\hspace{-3pt}\E\left[\sum_{i=0}^{\infty}\frac{  \sum_{j=k }^{k+h-1}\lambda_j}{2\Lambda(i)+\sum_{j=k }^{k+h-1}\lambda_j}\left|\langle f^{\star},e(i)\rangle_{\HH_K}\right|^2 \right]\right)^{\frac{1}{2}},  \label{sudumean}
\end{align}
 where $k_{0}=
7+ \lceil
 \big(2(\alpha_1\alpha_2+\alpha_1\kappa) \big)^{\frac{1}{\tau_1}}\rceil+
\lceil\big(\frac{32}{\alpha_1\alpha_2(1-\tau_1-\tau_2)}\big)^{\frac{2}{1-\tau_1-\tau_2}}\rceil+
\lceil\frac{\tau_1-3\tau_2}{\alpha_1\alpha_2}\big)^{\frac{2}{1-\tau_1-\tau_2}}\rceil$,   $C_7=2 \hat{C}_5  \alpha_1 (\kappa+\alpha_{2})+  C_6$, $C_6= C_0  ( C_{01} C_3 ( 1 + k_0(\kappa+\alpha_2+1)(\alpha_1^2+\alpha_1)) + C_2 ( \kappa+\alpha_2 + \frac{\tau_1 2^{\tau_1}}{\alpha_1} ) ) + C_{01} C_3\left\|f_{0}\right\|_{L^2\left(\Omega;\HH_K\right)}+
C_{01} C_3 \|f^{\star}\|_{\HH_K}+ \hat{C}_5 C_2
+ 2 C_{01} k_0 C_3 \|f^{*}\|_{\HH_K}$,
$C_0=(2\kappa+\alpha_2 )\|f^*\|_{\HH_K}+\sqrt{\beta} \sqrt{\kappa}$,  $C_{01}=(1+\alpha_{1}\kappa+\alpha_{1}\alpha_{2})^{k_0}$, $C_2=  \alpha_1^2
+
\alpha_1^2 4^{\tau_1}
\left(
\frac{2}{\alpha_1\alpha_2}+2
\right)^{\frac{3}{2}}$, $C_3=
\exp (\frac{\alpha_{1}\alpha_{2}}{1-\tau_1-\tau_2}(k_0+2)^{1-\tau_1-\tau_2} ),\ \kappa=\sup_{x\in \mathscr{X}}K(x,x),$   $\{\Lambda(i),e(i),\  i\in\mathbb N\}$ is  the eigensystem of $R$ with $\{\Lambda(i),\ i\in\mathbb N\} $ arranged in a non-increasing order.
}
\end{corollary}
\vskip 0.2cm
 \begin{proof}
  See Appendix \ref{fuluBBBB} for the proof.
\end{proof}
\vskip 0.2cm

\begin{remark}
On the r.h.s. of \eqref{sudumean}, the first term is explicit, whereas
the second term is associated with the lower-bound operator in the RKHS PE
condition and hence depends on the specific data stream. For a deterministic
lower-bound operator of the form
$
R=\frac{h}{12}\int_{\mathscr X}K_x\otimes K_x\,\gamma(dx),
$
where $\gamma$ is the Lebesgue measure on $\mathscr X$, the relevant
eigensystem can be approximated numerically through the associated kernel
integral eigenvalue problem. In particular, one may discretize the integral
operator on a set of quadrature nodes and compute a
Nystr\"om approximation from the corresponding quadrature-weighted kernel
matrix as shown in \cite{WilliamsSeeger}. The resulting numerical eigenpairs
can then be used to recover an approximate eigensystem of $R$ and to
approximate the second term in \eqref{sudumean} by truncation.
\end{remark}

By the tower property of conditional expectation and Theorem \ref{wuxian}, the following corollary follows immediately.

\vskip 0.1cm
\begin{corollary}\label{tuijil}
\rm{For the algorithm (\ref{AL}), if Assumptions \ref{ass2}-\ref{ass1} and Condition \ref{con1} hold,  there exists an integer $h>0$ and a strictly positive compact operator  $R\in \mathscr L(\HH_K)$, such that 
 $\sum_{i=k}^{k+h-1}T_i\succeq R~\mathrm{a.s.},\  \forall~k\in\mathbb N,$ and

(i) the random Tikhonov regularization path is slowly time-varying in the sense that
\bna
\left\|f_{\lambda,k+1}-f_{\lambda,k}\right\|_{L^2\left(\Omega;\HH_K\right)}=o\left(a_k\lambda_k\right),\notag
\ena
then $$\lim_{k\to\infty}\|f_k-f^{\star}\|_{L^2(\Omega;\HH_K)}=0;$$

(ii)  the random Tikhonov regularization path is conditionally slowly time-varying in the sense that
\bna
\left\|f_{\lambda,k+1}-f_{\lambda,k}\right\|_{\HH_K}=o\left(a_k\lambda_k\right),\ \text{a.s.},\notag
\ena
\begin{align}
\mathbb{E}\left[\left(\sup_{k\in\mathbb{N}}\frac{\|f_{\lambda,k+1}-f_{\lambda,k}
\|_{\mathscr{H}_K}}{\lambda_k}\right)^2\right]
<\infty,\notag
\end{align}
then $$\lim_{k\to\infty}\|f_k-f^{\star}\|_{\HH_K}=0 \ \text{a.s.}$$
}
\end{corollary}

 \begin{proof}
  See Appendix \ref{fuluBBBB} for the proof.
\end{proof}

\begin{remark}
\rm{
It follows from Assumption \ref{ass2} and Proposition D.3  in \cite{Zhang} that $\E[H_{i}|\F_{k-1}]$ is compact with countably infinite eigenvalues almost surely, which means that the $j$-th largest eigenvalue $\Lambda_j(\sum_{i=k }^{k+h-1} \E[H_{i}|\F_{k-1}])$ is well-defined. The \textit{RKHS persistence of excitation} (\ref{suanzi}) in Definition \ref{pe} implies that
 $\inf_{k\in\mathbb N}\Lambda_j \Big(\sum_{i=k }^{k+h-1}\E [ H_{i} |\F_{k-1} ] \Big)>0~\mathrm{a.s.},~j=1,2,\cdots.$  \hfill  $\blacksquare$
}
\end{remark}

\vskip 0.1cm
\begin{remark}
\rm{For the finite-dimensional filtering, Guo~(\cite{guo1}) proposed a stochastic PE condition: there exists an integer $h>0$ and a constant
$\alpha>0$ such that $$
\inf_{k\in\mathbb N}\Lambda_{\mathrm{min}}\bigg(\E\bigg[
\sum_{i=k}^{k+h-1}
\frac{x_ix_i^{\top}}{1+ \|x_i \|^2} |\F_{k-1}\bigg]\bigg)
\geq \alpha~\mathrm{a.s.}
$$
Zhang et al. (\cite{Zhang2025})  generalized
Guo's stochastic PE condition to decentralized regularized linear regression problems with a regularization parameter $\lambda_k\geq 0$.
For the special case with a single node and  $\lambda_k=0$, the PE condition in \cite{Zhang2025} becomes: there exists an integer $h>0$ and a constant
$\alpha>0$ such that $$
\inf_{k\in\mathbb N}\Lambda_{\mathrm{min}}\bigg(\E\bigg[
\sum_{i=k}^{k+h-1}
x_ix_i^{\top} |\F_{k-1}\bigg]\bigg)
\geq \alpha~\mathrm{a.s.}
$$
Different from \cite{Zhang2025}, here,
the essential infinite-dimensional framework  requires that the regularization parameter $\lambda_k>0$.
For the finite-dimensional space $\HH_K=\mathbb R^n$, where $K(x,y)=\langle x,y \rangle =x^Ty$, $\forall ~x,y\in \X\subseteq \mathbb R^n$, the statistical learning model (\ref{model}) degenerates to a parameter estimation problem with the measurement model
$
y_k=x_k^{\top}\theta_0+v_k,~\forall ~k\in \mathbb N,
$
where $\theta_0\in\mathbb R^n$ is the unknown vector. And the \textit{RKHS persistence of excitation} (\ref{suanzi}) in Definition \ref{pe} degenerates to that there exists an integer $h>0$ and a random variable $\alpha>0$, such that
$
\inf_{k\in\mathbb N}\Lambda_{\mathrm{min}}\big(\E\big[ \sum_{i=k }^{k+h-1}x_ix_i^{\top} |\F_{k-1}\big]\big)>\alpha~\mathrm{a.s.},
$
which seems weaker than \cite{Zhang2025}. However, the nonzero regularization parameter $\lambda_k$ leads to an additional slowly time-varying condition on the random Tikhonov regularization path in the consistency analysis.

The RKHS PE condition is not a finite-dimensional Gram-matrix invertibility condition or a restricted-eigenvalue condition, which says that for deterministic input points $x_k,\ldots,x_{k+h-1}$, or after fixing one realization of the input data, the Gram matrix
$\mathbf K=(K(x_i,x_j))_{i,j=k}^{k+h-1}$
induced by   $\{K_{x_k},\ldots,K_{x_{k+h-1}}\}$ is invertible, and this invertibility condition is equivalent to
$$
\left\langle f,
\left(
\sum_{i=k}^{k+h-1}K_{x_i}\otimes K_{x_i}
\right)f
\right\rangle_{\mathcal H_K}
\geq c\|f\|_{\mathcal H_K}^2>0,\, \forall \ f\in S_k\setminus\{0\},
$$
 where $c=\lambda_{\min}(\mathbf K)>0$, and $S_k= \mathrm{span}\{K_{x_i}:i=k,\ldots,k+h-1\}$.
Indeed, such a Gram-matrix invertibility condition implies the strict positivity of the associated operator $\sum_{i=k}^{k+h-1}K_{x_i}\otimes K_{x_i}$ on the finite-dimensional subspace $S_k$ spanned by the observed feature maps (\cite{Lazar2024}), while Definition~\ref{pe} requires a conditional auto-covariance operator inequality on the entire RKHS $\HH_K$. More specifically, Definition~4.1 requires auto-covariance operator inequality
$$
\sum_{i=k}^{k+h-1}
\mathbb E\big[
K_{x_i}\otimes K_{x_i}\mid\mathcal F_{k-1}
\big]
\succeq R
\quad \text{a.s.},
$$
where $R$ is a strictly positive compact operator. Equivalently,
$$
\left\langle f,
\left(
\sum_{i=k}^{k+h-1}
\mathbb E\big[
K_{x_i}\otimes K_{x_i}\mid\mathcal F_{k-1}
\big]
\right)f
\right\rangle_{\mathcal H_K}
\geq
\langle f,Rf\rangle_{\mathcal H_K}>0,
\
\forall \ f\in\mathcal H_K\setminus\{0\}.
$$

Moreover, requiring that the eigenvalues of the Gram matrix admit a uniformly strictly positive lower bound is essentially a finite-dimensional condition. For an infinite-dimensional RKHS, even if the data-induced conditional auto-covariance operator is strictly positive on $\HH_K$, the infimum of its eigenvalues is zero.

The RKHS PE condition implies that  no nontrivial element $f\in\mathcal H_K$  lies in the null space of the conditional auto-covariance operator
$
\sum_{i=k}^{k+h-1}\mathbb E [H_i|\mathcal F_{k-1}],
$
which ensures that every nontrivial element in the RKHS receives a nonvanishing amount of excitation in the conditional second-moment sense over each time window.
Besides, the condition  implies  that the only function that can be orthogonal to all feature maps $\{K_{x_i}, i=k,\dots,k+h-1\}$ over any window is the trivial element, which guarantees that the directions along which the  algorithm updates are sufficiently rich, so that the estimator does not collapse onto an unidentifiable subspace.
Thus, this condition can be viewed as a function-space analogue of the classical PE condition in \cite{Green} in adaptive estimation and control.
\hfill  $\blacksquare$
}
\end{remark}
\vskip 0.1cm

\begin{remark}
\rm{
Zhang and Li (\cite{ZXWLT}) studied the online learning theory with non-i.i.d. data in RKHS, and proposed a  PE condition, that is,  the auto-covariance operators of the input data over a fixed length time period have a strictly positive compact lower bound $R\in\mathscr L(\HH_K)$, i.e.
$
\sum_{i=k}^{k+h-1}\E\left[H_{i}\right]\succeq R,~\forall~k\in\mathbb N,
$
and
$
\lim_{i\to\infty}\sup_{\substack{u_i\in \F_{i-1},  \|u_i\|_{\HH_K}=1}}\E\big[  \|  ( \E\left[H_{i}\right]\hspace{-2pt}-\hspace{-2pt} T_{i} )u_i \|^2_{\HH_K}\big]^{\frac{1}{2}} =0.
$
Different from the PE condition in \cite{ZXWLT}, the \textit{RKHS persistence of excitation} condition is imposed  directly on the conditional auto-covariance operators and
 no longer requires the above convergence.
\hfill  $\blacksquare$
}
\end{remark}

\vskip 0.2cm

\begin{remark}
\rm{
Choosing appropriate algorithm  gains and regularization parameters is crucial for the consistency of the  algorithm.
We select a decaying gain sequence $a_k$ in Condition \ref{con1} to attenuate the effect of noise, and a decaying regularization parameter $\lambda_k$ to ensure that the random Tikhonov regularization path $f_{\lambda,k}$ approximates $f^{\star}$.
Moreover, Condition \ref{con1} guarantees that the influence of the initial value vanishes, as   $a_k\lambda_k$ satisfies $\sum_{k=0}^{\infty} a_k\lambda_k = \infty$.
To control the random fluctuations induced by sampling in the regularization path, we choose $a_k = (k+1)^{-\tau_1}$  decaying faster than $\lambda_k = (k+1)^{-\tau_2}$ with $3\tau_2 < \tau_1$.
It follows from Lemma \ref{THMM} and condition (\ref{cnm}) that if the drift $\|f_{\lambda,k+1}-f_{\lambda,k}\|_{L^2(\Omega;\HH_K)}$ decays faster than $a_k\lambda_k$, then the mean square error between $f_k$ and $f_{\lambda,k}$ converges to zero.
Finally, the \textit{RKHS persistence of excitation} condition ensures that $f_{\lambda,k}$ converges to $f^{\star}$ in mean square, which gives the mean square consistency of the algorithm (\ref{AL}). For choosing   algorithm  gains and regularization parameters satisfying Condition   \ref{con1}, one may first choose $\tau_1\in (0,1)$ and then choose $\tau_2\in \big(0,\min\{1-\tau_1,\ \tfrac{1}{3}\tau_1\}\big).$
\hfill  $\blacksquare$
}
\end{remark}
\vskip 0.2cm
\section{Special case with independent and non-identically distributed online data streams}
In this section, we will  consider the special case with independent and non-identically distributed online data streams. Let the input space
 $\X$ be a compact set in $\mathbb R^n$. It follows from Riesz representation theorem that $\mathcal M(\X)$ is the dual of the Banach space $(C(\X),\|\cdot\|_{\infty})$ consisting of all continuous functions defined on  $\X$ (\cite{Lax}), i.e. $\mathcal M(\X)=(C(\X))^*$. Denote the probability distribution of the observation data $(x_k,y_k)$ at instant $k$ as $\rho^{(k)}$, and $\rho_{\X}^{(k)}$ is the marginal probability measure induced by the input data $x_k$. For the independent data streams $\{(x_k,y_k),k\in \mathbb N\}$, we have the following proposition.
 \vskip 0.1cm

\begin{proposition}\label{mingti7}
\rm{Suppose that the online data streams $\{(x_k,y_k),k\in \mathbb N\}$ are mutually independent. If there exists an integer $h>0$ and a strictly positive measure $\gamma\in \mathcal M_{+}(\X)$, such that
\bna\label{cedu}
\frac{1}{h}\sum_{i=k }^{k+h-1}\rho^{(i)}_{\X}\ge \gamma,~\forall ~k\in \mathbb N,
\ena
then $\{(x_k,y_k),k\in \mathbb N\}$ satisfies the RKHS persistence of excitation condition.
}
\end{proposition}
\vskip 0.2cm
\begin{proof}
  See Appendix \ref{fulu1ccc} for the proof.
\end{proof}
\begin{remark}
\rm{For the RKHS persistence of excitation condition (\ref{suanzi}), we do not require the online data streams to be independent or stationary. Proposition \ref{mingti7} specifically characterizes  the  condition (\ref{suanzi}) using the probability measures of the dataset for the case of independent data streams, where the average $h^{-1}\sum_{i=k}^{k+h-1}\rho^{(i)}_{\X}$ of the marginal probability measures over each time interval of length $h$ has a uniformly strictly positive lower bound $\gamma\in\mathcal M_{+}(\X)$. Intuitively, if there exists an open set $U$ in $\X$, such that $\rho^{(k)}_{\X}(U)=0$, $\forall~ k\in \mathbb N$, then we cannot obtain any information about  $f^{\star}$ on $U$, which shows that the condition (\ref{cedu}) is necessary for the consistency of the algorithm (\ref{AL}) in some sense. Furthermore, we do not require each marginal measure at each time instant to be strictly positive. Instead, it suffices to require the averages of all marginal measures within the time interval $[k,k+h-1]$ to be strictly positive. Notably,  (\ref{cedu}) degenerates to the condition in \cite{gzc}, that is, $\gamma=\rho^{(0)}_{\X}$ is a strictly positive probability measure, for the case with i.i.d. online data streams.\hfill  $\blacksquare$
}
\end{remark}
\vskip 0.2cm

Denote the H\"{o}lder space by $C^s(\X)=\{f\in C(\X): \|f\|_{C^s(\X)}<\infty\}$, where  $0\leq s\leq 1$,  $\|f\|_{C^s(\X)}=\|f\|_{\infty}+|f|_{C^s(\X)}$, $\|f\|_{\infty}=\sup_{x\in \X}|f(x)|$, and
\[|f|_{C^s(\X)}=\sup_{x\neq y, \ x,\ y \in \X}\frac{|f(x)-f(y)|}{\|x-y\|^s}.\]
Here, $C^s(\X)$ is a Banach space (\cite{Lax}). If the sample space of the probability measure $\rho$ is $\X$, then $\rho$ is a bounded linear functional on $C^s(\X)$ (\cite{Lax}), i.e. $\rho\in (C^s(\X))^*$.
\begin{assumption}\label{ass3}
\rm{
There exist constants $0\leq s\leq 1$ and $\tau_s>0$, such that the kernel function $K\in C^s(\X\times \X)$, and for any $u_1,u_2,v_1,v_2\in\X$,
\ban
\left|K(u_1,v_1)-K(u_2,v_1)-K(u_1,v_2)+K(u_2,v_2)\right|\leq \tau_s\|u_1-u_2\|^s\|v_1-v_2\|^s.
\ean
}
\end{assumption}
\begin{remark}
\rm{In the works of online regularized learning algorithms based on i.i.d. data   (\cite{Smale5}-\cite{Hu}), Assumption \ref{ass3} is referred to as the $s$-order kernel condition. Specifically, if $K\in C^2(\X\times\X)$ and $\X$ is a smooth and bounded region in $\mathbb R^n$, then Assumption \ref{ass3} holds (\cite{dxzhou}).\hfill $\square$
}

\end{remark}

Combining Proposition \ref{mingti7} and Assumption \ref{ass3}, the following corollary provides  sufficient conditions for the consistency of the online regularized learning algorithm (\ref{AL}) by characterizing the marginal probability measure $\rho^{(k)}_{\X}$ induced by the random input data.
\vskip 1mm
\begin{corollary}\label{corollary1}
\rm{For the algorithm (\ref{AL}), suppose that (i) Assumption  \ref{ass1}, Assumption \ref{ass3} and Condition \ref{con1} hold; (ii) the online data streams $\{(x_k,y_k),k\in \mathbb N\}$ are mutually independent,  and there exists an integer $h>0$ and a strictly positive  measure $\gamma\in \mathcal M_{+}(\X)$, such that
\bna\label{cedu111}
\frac{1}{h}\sum_{i=k }^{k+h-1}\rho^{(i)}_{\X}\ge \gamma,~\forall ~k\in \mathbb N;
\ena
(iii)
\bna\label{over1}
\left\|\rho_{\X}^{(k+1)}-\rho_{\X}^{(k)}\right\|_{\left(C^s(\X)\right)^*}=O\left(a_k\lambda^2_k\right).
\ena
Then
$\lim_{k\to\infty}\|f_k-f^{\star}\|^2_{L^2(\Omega;\HH_K)}=0$, $\lim_{k\to\infty}\mathbb{E}\big[|f_k(x)-f^{\star}(x)|^2\big]=0, \ \forall \ x\in \X$, $\lim_{k\to\infty}\|f_k-f^{\star}\|_{\HH_K}=0 \ \text{a.s.}$ and $\lim_{k\to\infty} f_k(x) =f^{\star}(x) \text{ a.s. } \ \forall \ x\in \X$.}
\end{corollary}
\begin{proof}
  See Appendix \ref{fulu1ccc} for the proof.
\end{proof}

If $\mathscr{X}\subseteq \mathbb{R}^n$ and the independent data stream ${(x_k,y_k),k\in\mathbb{N}}$ satisfy the RKHS persistence of excitation condition restricted on a compact subset, then the following corollary shows that the output of the algorithm (\ref{AL})  converges to $f^*$ pointwisely on the subset.

\vskip 1mm
\begin{corollary}\label{zasihenghexianz}
\rm{Let $\mathscr{X} \subseteq \mathbb{R}^n $ and $\hat{\mathscr{X}}  \subseteq \mathscr{X} $ be a nonempty compact set. For the algorithm (\ref{AL}), suppose that (i) Assumption  \ref{ass1}, Assumption \ref{ass3} and Condition \ref{con1} hold; (ii) the online data streams $\{(x_k,y_k),k\in \mathbb N\}$ are mutually independent,  and there exists an integer $h>0$ and a strictly positive  measure $\gamma\in \mathcal M_{+}(\hat{\mathscr{X}})$, such that
\bna\label{cedu111111}
\frac{1}{h}\sum_{i=k }^{k+h-1}\rho^{(i)}_{\hat{\mathscr{X}}}\ge \gamma,~\forall ~k\in \mathbb N;
\ena
(iii)
\bna\label{over111}
 \|\rho_{\hat{\mathscr{X}}}^{(k+1)}-\rho_{\hat{\mathscr{X}}}^{(k)} \|_{\left(C^s(\hat{\mathscr{X}} )\right)^*}=O\left(a_k\lambda^2_k\right).
\ena
Then
$\lim_{k\to\infty}\|f_k-f^{\star}\|^2_{L^2(\Omega;\hat{\mathscr{H}}_{K})}=0$, $\lim_{k\to\infty}\mathbb{E}\big[|f_k(x)-f^{\star}(x)|^2\big]=0, \ \forall \ x\in  \hat{\mathscr{X}}$, $\lim_{k\to\infty}\|f_k-f^{\star}\|_{\hat{\mathscr{H}}_{K}}=0 \ \text{a.s.}$ and $\lim_{k\to\infty} f_k(x)=f^{\star}(x),\ \text{ a.s. } \ \forall \ x\in  \hat{\mathscr{X}}$}, where $\hat{\mathscr{H}}_{K} = \{ f_{1} \mid f_{1} = f|_{\hat{\mathscr{X}}},\ f \in \mathscr{H}_{K} \}$ and  $f|_{\hat{\mathscr{X}}}$ is the restriction  of  $f$ to   $\hat{\mathscr{X}}$.
\end{corollary}
\begin{proof}
  See Appendix \ref{fulu1ccc} for the proof.
\end{proof}

\begin{remark}
\rm{Compared with the online learning algorithms with i.i.d. data streams, the consistency of online algorithms with independent but non-stationary data depends on the sequence of marginal probability measures  $\{\rho^{(k)}_{\X},k\in \mathbb N\}$. To analyze the algorithm (\ref{AL}) with the above settings, Smale and Zhou (\cite{Smale5}) established the  exponential convergence condition of the sequence of marginal probability measures in $(C^s(\X))^*$, i.e. there exists a probability measure $\rho_{\X}$ on $\X$, and constants $C_1>0$, $0<\alpha<1$, such that
\bna\label{bj1}
 \|\rho_{\X}^{(k)}-\rho_{\X} \|_{(C^s(\X))^*}\leq C_1\alpha^k,~\forall~ k\in \mathbb N,
\ena
then the algorithm (\ref{AL}) is consistent in mean square. Subsequently, Hu and Zhou (\cite{Hu}) investigated the consistency of the online regularized algorithms with general loss functions and weakened the above condition (\ref{bj1}) to the polynomial convergence of the sequence of marginal probability measures in $(C^s(\X))^*$, i.e. there exists a probability measure $\rho_{\X}$ on $\X$, and constants $C_2>0$, $b>1$, such that
\bna\label{bj2}
 \|\rho_{\X}^{(k)}-\rho_{\X} \|_{(C^s(\X))^*}\leq C_2k^{-b},~\forall ~k\in \mathbb N.
\ena
Compared to the restrictions in \cite{Smale5}-\cite{Hu} on the sequence of marginal probability measures, which are required to converge  to a limiting probability measure in $(C^s(\X))^*$, in the condition (\ref{over1}) of Corollary \ref{corollary1}, we no longer require  the convergence of marginal probability measures, instead of which, we only require the drifts of marginal probability measures $\rho^{(k)}_{\X}$ to be of $O(a_k\lambda^2_k)$. In particular, if the algorithm gains and regularization parameters are chosen as $a_k=(k+1)^{-0.7}$ and $\lambda_k=(k+1)^{-0.15}$, it can be verified that Condition \ref{con1} holds and $a_k\lambda_k^2=(k+1)^{-1}$. Furthermore, if the marginal probability measures satisfy (\ref{bj2}), then
$
\big\|\rho_{\X}^{(k+1)}-\rho_{\X}^{(k)}\big\|_{\left(C^s(\X)\right)^*} \leq 2C_2k^{-b}.
$
Note that $b>1$, which shows that  (\ref{over1}) in Corollary \ref{corollary1} is satisfied. Therefore, (\ref{bj1})-(\ref{bj2}) are both sufficient conditions for (\ref{over1}). On the other hand, to ensure the consistency of the online regularized algorithm,  Smale and Zhou (\cite{Smale5}), Hu and Zhou (\cite{Hu}) both required the regression function to satisfy the regularity condition involving the limiting probability measure $\rho_{\X}$. Different from this, the condition (\ref{cedu111}) in Corollary \ref{corollary1} does not require any prior information about the unknown function and only requires that the average $h^{-1}\sum_{i=k }^{k+h-1}\rho^{(i)}_{\X}$ of marginal probability measures has a uniformly strictly positive lower bound $\gamma\in\mathcal M_{+}(\X)$ within each time interval of length $h$. In summary, even for the independent and non-identically distributed online data streams, we have obtained more general results.\hfill  $\blacksquare$
}
\end{remark}

\section{Numerical examples}
\subsection{One-dimensional Regression Tasks}
Let $\X=[-1,5]$. The observation data $(x_k,y_k)$ at instant $k$ satisfies $y_k = f^{\star}(x_k) + v_k$, where   $f^{\star}(x), ~\forall~ x \in \X$  is the true  function to be estimated.
	Take the  Gaussian kernel $K(x,y) = e^{-(x-y)^2}, ~\forall ~x,\ y \in \X$. It can be verified that Assumption \ref{ass3} holds with $s = 1$.
We use the  algorithm (\ref{AL}) to estimate $f^{\star}$. Let the initial value  $f_0 = 0$.
We sample $1000$ points $\{z_{l},\ l=1,\ldots,1000\}$ on $\X$ with $z_{l}=-1+\frac{6(l-1)}{1000}, \ l=1,\ldots,1000.$ Then, we   iterate the values of $f_k$ at the sampled points by algorithm (\ref{AL}), that is,
\begin{align}
f_{k+1}(z_{l})=f_k(z_{l})-a_k\left((f_k(x_k)-y_k)K(x_k,z_{l})
+\lambda_kf_k(z_{l})\right),\notag
\end{align}
$~\forall~k\in \mathbb N,\ l=1,\ldots,1000.$ If $x_k \notin \{z_{l},\ l=1,\ldots,1000\}$, we approximate $f_k(x_k)$ by the cubic spline interpolation method.

\subsubsection{independent and non-identically distributed online data streams}

The input data $\{x_k,\ k \in \mathbb{N}\}$ are  independent random variables, each of which is with the uniform distribution on $I_k$, where $I_k = \X$ for $k = 0$, and  $I_k = \big[ \frac{3(1+(-1)^k)}{k+1} - 1,\; \frac{3(1+(-1)^k)}{k+1} - \frac{6}{k+1} + 5 \big]$  for $k = 1, 2, \ldots$
  Denote the probability density function of $x_k$ by $m_k$.
  We have $m_0(x)=\frac{1}{6}$, $\forall~x\in\X$, and
	$$ m_k(x) = \displaystyle  \frac{k+1}{6k}, ~\forall~x\in I_k,~k=1,2,\cdots$$
	Denote $S_{2,k}(x) = m_k(x) +m_{k+1}(x)$.
	For the odd integer $k$, we obtain
	\begin{equation}\label{gaosifenbu1}
		S_{2,k}(x)  = \left\{
		\begin{aligned}
			&m_k(x)  &&=\displaystyle\frac{k+1}{6k}, && -1\leq x\leq -1+\frac{6}{k+2};\\
			&m_k(x) +m_{k+1}(x)  &&= \frac{k+1}{6k}+\frac{k+2}{6(k+1)}, && -1+\frac{6}{k+2} < x < 5-\frac{6}{k+1};\\
			&m_{k+1}(x)  &&= \frac{k+2}{6(k+1)}, && 5-\frac{6}{k+1} \leq x \leq 5.
		\end{aligned}
		\right.
	\end{equation}
	For the even integer $k>0$, we get
	\begin{equation}\label{gaosifenbu2}
		S_{2,k}(x)= \left\{
		\begin{aligned}
			&m_{k+1}(x)  &&=\displaystyle\frac{k+2}{6(k+1)}, && -1\leq x \leq -1+\frac{6}{k+1};\\
			&m_k(x) +m_{k+1}(x)  &&= \frac{k+1}{6k}+\frac{k+2}{6(k+1)}, && -1+\frac{6}{k+1} < x < 5-\frac{6}{k+2};\\
			&m_{k}(x)  &&= \frac{k+1}{6k}, && 5-\frac{6}{k+2} \leq x \leq 5.
		\end{aligned}
		\right.
	\end{equation}
Noting that $S_{2,0}(x)>\frac{1}{6}$, it follows from (\ref{gaosifenbu1})-(\ref{gaosifenbu2}) that
	$$ S_{2,k}(x) \geq \frac{1}{6},~\forall ~x \in \X,~\forall ~k \in \mathbb{N}.$$
	For any open set $(a,b) \subset [-1,5]$, we have
	$$\frac{1}{2}\left(\rho_{\mathscr{X}}^{(k)}+ \rho_{\mathscr{X}}^{(k+1)}\right)(a,b) =\frac{1}{2} \int_{a}^{b}\left(m_k(x)+m_{k+1}(x)\right)~\mathrm{d}x \geq \frac{1}{2}(b-a) = \gamma(a,b),~\forall~k\in\mathbb N,$$
where $\gamma:= \frac{1}{2} \gamma_1$ and $\gamma_1$ is the Lebesgue measure defined on $\X$. Noting that $\gamma\in \mathcal{M}_{+}(\X)$ and $\gamma$ is strictly positive, we know that
$$\frac{1}{2}\left(\rho_{\mathscr{X}}^{(k)}+ \rho_{\mathscr{X}}^{(k+1)}\right)\ge  \gamma,~\forall~k\in\mathbb N,$$
which implies that the condition (\ref{cedu111}) in Corollary \ref{corollary1} holds with $h=2$.
For the odd integer $k$, note that
	\begin{equation*}
		m_{k+1}(x)-m_{k}(x) = \left \{
		\begin{aligned}
			&-\frac{k+1}{6k}, && -1\leq x\leq -1+\frac{6}{k+2};\\
			&\frac{-1}{6k(k+1)}, && -1+\frac{6}{k+2} < x < 5-\frac{6}{k+1};\\
			&\frac{k+2}{6(k+1)}, && 5-\frac{6}{k+1} \leq x \leq 5.
		\end{aligned}
		\right.
	\end{equation*}
Moreover, for any $g\in C^1(\mathscr{X})$ satisfying $\| g \|_{C^1(\mathscr{X})} \leq 1$, we have $\| g \|_{\infty} \leq  1.$ Therefore,
\begin{align}
&\left\| \rho_{\mathscr{X}}^{(k+1)}-\rho_{\mathscr{X}}^{(k)} \right\| _{(C^1(\mathscr{X}))^*} \notag\\
=& \sup_{g\in    C^1(\mathscr{X}),\ \| g \|_{C^1(\mathscr{X})} \leq 1}{\left \| \int_{\X}g~\mathrm{d}\left(\rho_{\mathscr{X}}^{(k+1)}-\rho_{\mathscr{X}}^{(k)}\right) \right \| }\cr
 \leq & \sup_{g\in     C^1(\mathscr{X}),\ \| g \|_{\infty} \leq 1}{\left \| \int_{\X}g~\mathrm{d}\left(\rho_{\mathscr{X}}^{(k+1)}-\rho_{\mathscr{X}}^{(k)}\right) \right \| }\cr
= & \sup_{g\in     C^1(\mathscr{X}),\ \| g \|_{\infty} \leq 1}{\left \| \int_{-1}^{5}{g(x)\left(m_{k+1}(x)-m_{k}(x)\right)}~\mathrm{d}x \right \|}\cr
 \leq & \int_{-1}^5\left|m_{k+1}(x)-m_{k}(x)\right|~\mathrm{d}x\cr
 \leq & \frac{3}{k}.\notag
\end{align}
By the above inequality, we have $\| \rho_{\mathscr{X}}^{(k+1)}-\rho_{\mathscr{X}}^{(k)} \|_{(C^1(\mathscr{X}))^*}=O(k^{-1})$ for the odd integer $k$, which also holds for the even integer $k$ by the same way. Thus,  we obtain
\ban
\left\|\rho_{\X}^{(k+1)}-\rho_{\X}^{(k)}\right\|_{(C^1(\mathscr{X}))^*}
=O\left(a_k\lambda^2_k\right),
\ean
which implies that the condition (\ref{over1}) in Corollary \ref{corollary1} holds.
The measurement noises $\{v_k,k=0,1,...\}$ are independent random variables with the normal distribution $N(0, 0.1)$ and independent of the input data $\{x_k,k=0,1,...\}$.  It follows from Remark \ref{fangzhenzhu} that Assumption \ref{ass1} holds.

\begin{figure}
    \centering
     \subfigure{\includegraphics[scale=0.3]{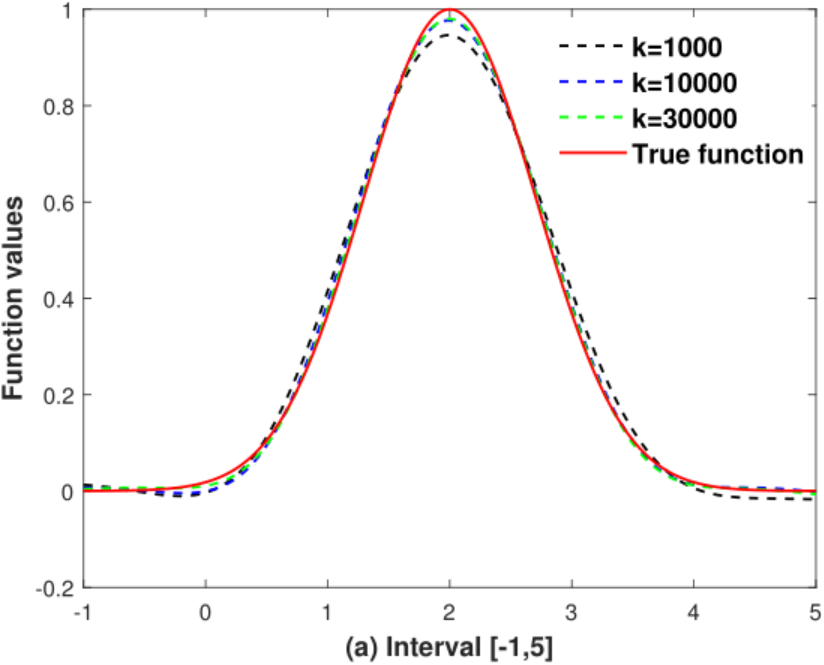}}
     	\subfigure{\includegraphics[scale=0.33]{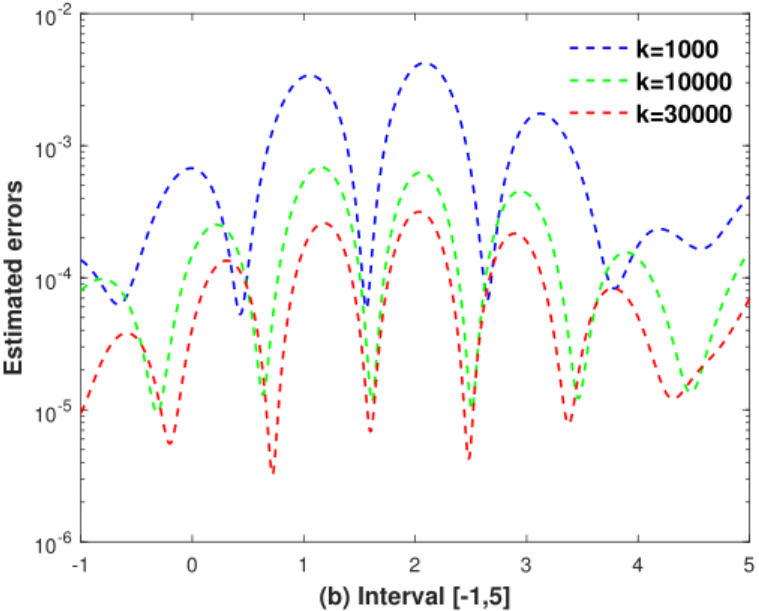}}
\caption{ (a) Approximated functions and true function ; (b) Mean square  errors of approximated functions.}\label{peconditions}
	\end{figure}

\begin{figure}
    \centering
     \subfigure{\includegraphics[scale=0.27]{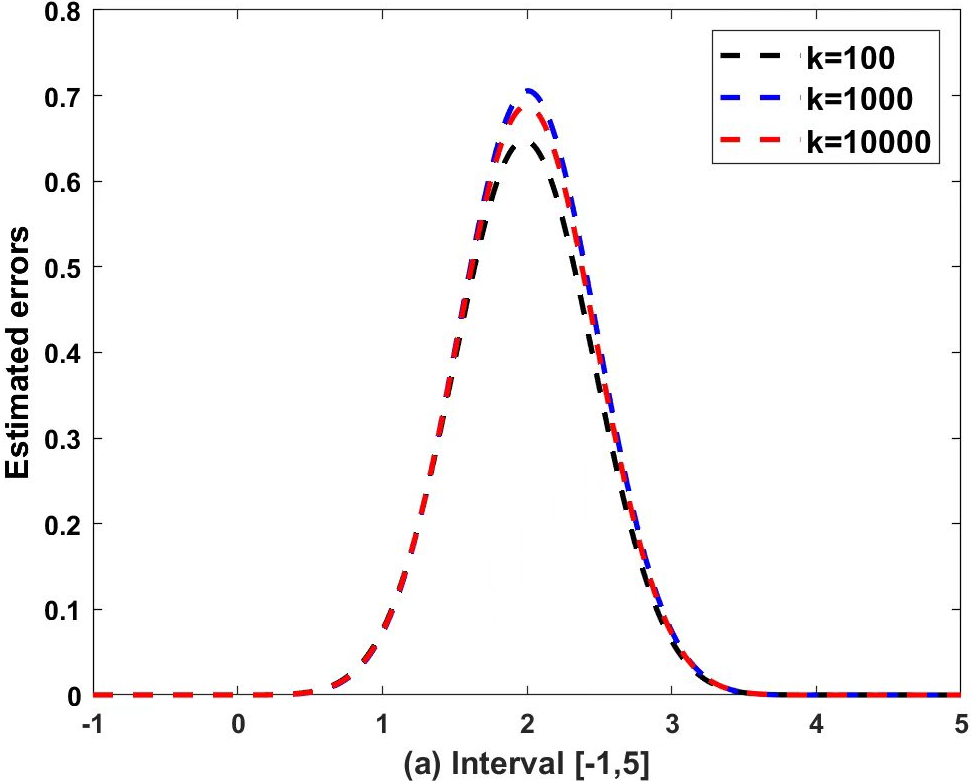}}
	\subfigure{\includegraphics[scale=0.35]{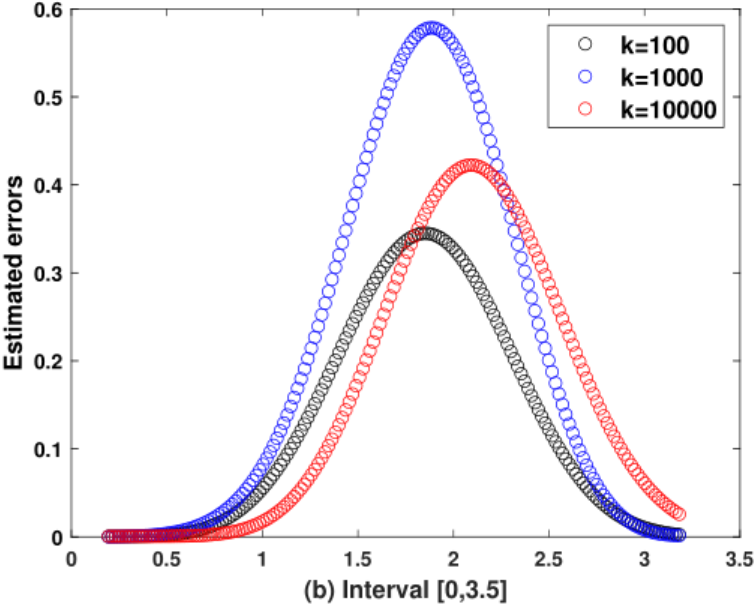}}

\caption{ (a) Mean square  errors of KLMS; (b) Mean square  errors of NORMA.}	
\label{fig2}
	\end{figure}

First, we examine the convergence of the algorithm (\ref{AL}) with the algorithm gain $a_k =  (k+1)^{-0.7} $ and the regularization parameter $\lambda_k =  10^{-4} (k+1)^{-0.15}$. Let  $f^{\star}(x)=e^{-(x-2)^2}, ~\forall~ x \in \X$ and note that $f^{\star}\in \HH_K.$ Fig.\ref{peconditions}(a)  shows
  $f_k(x,\omega_{1})$, $x \in \X$ at iterations $k=1000,\ 10000,\ 30000.$
Fig.\ref{peconditions}(b) shows the graph  of  $\mathbb{E}\left[|f_k(x)-f^{\star}(x)|^2\right],\  x\in \X$ for $k=1000,\ 10000,\ 30000$.  Here, $\mathbb{E}\big[|f_k(x)-f^{\star}(x)|^2\big],\  x\in \X$
 is approximated by $\frac{1}{100} \sum\limits_{i=1}^{100}|f_k(z_{l},\omega_{i})-f^{\star}(z_{l})|^2,\ l=1,\ldots,1000$, where $\omega_{i}$ is the $i$-th sample path.  It can be seen that the algorithm (\ref{AL}) converges almost surely and  in mean square,  which is consistent with the convergence result of Corollary \ref{corollary1}.
 We also implement  KLMS  and NORMA in  \cite{Kivinen04} and the results are shown in Fig.\ref{fig2}(a) and (b). The results indicate that  $\mathbb{E}\left[|f_k(x)-f^{\star}(x)|^2\right], x\in \X$ obtained by both algorithms does not converge to 0 as the number of iterations increases. In contrast, $\mathbb{E}\left[|f_k(x)-f^{\star}(x)|^2\right],\ x\in \X$ obtained by our algorithm does converge to zero.

\begin{figure}
    \centering
     \subfigure{\includegraphics[scale=0.15]{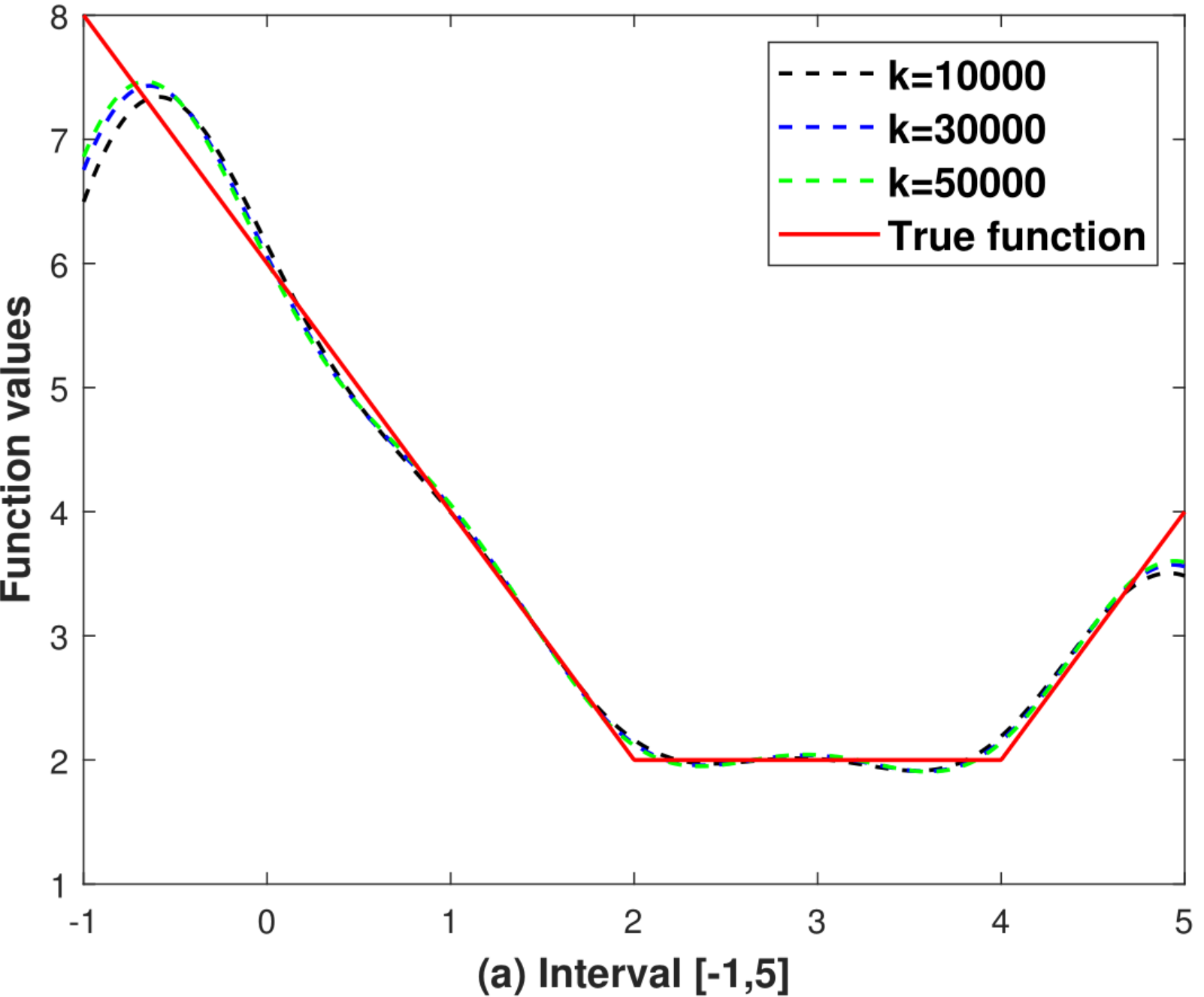}}
	\subfigure{\includegraphics[scale=0.15]{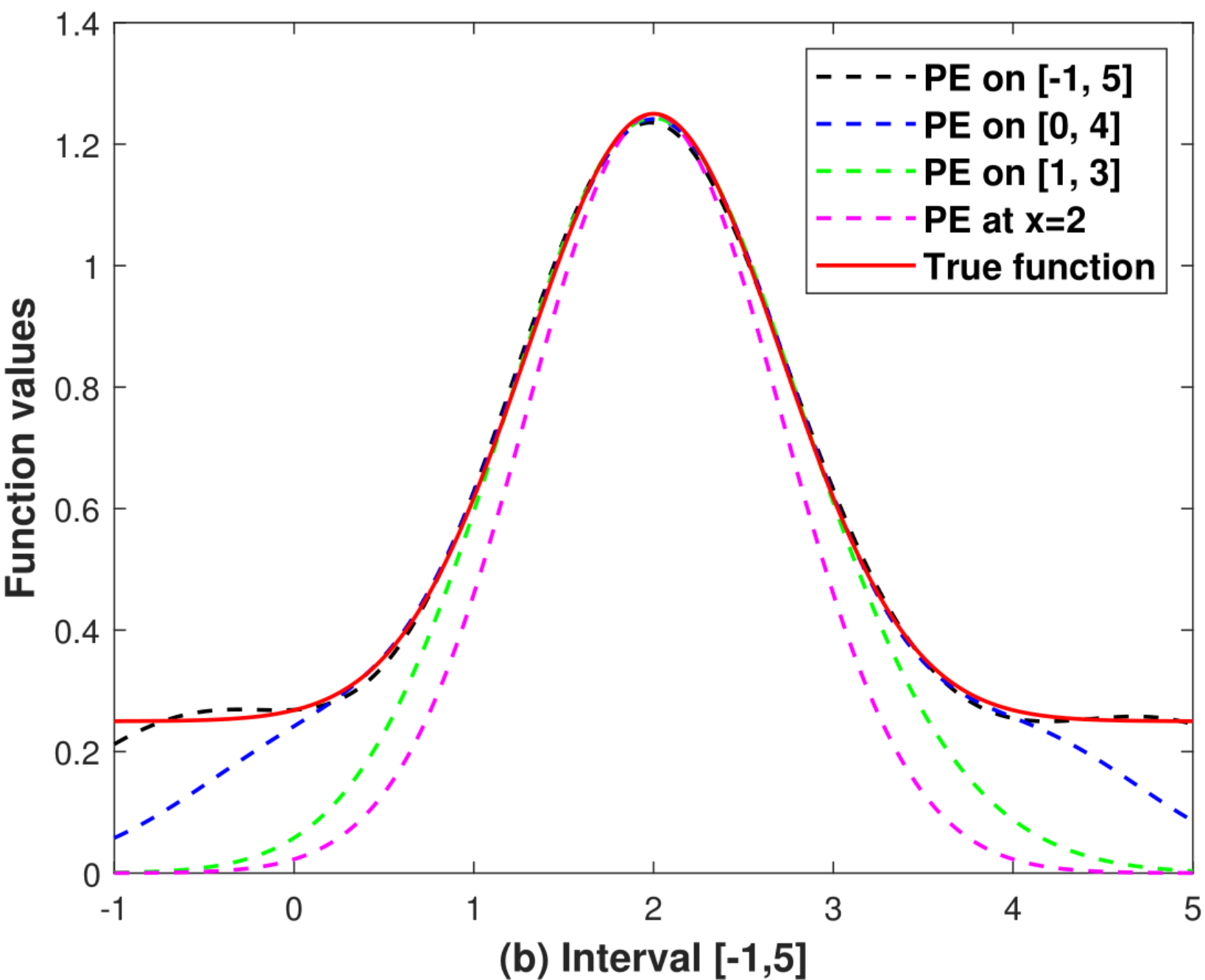}}

\caption{ (a) Approximated functions and true function ($f^{\star} \notin \HH_K$); (b) Approximated functions under different excitation intervals .}	
\label{fig2000}
	\end{figure}

We now consider the case where the unknown function does not belong to $\HH_K$. Specifically, let $f^{\star}(x)=|x-2|+|x-4|,\ x \in \X$. Since the RKHS $\HH_K$ is induced by the Gaussian kernel consists of analytic functions, whereas $f^{\star}$ is not differentiable at $x=2,\ 4$, it follows that $f^{\star} \notin \HH_K$.
Fig.\ref{fig2000}(a) illustrates the realizations of $f_k(x,\omega_{1})$, $x \in \X$ at iterations $k= 10000,\ 30000,$ and $50000$. The algorithm is implemented with  $a_k=\frac{1}{(k+1)^{0.7}}$, $\lambda_k=\frac{10^{-4}}{(k+1)^{0.15}}$.
The results show that the output of the algorithm (\ref{AL}) fails to converge to $f^{\star}$. Instead, a non-vanishing asymptotic error remains, which can be attributed to the model mismatch between the RKHS   and   $f^{\star}$.  Nevertheless, the algorithm can still identify the best approximation to $f^{\star}$ within the given RKHS.

To show the algorithm's sensitivity to the RKHS PE condition, we consider four types of input data streams and investigate the convergence behavior of the estimated function $f_k$ when the RKHS PE condition is satisfied only within restricted subregions $\Omega_s\subseteq \mathscr{X},\ s=1,2,3,4$.
 The input data $\{x_k^{(s)},\ k \in \mathbb{N}\}$ are  independent random variables, each of which is with the uniform distribution on $I_k^{(s)}$, where
$
I_k^{(s)} =  [ \frac{(4-s)(1+(-1)^k)}{k+1} + s-2, \; \frac{(4-s)(1+(-1)^k)}{k+1} - \frac{2(4-s)}{k+1} + 6-s],
$
$s=1,2,3,4.$ Therefore, the RKHS PE condition is satisfied over four subintervals $\Omega_1 = [-1, 5]$, $\Omega_2 = [0, 4]$, $\Omega_3 = [1, 3]$, and $\Omega_4 = \{2\}$. The realizations of the estimated functions $f_k(x, \omega_{1})$ for the four cases at $k=30,000$ iterations are illustrated in Fig.\ref{fig2000}(b). The results demonstrate that the algorithm effectively approximates the true function $f^*(x) = \frac{1}{4} + \exp(-(x-2)^2)$  exclusively within the   subintervals where the RKHS PE  condition is  satisfied.
In the  subintervals   where the PE condition fails, the estimated functions remain  near their initial values or exhibit large approximation errors due to the lack of informative updates.

\subsubsection{ M-dependent online data streams} 
\begin{figure}
    \centering
     \subfigure{\includegraphics[scale=0.25]{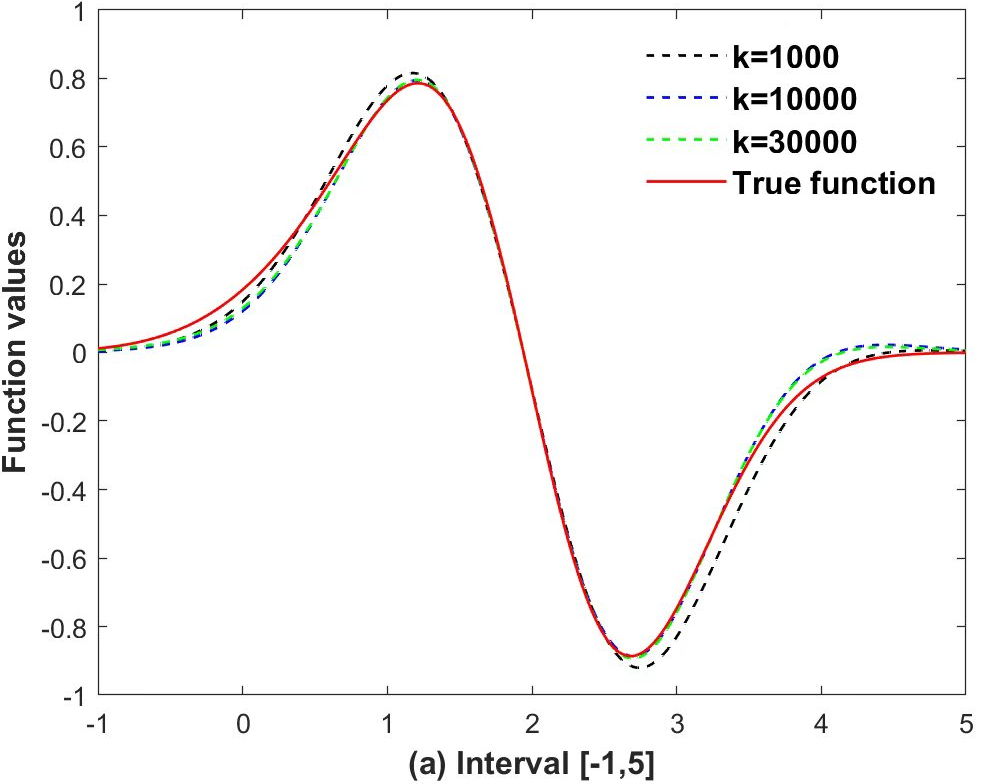}}
	\subfigure{\includegraphics[scale=0.25]{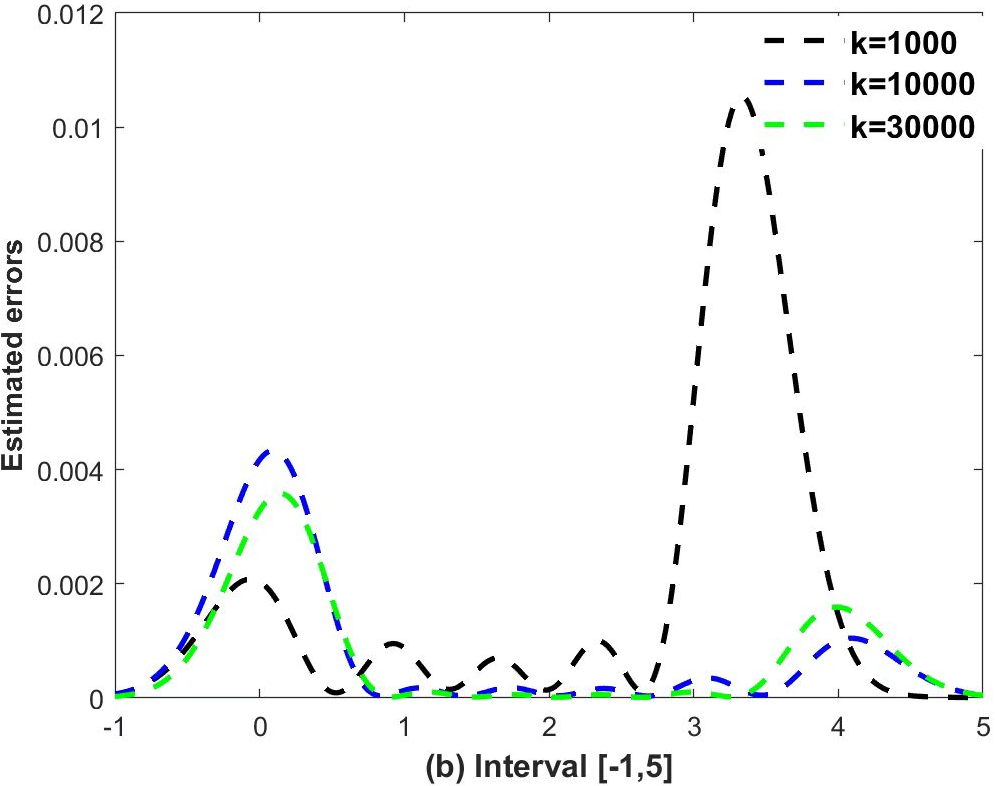}}

\caption{$M$-dependent online data streams: (a) Approximated functions and true function; (b) Mean square errors of approximated functions.}	
\label{fig20001}
	\end{figure}

The input data $\{x_k,\ k \in \mathbb{N}\}$ form an $m$-dependent sequence. Specifically, let $\{u_k,\ k \in \mathbb{N}\}$ be a sequence of independent random variables, each of which is uniformly distributed on $[-1,5]$. Then the input  data $\{x_k,\ k \in \mathbb{N}\}$ is given by
$
x_k = \frac{1}{m} \sum_{j=0}^{m-1} u_{k+j},\ k \in \mathbb{N}, m=10.
$
It follows that $\{x_k,\ k \in \mathbb{N}\}$ is  a  10-dependent sequence, each of which has the same distribution  $\gamma_1$, where $\gamma_1$  is  the distribution of the average of ten independent uniform random variables on $[-1,5]$. As $\gamma_1$ is a strictly positive measure on $[-1,5]$, then $E[H_i]=\int_{\X}K_x\otimes K_x\,\mathrm{d}\gamma_1$ is a strictly positive compact  operator, whose proof is similar to   Proposition \ref{mingti7}. Then, by Proposition  \ref{mixing}, the RKHS persistence of excitation condition is satisfied with $h= 11$ and   $R=\frac{1}{2}\int_{\X}K_x\otimes K_x\,\mathrm{d}\gamma_1$.
The measurement noises $\{v_k,\ k \in \mathbb{N}\}$ are independent random variables with the normal distribution $N(0, 0.1)$, and are independent of the input data $\{x_k,\ k \in \mathbb{N}\}$. Let  $f^{\star}(x)=-0.3(x-1)e^{-(x-1)^2}-2(x-2)e^{-(x-2)^2}, ~\forall~ x \in \X$ and note that $f^{\star}\in \HH_K.$ Fig.\ref{fig20001}(a) and (b) illustrate the   estimates $f_k(x,\omega_{1})$, $x \in \X$, together with the mean square errors $\mathbb{E}\big[|f_k(x)-f^{\star}(x)|^2\big]$ at iterations $k=10000,\ 30000,\ 50000$. The results demonstrate that the algorithm \eqref{AL} converges both almost surely and in  mean square.

\subsection{Two-dimensional Regression Tasks}
\begin{figure}[htbp]
    \centering
     \subfigure{\includegraphics[scale=0.28]{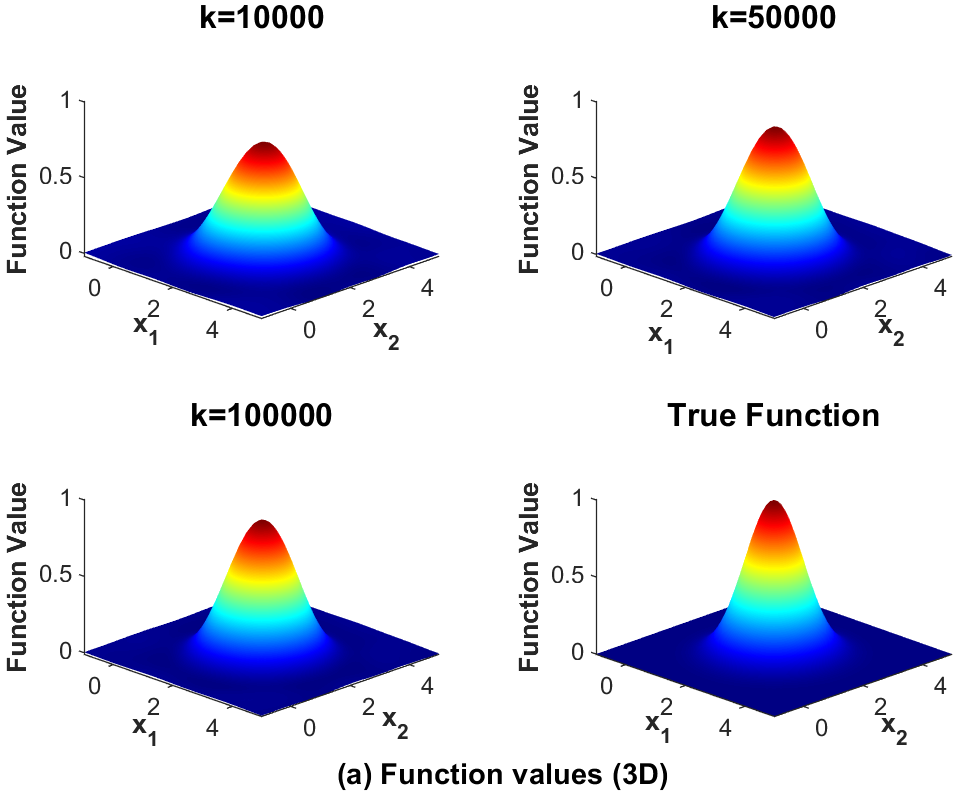}}
     \hspace{-1mm}
	\subfigure{\includegraphics[scale=0.28]{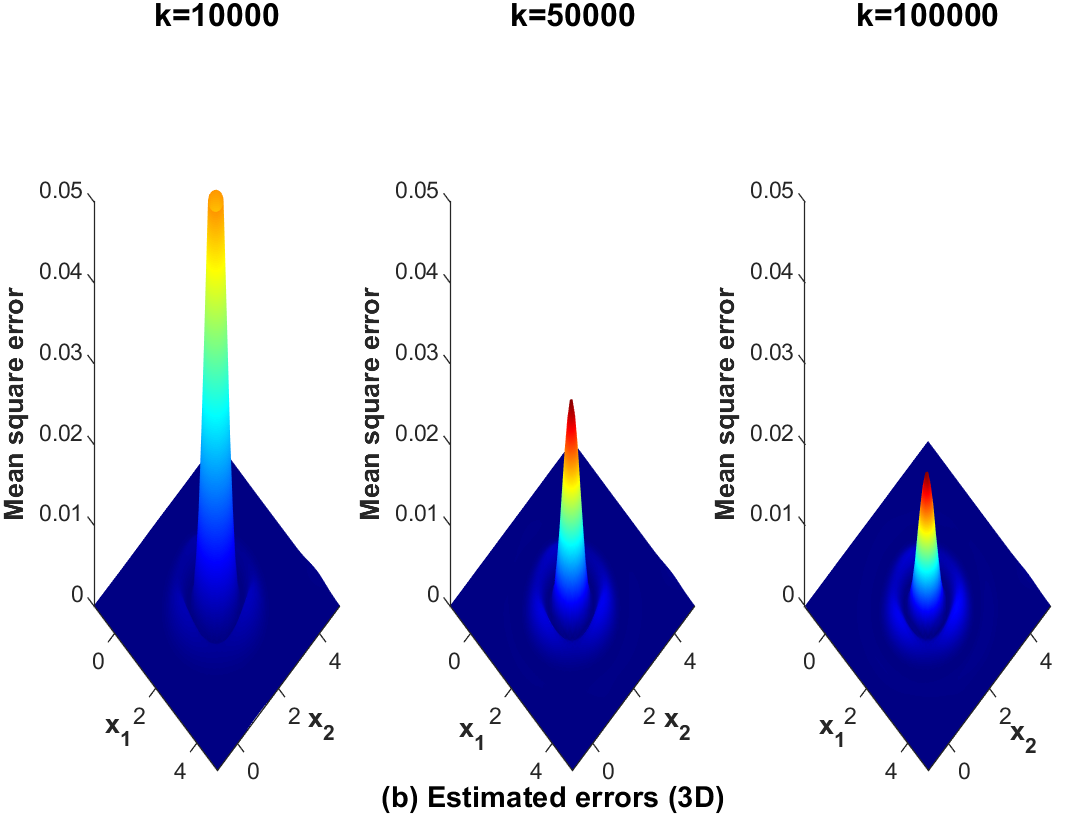}}

\caption{Two-dimensional input data: (a) Approximated functions and true function; (b) Mean square errors of approximated functions.}	
\label{twodimens}
	\end{figure}

Let $\X=[-1,5]^2$. The input data are generated following the rules below.
Let $\{(U_{1,k}, U_{2,k}),\ k\in \mathbb{N}\}$ be a sequence of independent two-dimensional random vectors, each with joint distribution function
$
C(u,v) = uv + \alpha u(1-u)v(1-v), \  (u,v)\in[0,1]^2,
$
where $\alpha=0.6$.
Both marginal distributions are uniform on $[0,1]$. The input data are given by
$
x_{1,k} = -1 + 6U_{1,k}, \
x_{2,k} =\frac{6(1+(-1)^{k-1})}{2k} - 1 +6\big(1 - \frac{1}{k}\big)U_{2,k}, \ k \in \mathbb{N}.
$
For each $k\in\mathbb{N}$, the joint distribution of $(x_{1,k},x_{2,k})$ admits a density
$
p_k(x_1,x_2)
 =
\frac{1}{36 (1-\frac{1}{k} )}
\big[
1+\alpha
 (1-\frac{x_1+1}{3} )
\big(
1- \frac{k}{3(k-1)}  (x_2- (\frac{6(1+(-1)^{k-1})}{2k}-1 ) )
\big)
\big]
\mathbf{1}_{[-1,5]}(x_1)\,
\mathbf{1}_{I_k}(x_2).
$
where
$
I_k=
 [
\frac{6(1+(-1)^{k-1})}{2k}-1,\
\frac{6(1+(-1)^{k-1})}{2k}-1+6 (1-\frac{1}{k} )
 ].
$
Take the  Gaussian kernel $K(\mathbf{x},\mathbf{y}) = e^{-\|\mathbf{x}-\mathbf{y}\|^2}, ~\forall ~\mathbf{x},\ \mathbf{y} \in \X$. The true function is given by $f^{\star}(\mathbf{x})=e^{-(x_1-2)^2-(x_2-2)^2}, ~\forall~ \mathbf{x}=(x_1,x_2) \in \X$. Fig.\ref{twodimens} (a) and (b) show
  $f_k(\mathbf{x},\omega_{1})$, $\mathbf{x} \in \X$, together with the   errors $\mathbb{E}\big[|f_k(\mathbf{x})-f^{\star}(\mathbf{x})|^2\big]$, $\mathbf{x} \in \X$, at iterations $k=10000,\ 50000,\ 100000$.  It follows that the algorithm (\ref{AL}) converges almost surely and in  mean square.

\section{Conclusions}

We have studied a recursive regularized learning algorithm in RKHS with dependent and non-stationary online data streams. We introduce the concept of a random Tikhonov regularization path and use it as a reference to analyze the convergence of the online algorithm. By reformulating the learning problem as a time-varying inverse problem and introducing an RKHS persistence of excitation condition, we show that the evolution of the regularization path converges to the unknown function.
We then analyze the mean square asymptotic stability of the associated random difference equations and establish the tracking  error  between the algorithm output and the regularization path. In particular, we show that if the regularization path varies sufficiently slowly, the tracking error converges to zero under appropriate choices of the algorithm gain  and the regularization parameter. Finally, for independent but non-identically distributed data streams, we provide more interpretable sufficient conditions based on marginal probability measures.

In the model (\ref{model}), the unknown function to be learned is time-invariant. However, in manufacturing systems (\cite{HU2017}), the underlying models may vary over time due to changes in the environment or input data. This motivates the study of learning time-varying models in future work.
Moreover, it is of interest to investigate acceleration techniques, such as averaged stochastic gradient algorithms (\cite{Polyak}-\cite{Moulinese}), the heavy-ball method (\cite{Polyak1}), and Nesterov’s accelerated gradient method (\cite{Nesterov}).

\begin{appendices}

\section{Proof in Section III}\label{xiaosiAAA}
\setcounter{equation}{0}
\def\theequation{A.\arabic{equation}}
\noindent
\textbf{Proof of Proposition \ref{mingti6}:}
For any given $k\in\mathbb N$, by the reproducing property of RKHS, Assumption \ref{ass2} and Proposition \ref{mingti2}, we get
\begin{align}
  &\mathrm{grad}\,J_k(f)\cr
 =&\frac{1}{2}\mathrm{grad}\,\E\left[(y_k-f(x_k))^2|\F_{k-1}\right]+\frac{1}{2}\lambda_k\mathrm{grad}\,\|f\|^2_{\HH_K}\cr
=&\frac{1}{2}\mathrm{grad}\,\E\left[f^2(x_k)|\F_{k-1}\right]-\mathrm{grad}\,\E[y_kf(x_k)|\F_{k-1}] +\frac{1}{2}\lambda_k\mathrm{grad}\,\|f\|^2_{\HH_K}\cr
=&\frac{1}{2}\mathrm{grad}\,\E\left[f(x_k)\left\langle K_{x_k},f\right\rangle_{\HH_K} |\F_{k-1}\right]-\mathrm{grad}\,\E[y_kf(x_k)|\F_{k-1}] +\frac{1}{2}\lambda_k\mathrm{grad}\,\|f\|^2_{\HH_K}\cr
=&\frac{1}{2}\mathrm{grad}\,\E\left[\left\langle f(x_k)K_{x_k},f\right\rangle_{\HH_K} |\F_{k-1}\right]-\mathrm{grad}\,\E[y_kf(x_k)|\F_{k-1}] +\frac{1}{2}\lambda_k\mathrm{grad}\,\|f\|^2_{\HH_K}\cr
 =& \frac{1}{2}\mathrm{grad}\,\langle T_{k}f,f\rangle_{\HH_K} -\mathrm{grad}\,
 \E[y_kf(x_k)|\F_{k-1}] +\lambda_kf~\mathrm{a.s.},\notag
\end{align}
where $\mathrm{grad}\,J_k:\HH_K\to\HH_K$ is the gradient operator. It follows from Proposition \ref{xingzhi1} that $T_{k}$ is self-adjoint a.s. By Proposition \ref{mingti2} and the reproducing property of RKHS, we obtain
$$ \mathrm{grad}\,\langle T_{k}f,f\rangle_{\HH_K} =2T_{k}f=2\E[f(x_k)K_{x_k}|\F_{k-1}]
~\mathrm{a.s.}$$
By the reproducing property of RKHS, Assumption \ref{ass2} and Proposition 2.6.31 in  \cite{Hytonen}, we have
\begin{align}
&\lim_{t\to 0}(\E\left[y_k(f+tg)(x_k)|\F_{k-1}\right]
-\E\left[y_kf(x_k)|\F_{k-1}\right])t^{-1}\notag\\
=& \langle \E[y_kK_{x_k}|\F_{k-1}],g\rangle_{\HH_K} ~\mathrm{a.s.},~\forall~ g\in \HH_K,\notag
\end{align}
which leads to $\mathrm{grad}\,\E[y_kf(x_k)|\F_{k-1}]=\E[y_kK_{x_k}|\F_{k-1}]$ a.s. Thus, we get (\ref{gradientoperator}).
Since $f_{\lambda,k}$ is the optimal solution of the optimization problem (\ref{youhua}), then $\mathrm{grad}\,J_k(f_{\lambda,k})=0$ a.s. Noting that $f_{\lambda,k}\in L^2(\Omega,\F_{k-1};\HH_K)$, by Assumption \ref{ass2} and Proposition \ref{mingti2}, we get (\ref{tidufcdddd}).

Especially, when $\lambda_k=0$, we know that $2J_k(f)=\E[(y_k  -  f(x_k))^2|\F_{k-1}]$. It follows from the statistical learning model (\ref{model}), Assumptions \ref{ass2}-\ref{ass1},  Proposition 2.6.31 in  \cite{Hytonen} and the reproducing property of RKHS that
\begin{align}
&\E[(y_k-f^{\star}(x_k))(f^{\star}(x_k)-f_k(x_k))|\F_{k-1}]\notag\\
 = & \E[v_k f^{\star}(x_k)|\F_{k-1}]-\E[v_kf_k(x_k)|\F_{k-1}]\notag\\
 = & \E[v_k\langle f^{\star},K_{x_k}\rangle_{\HH_K} |\F_{k-1}]-\E[v_k\langle f_k,K_{x_k}\rangle_{\HH_K} |\F_{k-1}]\notag\\
 = &\E[\langle f^{\star},  v_k   K_{x_k}\rangle_{\HH_K} |\F_{k-1}]  -\E[\langle f_k,v_kK_{x_k}\rangle_{\HH_K} |\F_{k-1}]\notag\\
 =& \langle f^{\star},  \E[v_kK_{x_k} |\F_{k-1}]\rangle_{\HH_K}   -\langle f_k,\E[v_kK_{x_k}|\F_{k-1}]\rangle_{\HH_K}
 =  0~\mathrm{a.s.},\notag
\end{align}
$\forall~ f_k\in L^0(\Omega,\F_{k-1};\HH_K).$
By the above, we get
\begin{align}
& \E\left[(y_k-f_k(x_k))^2|\F_{k-1}\right]\cr
 =& \E\left[(y_k-f^{\star}(x_k))^2|\F_{k-1}\right]+\E\left[(f^{\star}(x_k)-f_k(x_k))^2|\F_{k-1}\right]\cr
& +2\E\left[\left(y_k-f^{\star}(x_k)\right)\left(f^{\star}(x_k)-f_k(x_k)\right)|\F_{k-1}\right]\cr
 =&\E\left[(y_k-f^{\star}(x_k))^2|\F_{k-1}\right]+\E\left[(f^{\star}(x_k)-f_k(x_k))^2|\F_{k-1}\right]\cr
 \ge & \E\left[(y_k-f^{\star}(x_k))^2|\F_{k-1}\right]~\mathrm{a.s.},\notag
\end{align}
$\forall~ f_k\in L^0(\Omega,\F_{k-1};\HH_K),$ which shows that $f_{\lambda,k}=f^{\star}$.

When $\lambda_k>0$, it follows from Assumption \ref{ass2} and Proposition \ref{xingzhi2} that $\E[H_{k}+\lambda_k I|\F_{k-1}]$ is invertible a.s. By Assumption \ref{ass1}, we get
$
\E[v_kK_{x_k}|\F_{k-1}]=0~\mathrm{a.s.}
$
 Combining the statistical model (\ref{model}), (\ref{tidufcdddd}) and the reproducing property of RKHS gives
\begin{align}
 f_{\lambda,k}  =&\left( T_{k}+\lambda_k I \right)^{-1}
\E\left[y_kK_{x_k}|\F_{k-1}\right]\cr
 =&\left( T_{k}+\lambda_k I \right)^{-1}
\big(\E\left[f^{\star}(x_k)K_{x_k}|\F_{k-1}\right]
 +\E\left[v_kK_{x_k}|\F_{k-1}\right]\big)\cr
 =&\left(T_{k}+\lambda_k I\right)^{-1}T_{k}f^{\star}~\mathrm{a.s.},\hspace{-4pt}\notag
\end{align}
which shows that (\ref{tidufcddddcc}) holds.
\hfill $\blacksquare$

\section{Proofs in Section IV}\label{fuluBBBB}
\setcounter{equation}{0}
\def\theequation{B.\arabic{equation}}
\setcounter{lemma}{0}
\def\thelemma{B.\arabic{lemma}}
\setcounter{remark}{0}
\def\theremark{B.\arabic{remark}}
\setcounter{proposition}{0}
\def\theproposition{B.\arabic{proposition}}
We denote $\Phi_{i,j}=\prod_{k=j}^i\left(I-a_k\left(H_{k}+\lambda_kI\right)\right)$, if $i\ge j$, $\Phi_{i,j}=I$, otherwise.

\vskip 0.2cm
For analyzing the tracking error equation (\ref{EERR}), we consider the following two types of random difference equations with values in $\HH_K$, that is,
\begin{align}\label{cha1}
M_{k+1}=\left(I-a_k\left(H_{k}+\lambda_kI\right)\right)M_k-a_kw_k,
\end{align}
and
\begin{align}\label{cha2}
D_{k+1}\hspace{-2pt}=\hspace{-2pt}\left(I\hspace{-2pt}-a_k\left(
H_{k} +\lambda_kI\right)\right)D_k\hspace{-2pt}
-\hspace{-2pt}(d_{k+1}\hspace{-2pt}-d_k),
\end{align}
$\|M_0\|_{L^2\left(\Omega;\HH_K\right)}<\infty, ~\|D_0\|_{L^2\left(\Omega;\HH_K\right)}<\infty,~\forall~k\in \mathbb N,$ where $\{w_k,k\in \mathbb N\}$ and $\{d_k,k\in \mathbb N\}$ are both sequences of random elements with values in $\HH_K$. The following proposition provides a structural decomposition of the tracking error $\d_k$.
\vskip 0.2cm
\begin{proposition}\label{mingti100}
\rm{If the non-homogeneous terms and initial values of (\ref{cha1}) and (\ref{cha2}) are respectively given by
$w_k=(H_{k}+\lambda_kI)f_{\lambda,k}- H_{k} f^{\star}-v_kK_{x_k},\ d_k=f_{\lambda,k},\  M_0=f_0, D_0=-f_{\lambda,0},~\forall~k\in \mathbb N,$
then
\bna\label{fenjie}
\d_k=M_k+D_k,~\forall~ k\in \mathbb N.
\ena
}
\end{proposition}

\vskip 0.2cm
\begin{proof}
By the random difference equations (\ref{cha1})-(\ref{cha2}), as well as the tracking error equation (\ref{EERR}), we obtain
\begin{align}\label{yiy}
  M_{k+1}+D_{k+1}-\d_{k+1}
 =&\left(I-a_k\left(H_{k}+\lambda_kI\right)\right)\left(M_k+D_k-\d_k\right)\cr
 =& \Phi(k,0)\left(M_0+D_0-\d_0\right),~\forall~k\in\mathbb N.
\end{align}
Noting that $M_0+D_0-\d_0=f_0-f_{\lambda,0}-\d_0=0$, it follows from (\ref{yiy}) that (\ref{fenjie}) holds.
\end{proof}
\vskip 0.2cm

Proposition \ref{mingti100} shows that the tracking error $\d_k$ can be decomposed into two parts including (i) $M_k$, which is jointly determined by the sampling error of the Tikhonov regularization path and the multiplicative noise; (ii) $D_k$, which is determined by the drift error of the Tikhonov regularization path. In fact, by Assumptions \ref{ass2}-\ref{ass1}, Proposition \ref{mingti2} and Proposition \ref{mingti6}, we get
$$\E[w_k|\F_{k-1}]
 = (T_{k}+\lambda_k I ) f_{\lambda,k}
-T_{k}f^{\star}-\E\left[v_kK_{x_k}|\F_{k-1}\right]=0,
$$
which means that $\{w_k,\F_k,k\in \mathbb N\}$ is a martingale difference sequence with values in $\HH_K$.
Thus, the tracking error equation (\ref{EERR}) can be essentially decomposed into two types of random difference equations including (i) the random difference equation (\ref{cha1}), whose non-homogeneous term is a martingale difference sequence dependent on the homogeneous term; and (ii) the random difference equation (\ref{cha2}), whose non-homogeneous term is the drift of the Tikhonov regularization path.

\vskip 0.2cm
We obtain the lemmas on asymptotic mean square stabilities  of (\ref{cha1})-(\ref{cha2}), which are crucial for the consistency analysis of the algorithm.

\vskip 0.1cm
\begin{lemma}\label{LLEEMM6}
\rm{Suppose that Assumption \ref{ass2} and Condition \ref{con1} hold. For the random difference equation (\ref{cha1}), if $\{w_k,\F_k,k\in \mathbb N\}$ is a martingale difference sequence with values in  $\HH_K$ satisfying  $\sup_{k\in \mathbb N}\|w_k\|_{L^2(\Omega;\HH_K)}<\infty$, then the solution sequence $\{M_k,k\in \mathbb N\}$ of (\ref{cha1}) is asymptotically  mean square stable, i.e.  $\lim_{k\to\infty}\|M_k\|_{L^2(\Omega;\HH_K)}=0,$  and
$$\left\|M_{k+1}\right\|_{L^2\left(\Omega;\HH_K\right)}
\leq C_F \beta_0(k),\ \forall \ k\geq k_0,$$
where $\beta_0(k)=(k+1)^{-\frac{
 \tau_1-3\tau_2}{2}} \ln^{\frac{3}{2}}(k+1)$,  $C_F =C_{01}C_3\big(
\alpha_1 C_0
+(\kappa+\alpha_2+1)(\alpha_1^2+\alpha_1)C_0 k_0
+\|M_0\|_{L^2(\Omega;\mathcal{H}_K)}
\big)
+
C_0 C_2\big((\kappa+\alpha_2)+\frac{\tau_1 2^{\tau_1}}{\alpha_1}\big)$, $C_0=\sup_{k\in\mathbb N}\|w_k\|_{L^2(\Omega;\HH_K)}$, $k_{0}, \ C_{01}, \ C_2, \ C_3 $ are given in  Corollary \ref{wuxiannsuanfasud}.
}
\end{lemma}


\begin{proof}
 Denote    $S_{k,i}=\sum_{j=i}^kw_j, \forall ~k,i\in\mathbb N.$
For integers $i>j\ge 0$, we have $w_j\in L^0(\Omega,\F_{i-1};\HH_K)$ and $w_i\in L^2(\Omega;\HH_K)$, which together with Proposition 2.6.31 in \cite{Hytonen} gives
 $\E\left[\langle w_i,w_j\rangle_{\HH_K} \right]=\E\left[\E\left[\langle w_i,w_j\rangle_{\HH_K} |\F_{i-1}\right]\right]=\E\left[\langle \E[w_i|\F_{i-1}], w_j\rangle_{\HH_K} \right]=0,~\forall ~i>j\ge 0.$ Then, we have
\begin{align}\label{xak}
 \hspace{-14pt}\|S_{k,i}\|_{L^2\left(\Omega;\HH_K\right)}\hspace{-3pt}
 =&  \bigg(  \sum_{j=i}^k\E\big[ \|w_j \|_{\HH_K}^2\big] \bigg)^{\frac{1}{2}}
 \hspace{-3pt} \leq \hspace{-3pt}  C_0\sqrt{k-i+1}.\hspace{-5pt}
\end{align}
By Condition \ref{con1} and $\ln (k+1)^{\frac{1}{2}+\frac{\tau_{1}-3\tau_{2}}{2}}=o\big(\frac{\alpha_{1}\alpha_{2}}{1-\tau_1-\tau_2} (k+1)^{1-\tau_1-\tau_2}\big)$,  we know that $0<1-a_{k}\lambda_k<1$ and $\ln (k+1)^{\frac{1}{2}+\frac{\tau_{1}-3\tau_{2}}{2}}\leq   \alpha_{1}\alpha_{2}(1-\tau_1-\tau_2)^{-1}  (k+1)^{1-\tau_1-\tau_2}$, $\forall \ k\geq k_{0}.$
Noting that
\begin{align}
& \sum_{i=0}^ka_i\prod_{j=i+1}^k(I-a_j(H_j+\lambda_jI))w_i\cr
 =& \sum_{i=0}^ka_i\Phi_{k,i+1}(S_{k,i}-S_{k,i+1})\cr
 =&a_0\Phi_{k,1}S_{k,0}+\sum_{i=1}^k\left(a_i\Phi_{k,i+1}-a_{i-1}\Phi_{k,i}\right)S_{k,i}\cr
 =&a_0\Phi_{k,1}S_{k,0}+\sum_{i=1}^k\big(a^2_i\Phi_{k,i+1}
 (H_i+\lambda_iI) +(a_i-a_{i-1})\Phi_{k,i}\big)S_{k,i},~\forall~k\in\mathbb N,\notag
\end{align}
and $ \|H_k+\lambda_kI\|_{\mathscr L(\HH_K)} \leq \|H_k\|_{\mathscr L(\HH_K)}+\lambda_k
 \leq  K(x_k,x_k)+\lambda_k
 \leq \kappa+\alpha_{2}~\mathrm{a.s.},~\forall~k\in\mathbb N,$
then by Lemma \ref{lemma1111} and Minkowski inequality, we get
\begin{align} \label{ppff}
& \bigg\|\sum_{i=0}^ka_i\prod_{j=i+1}^k(I-a_j(H_j
+\lambda_jI))w_i \bigg\|_{L^2\left(\Omega;\HH_K\right)}\cr
 \leq & a_0\|\Phi_{k,1}S_{k,0}\|_{L^2\left(\Omega;\HH_K\right)} +\sum_{i=1}^k\left\|a^2_i\Phi_{k,i+1}(H_i+\lambda_iI)S_{k,i}
 \right\|_{L^2\left(\Omega;\HH_K\right)}\cr & +\sum_{i=1}^k\left\|(a_i-a_{i-1})\Phi_{k,i}S_{k,i}\right\|_{L^2\left(\Omega;\HH_K\right)}\cr
 \leq & a_0\|\Phi_{k,1}\|_{\mathscr L(\HH_K)}\|S_{k,0}\|_{L^2\left(\Omega;\HH_K\right)} +\hspace{-5pt}\sum_{i=1}^k \hspace{-3pt} a^2_i\left\|H_i\hspace{-3pt}+\hspace{-3pt}\lambda_iI\right\|_{\mathscr L(\HH_K)}\left\|\Phi_{k,i+1}\right\|_{\mathscr L(\HH_K)} \left\|S_{k,i}\right\|_{L^2\left(\Omega;\HH_K\right)}\notag\\
 &+\sum_{i=1}^k(a_{i-1}-a_i)\left\|\Phi_{k,i}\right\|_{\mathscr L(\HH_K)}\left\|S_{k,i}\right\|_{L^2\left(\Omega;\HH_K\right)}\cr
 \leq & \Gamma_1+\Gamma_2+\Gamma_3,\ \forall~ k\ge k_0,
\end{align}
where $\Gamma_1=a_0 C_{01} \prod_{i=k_0}^k(1-a_i\lambda_i)\|S_{k,0}\|_{L^2\left(\Omega;\HH_K\right)}, \ \Gamma_2 =(\kappa +\alpha_{2}) C_{01} \sum_{i=1}^{k_0-1}a_i^2\prod_{j=k_0}^k(1-a_j\lambda_j)\\ \|S_{k,i}\|_{L^2\left(\Omega;\HH_K\right)}+ C_{01} \sum_{i=1}^{k_0-1} (a_{i-1} -a_i) \prod_{j=k_0}^k(1-a_j\lambda_j)  \|S_{k,i}\|_{L^2\left(\Omega;\HH_K\right)}, \ \Gamma_3  =(\kappa+ \alpha_{2})\sum_{i=k_0}^ka_i^2 \\ \prod_{j=i+1}^k(1-a_j\lambda_j)\|S_{k,i}\|_{L^2\left(\Omega;\HH_K\right)}+\sum_{i=k_0}^k(a_{i-1}-a_i)\prod_{j=i+1}^k  (1-a_j\lambda_j)
\|S_{k,i}\|_{L^2\left(\Omega;\HH_K\right)}$.
Below we analyze the right-hand side of the last inequality in (\ref{ppff}) term by term. By Condition \ref{con1} and (\ref{xak}), we get
\begin{align}
 \Gamma_1
\leq & a_0 C_{01} C_0\sqrt{k+1}\prod_{i=k_0}^k\bigg(1-\frac{\alpha_{1}
\alpha_{2}}{(i+1)^{\tau_1+\tau_2}}\bigg),\label{exp1}
\end{align}
$\forall~ k\ge k_0.$ Noting that
$
\prod_{j=k_0}^k\big(1-\frac{\alpha_{1}\alpha_{2}
}{(j+1)^{\tau_1+\tau_2}}\big)\leq \exp\big(-\sum_{j=k_0}^k\frac{\alpha_{1}
\alpha_{2}}{(j+1)^{\tau_1+\tau_2}}\big),
$
$~\forall~ k\ge k_0,$ and $\sum_{j=k_0}^k  \frac{1}{(j+1)^{\tau_1+\tau_2}}
\ge  \frac{\left((k+1)^{1-\tau_1-\tau_2} - (k_0+1)^{1-\tau_1-\tau_2}\right)}{1-\tau_1-\tau_2},$
by Condition \ref{con1}, we obtain
\begin{align}
&\prod_{j=k_0}^k\big(1- \alpha_{1}\alpha_{2}
 (j+1)^{-\tau_1-\tau_2} \big)\notag\\
\leq & \exp\left(- \alpha_{1}\alpha_{2}(1-\tau_1-\tau_2)^{-1} (k+1)^{1-\tau_1-\tau_2}  \right)   \exp\left(  \alpha_{1}\alpha_{2} (1-\tau_1-\tau_2)^{-1} (k_0+1)^{1-\tau_1-\tau_2} \right)\notag\\
\leq &
C_3 (k+1)^{-\frac{\frac{1}{2}+
 \tau_1-3\tau_2}{2}} \ln^{\frac{3}{2}}(k+1). \label{mk1}
\end{align}
%
It follows from Condition \ref{con1}, (\ref{xak}) and (\ref{mk1}) that
\begin{align}\label{ppff1}
\Gamma_1
 \leq  &\alpha_{1} C_0 C_{01} C_3  \beta_0(k),
\end{align}
which leads to
\begin{align}
 \Gamma_2
 \leq &(\kappa+\alpha_{2}+1)   (\alpha_{1}^2+\alpha_{1})C_0k_0   C_{01}  C_3  \beta_0(k).\label{wwiijj}
\end{align}
By Condition \ref{con1}, we get
$ \frac{a_{k-1} -a_k}{a^2_k} \leq \frac{\tau_1 2^{\tau_1}}{\alpha_1}, \forall \ k\geq 1.
$ This together with Lemma \ref{lemma111} shows that
\begin{align}\label{ppff2}
 \Gamma_3 \leq
 C_0 C_2 \Big((\kappa+\alpha_{2}) + \frac{\tau_1 2^{\tau_1}}{\alpha_1}\Big)   \beta_0(k).
\end{align}
Taking (\ref{ppff1})-(\ref{ppff2}) into (\ref{ppff}) leads to
\begin{align}\label{mk2}
&\bigg\|\sum_{i=0}^ka_i\prod_{j=i+1}^k(I-a_j(H_j
+\lambda_jI))w_i\bigg\|_{L^2\left(\Omega;\HH_K\right)}\notag\\
\leq & (C_F-  C_{01} C_3\|M_0\|_{L^2\left(\Omega;\HH_K\right)}    ) \beta_0(k).
\end{align}
Hence, by  (\ref{cha1}),  (\ref{mk2}), Lemma  \ref{lemma1111} and Minkowski inequality, we obtain
\begin{align}
& \|M_{k+1}\|_{L^2\left(\Omega;\HH_K\right)}\notag\\
 \leq& \bigg\|\prod_{i=0}^k\left(I-a_i\left(H_i+\lambda_iI\right)\right)M_0\bigg\|_{L^2\left(\Omega;\HH_K\right)} +\bigg\|\sum_{i=0}^ka_i\prod_{j=i+1}^k(I-a_j(H_j+\lambda_jI))w_i\bigg\|_{L^2\left(\Omega;\HH_K\right)}\notag\\
 \leq& \left\|\Phi_{k,0}\right\|_{\mathscr L(\HH_K)}\left\|M_0\right\|_{L^2\left(\Omega;\HH_K\right)} +\bigg\|\sum_{i=0}^ka_i\prod_{j=i+1}^k(I-a_j(H_j+\lambda_jI))w_i\bigg\|_{L^2\left(\Omega;\HH_K\right)}\notag\\
 \leq & C_{01} \prod_{i=k_0}^k\left(1-a_i\lambda_i\right)\|M_0\|_{L^2\left(\Omega;\HH_K\right)} +\bigg\|\sum_{i=0}^ka_i\prod_{j=i+1}^k(I-a_j(H_j+\lambda_jI))w_i\bigg\|_{L^2\left(\Omega;\HH_K\right)}\notag\\
\leq &C_{F} \beta_0(k).  \notag
\end{align}
Then, by Condition \ref{con1}, we have $\lim\limits_{k\to\infty}\|M_k\|_{L^2(\Omega;\HH_K)}=0.$
\end{proof}

\begin{lemma}\label{LLEEMM7}
\rm{Suppose that Assumption \ref{ass2} and Condition \ref{con1} hold. For the random difference equation (\ref{cha2}), if $\{d_k,k\in \mathbb N\}$ is a sequence of random elements with values in  $\HH_K$ satisfying

(i)  $\sup_{k\in \mathbb N}\|d_k\|_{L^2(\Omega;\HH_K)}<\infty$, and
\bna\label{jianjindj}
\lim_{k\to\infty}\sum_{i=0}^k\left\|d_{i+1}-d_i\right\|_{L^2\left(\Omega;\HH_K\right)}\prod_{j=i+1}^k\left(1-a_j\lambda_j\right)=0,
\ena
then the solution sequence $\{D_k,k\in \mathbb N\}$ of (\ref{cha2}) is asymptotically  mean square stable, i.e. $\lim_{k\to\infty}\|D_k\|_{L^2(\Omega;\HH_K)}=0$;

(ii) $\sup_{k\in \mathbb N}\|d_k\|_{\HH_K}<\infty$, and
\bna\label{jianjindj000}
\lim_{k\to\infty}\sum_{i=0}^k\left\|d_{i+1}-d_i\right\|_{\HH_K}\prod_{j=i+1}^k\left(1-a_j\lambda_j\right)=0,
\ena
then the solution sequence $\{D_k,k\in \mathbb N\}$ of (\ref{cha2}) is almost surely  asymptotically stable, i.e. $\lim_{k\to\infty}\|D_k\|_{\HH_K}=0$ a.s.

(iii) $\sup_{k\in \mathbb N}\|d_k\|_{L^2(\Omega;\HH_K)}<\infty$, and there exists $C_4>0$, such that
\bna\label{sududk1}
 \left\|d_{k+1}-d_k\right\|_{L^2\left(\Omega;\HH_K\right)}\leq C_4 a_k^2, \ \forall \ k\geq  0,
\ena
then, for any $ k \geq k_{0},$
\begin{align}\label{dksudu}
 \|D_k\|_{L^2(\Omega;\HH_K)}\leq  C_5 \beta_0(k),
\end{align}
where $ C_5 =
C_{01}  C_3 \left\|D_0\right\|_{L^2\left(\Omega;\HH_K\right)}+ C_4 C_2
+ 2 C_{01}  k_0 C_3   \sup_{k\in \mathbb N}  \|d_k\|_{L^2(\Omega;\HH_K)}$,  $k_{0}, \ C_{01}, \ C_2, \ C_3 $ are given in  Corollary \ref{wuxiannsuanfasud}.

}
\end{lemma}
\begin{proof}
 By the difference equation (\ref{cha2}), we get
\begin{align}\label{henfan1}
D_{k+1}
=\Phi_{k,0}D_0-\sum_{i=0}^k\Phi_{k,i+1}\left(d_{i+1}-d_i\right),
\end{align}
$\forall~k\in\mathbb N.$ It follows from (\ref{henfan1}), Assumption \ref{ass2}, Condition \ref{con1}, Lemma D.1  and Minkowski inequality that, there exists $k_{0} \in \mathbb{N}$, such that
\begin{align}\label{henfan2}
&\|D_{k+1}\|_{L^2 (\Omega;\HH_K )}\notag\\
\leq & \left\|\Phi_{k,0}D_0\right\|_{L^2\left(\Omega;\HH_K\right)}\hspace{-3pt}
+\bigg\|\sum_{i=0}^k\Phi_{k,i+1}\left(d_{i+1}\hspace{-3pt}-d_i\right)\bigg\|_{L^2\left(\Omega;\HH_K\right)}\notag\\
\leq & \left\|\Phi_{k,0}\right\|_{\mathscr L(\HH_K)}\left\|D_0\right\|_{L^2\left(\Omega;\HH_K\right)}  +\sum_{i=0}^k\|\Phi_{k,i+1}\left(d_{i+1}-d_i\right)\|_{L^2\left(\Omega;\HH_K\right)}\notag\\
\leq & \left\|\Phi_{k,0}\right\|_{\mathscr L(\HH_K)}\left\|D_0\right\|_{L^2\left(\Omega;\HH_K\right)}
 +\sum_{i=0}^k\left\|\Phi_{k,i+1}\right\|_{\mathscr L(\HH_K)}\left\|d_{i+1}-d_i\right\|_{L^2\left(\Omega;\HH_K\right)}\notag\\
\leq & C_{01}  \prod_{i=k_0}^k\left(1-a_i\lambda_i\right)\left\|D_0\right\|_{L^2\left(\Omega;\HH_K\right)}
 +\sum_{i=0}^{k_0-1}\left\|\Phi_{k,i+1}\right\|_{\mathscr L(\HH_K)}\left\|d_{i+1}-d_i\right\|_{L^2\left(\Omega;\HH_K\right)}\notag\\
& +\sum_{i=k_0}^k\left\|d_{i+1}-d_i\right\|_{L^2\left(\Omega;\HH_K\right)}\prod_{j=i+1}^k\left(1-a_j\lambda_j\right)\notag\\
\leq & C_{01} \exp\bigg(-\sum_{i=k_0}^ka_i\lambda_i\bigg)\left\|D_0\right\|_{L^2\left(\Omega;\HH_K\right)} +C_{01} \sum_{i=0}^{k_0-1}\exp\bigg(-\sum_{j=k_0}^ka_j\lambda_j\bigg)\left\|d_{i+1}-d_i\right\|_{L^2\left(\Omega;\HH_K\right)}\notag\\
& +\sum_{i=k_0}^k\left\|d_{i+1}-d_i\right\|_{L^2\left(\Omega;\HH_K\right)}\prod_{j=i+1}^k\left(1-a_j\lambda_j\right),
\end{align}
$\forall~k\ge k_0.$
By Condition \ref{con1}, we obtain
\bna\label{henfan3}
\lim_{k\to\infty}\exp\bigg(-\sum_{i=k_0}^ka_i\lambda_i\bigg)=0.
\ena
Noting that $\|D_0\|_{L^2(\Omega;\HH_K)}<\infty$ and $\sup\limits_{k\in\mathbb N}\|d_k\|_{L^2(\Omega;\HH_K)}<\infty$, then by (\ref{jianjindj}), (\ref{henfan2})-(\ref{henfan3}), we have
$
\lim\limits_{k\to\infty}\|D_k\|_{L^2\left(\Omega;\HH_K\right)}=0.
$
Similar to the above proof, it can be verified that if $$\lim_{k\to\infty}\sum_{i=0}^k \|d_{i+1}-d_i \|_{\HH_K}\prod_{j=i+1}^k\left(1-a_j\lambda_j\right)=0,$$ then $\lim\limits_{k\to\infty}\|D_k\|_{\HH_K}=0$ a.s.
By (\ref{sududk1}),   (\ref{henfan2}) and Lemma   \ref{lemma111}, we have (\ref{dksudu}).
\end{proof}

\vskip 2mm

Based on the above proposition and lemmas, we can prove  Lemma \ref{THMM}.

\vskip 2mm
\noindent
\textbf{Proof of Lemma \ref{THMM}:} By Proposition \ref{mingti100}, we have
\begin{align}
\d_k=M_k+D_k,~\forall~ k\in \mathbb N,\label{ECX}
\end{align}
where $
M_{k+1}=\left(I-a_k\left(H_{k}+\lambda_kI\right)\right)M_k-a_kw_k,$
and $
D_{k+1} =\hspace{-2pt}\left(I -a_k\left(
H_{k}+\lambda_kI\right)\right)D_k\hspace{-2pt}
- (d_{k+1} -d_k),$
$w_k=(H_{k}+\lambda_kI)f_{\lambda,k}- H_{k} f^{\star}-v_kK_{x_k},\ d_k=f_{\lambda,k},\  M_0=f_0, D_0=-f_{\lambda,0},~\forall~k\in \mathbb N$.
 It follows from the definition of the regularization path $f_{\lambda,k}$  and Assumption \ref{ass1} that
$ \E[(H_k+\lambda_kI)f_{\lambda,k}-H_k f^{\star}-v_kK_{x_k}|\F_{k-1}]
 = ( T_{k} + \lambda_kI)f_{\lambda,k}-
 T_{k} f^{\star}
 =0,~\forall~ k\in\mathbb N.
$
By Minkowski inequality and Assumption \ref{ass2}, we know that
$\sup_{k\in\mathbb N}\|(H_k+\lambda_kI)f_{\lambda,k}-H_k f^{\star}\|_{L^2\left(\Omega;\HH_K\right)}
 \leq \sup_{k\in\mathbb N}(\kappa+\alpha_{2})\|f_{\lambda,k}\|_{L^2\left(
 \Omega;\HH_K\right)}
+\kappa\|f^{\star}\|_{\HH_K} \leq (2\kappa+\alpha_{2})\|f^{\star}\|_{\HH_K},
$
from which we conclude that $\{(H_k+\lambda_kI)f_{\lambda,k}-H_k f^{\star} -v_kK_{x_k},\F_{k},k\in\mathbb N\}$ is a $L_2$-bounded martingale difference sequence. Thus,   by Assumption \ref{ass2}, Condition \ref{con1} and Lemma \ref{LLEEMM6}, we get
\begin{align}\label{ECX3}
  \lim_{k\to\infty}\left\|M_k\right\|_{L^2
 \left(\Omega;\HH_K\right)}=0.
\end{align}
By (\ref{hmp}) and Lemma \ref{LLEEMM7}, we have
$
  \lim_{k\to\infty}\left\|D_k\right\|_{L^2
 \left(\Omega;\HH_K\right)}=0.
$
Then, by Minkowski inequality, we obtain (\ref{dkjunf}).

By Assumptions \ref{ass2}-\ref{ass1}, $C_r$ inequality, noting that $M_k \in \F_{k-1}$, then
\begin{align}
& E\left[\left\|M_{k+1}\right\|_{\HH_K}^2|  \F_{k-1}  \right]\notag\\
=&  E\left[\left\| \left(I-a_k\left(H_{k}+\lambda_kI\right)\right)M_k-a_kw_k \right\|_{\HH_K}^2|  \F_{k-1}  \right]\notag\\
\leq &  E\left[\left\| \left(I-a_k\left(H_{k}+\lambda_kI\right)\right)M_k\right\|_{\HH_K}^2|  \F_{k-1}  \right]\notag\\
&-2  E\left[\left\langle \left(I-a_k\left(H_{k}+\lambda_kI\right)\right)^2 M_k, w_k  \right\rangle_{\HH_K}  |  \F_{k-1}  \right] + a_k^2  E\left[\left\| w_k \right\|_{\HH_K}^2 \big|  \F_{k-1}  \right]\notag\\
\leq & (1 + a_k^2 (\kappa+\alpha_2)^2 + 4 a_k^2 \kappa^2) \left\|M_{k}\right\|_{\HH_K}^2  +5 a_k^2  E\left[\left\| w_k \right\|_{\HH_K}^2|  \F_{k-1}  \right]. \notag
\end{align}
By the above inequality, Theorem 1 in \cite{Robbins} and Condition \ref{con1}, we know that $M_k$ converges a.s. as $k$ tends to infinity, which together with (\ref{ECX3}) shows that $ \lim_{k\to\infty}\left\|M_k\right\|_{ \HH_K }=0,$ a.s. By (\ref{hmp00}) and Lemma \ref{LLEEMM6}, we have
 $
  \lim_{k\to\infty}\left\|D_k\right\|_{ \HH_K }=0,$ a.s. These together with triangle inequality show (\ref{dkjihu}). 
\hfill $\blacksquare$

\vskip 0.2cm
Now, we will analyse the difference between $f_{\lambda,k}$ and $f^{\star}$. At first, we introduce the following  auxiliary variable and develop some lemmas.

\vskip 0.2cm
For any given integer $h>0$, let $f_{\lambda,k,h}=\big( \phi_{k,k+h-1}
+ \psi_{k,k+h-1}I\big)^{-1}  \phi_{k,k+h-1}  f^{\star},$
$\phi_{k,k+h-1}=\sum_{i=k}^{k+h-1}   \E[H_i|\F_{k-1}]$, $\psi_{k,k+h-1}=\sum_{i=k }^{k+h-1}\lambda_i.$
\vskip 0.2cm

\begin{lemma}\label{lemma222}
\rm{If Assumption \ref{ass2} and Condition \ref{con1} hold,  and
\begin{itemize}
  \item[(i)] $\left\|f_{\lambda,k+1}-f_{\lambda,k}\right\|_{L^2\left(\Omega;\HH_K\right)}=o\left(\lambda_k\right),$
then
\bna\label{wds}
\lim_{k\to\infty}\left\|f_{\lambda,k,h}-f_{\lambda,k}\right\|_{L^2\left(\Omega;\HH_K\right)}=0;
\ena
  \item[(ii)] there exists $\hat{C}_5>0$, such that $ \|f_{\lambda,k+1}-f_{\lambda,k} \|_{L^2\left(\Omega;\HH_K\right)}\leq \hat{C}_5 a_k^2,$
then, for $ k\geq 1$,
\begin{align}\label{wds00}
 \left\|f_{\lambda,k,h}-f_{\lambda,k}\right\|_{L^2\left(\Omega;\HH_K\right)}
 \leq   2\hat{C}_5  \alpha_1 (\kappa+\alpha_{2})\beta_0(k);
\end{align}
  \item[(iii)]  for $j=k,\cdots, k+h-2,$ $\mathbb{E}\big[ \|f_{\lambda,j+1}-f_{\lambda,j} \|_{\HH_K}^2|\mathcal{F}_{k-1}\big]^{\frac{1}{2}}=o(\lambda_k)~\text{a.s.},$
 then
\begin{align}
\lim _{k \rightarrow \infty}\left\| f_{\lambda,k,h}-f_{\lambda,k} \right\|_{\HH_K}=0~\text{a.s.} \label{assumonre3as}
\end{align}

\end{itemize}

}
\end{lemma}

\vskip 0.1cm
\begin{proof}
It follows from the definitions of $f_{\lambda,k}$ and $f_{\lambda,k,h}$ that
\begin{align}\label{bo1}
 & f_{\lambda,k,h}-f_{\lambda,k}\notag\\
 = &\big(\phi_{k,k+h-1}+ \psi_{k,k+h-1} I\big)^{-1}
 \bigg(\sum_{i=k }^{k+h-1}\E\big[ (\E\left[H_i|\F_{i-1}\right] +\lambda_iI )(f_{\lambda,i}-f_{\lambda,k})|\F_{k-1}\big]\bigg).
\end{align}
Noting that
$
\big\|\big(\phi_{k,k+h-1}+ \psi_{k,k+h-1} I\big)^{-1}\big\|  \leq \psi^{-1}_{k,k+h-1} ~\mathrm{a.s.},\  \forall~ k\in\mathbb N,
$
then by Assumption \ref{ass2}, Condition \ref{con1}, Minkowski inequality and (\ref{bo1}), we get
\begin{align}\label{hhb1}
& \left\|f_{\lambda,k,h}-f_{\lambda,k}\right\|_{L^2\left(\Omega;\HH_K\right)}\cr
 \leq & \frac{1}{\alpha_{2} h}(k+h+1)^{\tau_2} \bigg\|\sum_{i=k}^{k+h-1}
\E\big[\left(\E\left[H_i|\F_{i-1}\right]+\lambda_iI\right) (f_{\lambda,i}-f_{\lambda,k})|\F_{k-1}\big]
\bigg\|_{L^2\left(\Omega;\HH_K\right)}\cr
 \leq &  \frac{1}{\alpha_{2}h}(k+h+1)^{\tau_2} \sum_{i=k}^{k+h-1}\big(\E\big[\big\|
\E\big[\big(\E\left[H_i|\F_{i-1}\right]+\lambda_iI\big)  (f_{\lambda,i}-f_{\lambda,k})|\F_{k-1}\big]\big\|_{\HH_K}^2\big]\big)^{\frac{1}{2}}\cr
 \leq & \frac{1}{\alpha_{2}h}(k+h+1)^{\tau_2}\sum_{i=k}^{k+h-1}\Big(\E\big[\big(
\E\big[ \|\left(\E\left[H_i|\F_{i-1}\right]+\lambda_iI\right) (f_{\lambda,i}-f_{\lambda,k}) \|_{\HH_K}
|\F_{k-1}\big]\big)^2\big]\Big)^{\frac{1}{2}}\cr
 \leq & \frac{1}{\alpha_{2}h}(k+h+1)^{\tau_2}\sum_{i=k}^{k+h-1}\big(\E\big[
\E\big[\big\|\big(\E\left[H_i|\F_{i-1}\right]+\lambda_iI\big)
(f_{\lambda,i}-f_{\lambda,k})
\big\|_{\HH_K}^2|\F_{k-1}\big]\big]\big)^{\frac{1}{2}}\cr
 \leq & \frac{\kappa+\alpha_{2}}{\alpha_{2}h}(k+h+1)^{\tau_2}
\sum_{i=k}^{k+h-1}\left\|f_{\lambda,i}-f_{\lambda,k}\right\|_{L^2\left(\Omega;\HH_K\right)}\cr
 =& O\bigg((k+1)^{\tau_2}\sum_{i=k}^{k+h-1}\|f_{\lambda,i+1}-f_{\lambda,i}\|_{L^2\left(\Omega;\HH_K\right)}\bigg).
\end{align}
Then, by Condition \ref{con1} and $\left\|f_{\lambda,k+1}-f_{\lambda,k}\right\|_{L^2\left(\Omega;\HH_K\right)}=o\left(\lambda_k\right),$  we obtain (\ref{wds}).
 By $ \|f_{\lambda,k+1}-f_{\lambda,k} \|_{L^2\left(\Omega;\HH_K\right)}\leq \hat{C}_5 a_k^2$ and (\ref{hhb1}), we have $ \left\|f_{\lambda,k,h}-f_{\lambda,k}\right\|_{L^2\left(\Omega;\HH_K\right)}\leq   2\hat{C}_5  \alpha_1 (\kappa+\alpha_{2})  (k+1)^{\tau_2-2\tau_1}   \leq 2 \hat{C}_5   \alpha_1 (\kappa+\alpha_{2})(k+1)^{-\frac{\tau_1-3
\tau_2}{2}} \ln^{\frac{3}{2}}(k+1),$
that is, (\ref{wds00}) holds.
By (\ref{bo1}),  Assumption \ref{ass2}, conditional Jensen inequality, Similar to the proof of (\ref{hhb1}), we have
\begin{align}
&\|f_{\lambda,k,h}-f_{\lambda,k}\|_{\HH_K}\notag\\
\leq &  \frac{1}{\alpha_{2} h }(k+h+1)^{\tau_2} \sum_{i=k}^{k+h-2} \left\|\E\big[ \| f_{\lambda,i+1}-f_{\lambda,i}\|_{\HH_K}|\F_{k-1}\big]\right\|\notag\\
\leq &  \frac{1}{\alpha_{2} h }(k+h+1)^{\tau_2} \sum_{i=k}^{k+h-2}\left(  \E\big[ \| f_{\lambda,i+1}-f_{\lambda,i}\|_{\HH_K}|^2\F_{k-1}\big]\right)^{\frac{1}{2}}.\notag
\end{align}
This together with $\mathbb{E}\big[ \|f_{\lambda,i+1}-f_{\lambda,i} \|_{\HH_K}^2|\mathcal{F}_{k-1}\big]^{\frac{1}{2}}=o(\lambda_k)~\text{a.s.}$ gives (\ref{assumonre3as}).
\end{proof}

%
%
\vskip 0.1cm

\begin{lemma}\label{lemma333}
\rm{If Assumption \ref{ass2}} and Condition \ref{con1} hold, and the online data streams $\{(x_k,y_k),\ k  \in\mathbb N\}$ satisfy the RKHS persistence of excitation condition, then
$\lim_{k\to\infty}\|f_{\lambda,k,h}-f^{\star}\|_{L^2\left(\Omega;\HH_K\right)}=0$, $\lim_{k\to\infty} \| f_{\lambda,k,h}-f^{\star}\|_{ \HH_K } =0~\text{a.s.}$,  and
\begin{align}\label{Bk}
 \|f_{\lambda,k,h}-f^{\star}\|^2_{L^2\left(\Omega;\HH_K\right)}
 \leq &\E \left[ \sum_{i=0}^{\infty}\frac{  \psi_{k,k+h-1}}{2\Lambda(i)+ \psi_{k,k+h-1}} \left|\langle f^{\star},e(i)\rangle_{\HH_K}\right|^2 \right],   \forall \ k\geq 1.
\end{align}
\end{lemma}

\vskip 0.1cm

\begin{proof}
It follows from the definition of $f_{\lambda,k,h}$ that
\begin{align}\label{xiangsile}
 \left\|f_{\lambda,k,h}-f^{\star}\right\|^2_{\HH_K}
 = & \big\|\big(\psi_{k,k+h-1}\big)\big(\phi_{k,k+h-1}  +\psi_{k,k+h-1}I\big)^{-1}f^{\star}\big\|^2_{\HH_K}.
\end{align}
As
$
\phi_{k,k+h-1}\succeq R~\mathrm{a.s.},~\forall ~k\in \mathbb N,
$
then,
\begin{align}
& \big(\phi_{k,k+h-1}+\psi_{k,k+h-1}I\big)^2\cr
 =&\big(\phi_{k,k+h-1}\big)^2+2\big(\psi_{k,k+h-1}\big)\big(\phi_{k,k+h-1}\big)  +\big(\psi_{k,k+h-1}\big)^2I\cr
   \succeq & 2\big(\psi_{k,k+h-1}\big)\big(\phi_{k,k+h-1}\big)+\big(\psi_{k,k+h-1}\big)^2I\cr
 \succeq&
\big(\psi_{k,k+h-1}\big)\big(2R+\psi_{k,k+h-1}I\big)~\mathrm{a.s.},\notag
\end{align}
$\forall ~k\in \mathbb N.$
Noting that for any given $k\in\mathbb N$, $2R+\psi_{k,k+h-1}I$ almost surely has a bounded inverse, then by Theorem 2.3 in \cite{Dehimi}, we get
\begin{align}\label{wushishui3}
& \big(\big(\psi_{k,k+h-1}\big)\big(2R+\psi_{k,k+h-1}I\big)\big)^{-1}\cr
 =& \big(\psi_{k,k+h-1}\big)^2\big(\psi_{k,k+h-1}\big(2R+\psi_{k,k+h-1}I\big)\big)^{-1}\cr
 \succeq&
\big(\psi_{k,k+h-1}\big(\psi_{k,k+h-1}I+\phi_{k,k+h-1}\big)^{-1}\big)^2 \mathrm{a.s.},
\end{align}
$\forall ~k\in \mathbb N.$ It follows from the spectral theorem of the compact operator in \cite{Rudin1991} that
$$
f^{\star}=\sum_{i=0}^{\infty}\langle f^{\star},e(i)\rangle_{\HH_K} e(i)~\mathrm{a.s.},
$$
which together with the functional calculus for self-adjoint operators and $g(x)=\frac{ \psi_{k,k+h-1}}{2x+ \psi_{k,k+h-1} }\leq 1,\ \forall \ x \geq 0$  gives
\begin{align}\label{Bj}
 &\big\langle f^{\star},\psi_{k,k+h-1}\big(2R+\psi_{k,k+h-1}I\big)^{-1}f^{\star}\big\rangle_{\HH_K}\notag\\
 =&\left\langle f^{\star}, \left(\sum_{i=0}^{\infty}\frac{ \psi_{k,k+h-1}}{2\Lambda(i)+ \psi_{k,k+h-1} }(e(i)\otimes e(i)) f^{\star} \right)\right\rangle_{\HH_K} \notag\\
 =&\left\langle f^{\star}, \left(\sum_{i=0}^{\infty}\frac{ \psi_{k,k+h-1}}{2\Lambda(i)+ \psi_{k,k+h-1} } \langle f^{\star},e(i)\rangle_{ \HH_K } e(i)\right)\right\rangle_{\HH_K} \notag\\
 = & \sum_{i=0}^{\infty}\frac{ \psi_{k,k+h-1}}{2\Lambda(i)+ \psi_{k,k+h-1} }\left|\langle f^{\star},e(i)\rangle_{\HH_K}\right|^2~\mathrm{a.s.},
\end{align}
$\forall~ k\in\mathbb N.$
By (\ref{xiangsile}), (\ref{wushishui3}) and (\ref{Bj}), we have
\begin{align}\label{wushishui4}
&\left\|f_{\lambda,k,h}-f^{\star}\right\|^2_{\HH_K}\cr
=&\big\langle f^{\star},\big(\psi_{k,k+h-1}\big(\psi_{k,k+h-1}I+\phi_{k,k+h-1}\big)^{-1}\big)^2 f^{\star} \big\rangle_{\HH_K} \cr
 \leq & \left\langle f^{\star}, \left(\psi_{k,k+h-1}\right)\left(2R+\psi_{k,k+h-1}I\right)^{-1}f^{\star}\right\rangle_{\HH_K} \cr  =&\hspace{-3pt} \sum_{i=0}^{\infty}\hspace{-3pt}\frac{ \psi_{k,k+h-1}}{2\Lambda(i)
+ \psi_{k,k+h-1}}\hspace{-3pt}\left|\langle f^{\star},e(i)\rangle_{\HH_K}\right|^2\hspace{-3pt}\mathrm{a.s.}, \forall~ k\in\mathbb N.\hspace{-3pt}
\end{align}
 By (\ref{wushishui4}), we get (\ref{Bk}).
Noting that $\Lambda(i)>0~\mathrm{a.s.},\ \forall~i\in\mathbb N$, and
$$
\frac{ \psi_{k,k+h-1}}{2\Lambda(i)+ \psi_{k,k+h-1} }\left|\langle f^{\star},e(i)\rangle_{\HH_K} \right|^2\leq \left|\langle f^{\star},e(i)\rangle_{\HH_K} \right|^2~\mathrm{a.s.},
$$
  $\sum_{i=0}^{\infty}|\langle f^{\star},e(i)\rangle_{\HH_K} |^2=\|f^{\star}\|^2_{\HH_K}<\infty~\mathrm{a.s.}$, then by Condition \ref{con1} and the dominated convergence theorem, we have
$ \lim_{k\to\infty}\E\big[\sum_{i=0}^{\infty}\frac{
 \psi_{k,k+h-1} }{2\Lambda(i)+  \psi_{k,k+h-1} }\big|\langle f^{\star},e(i)\rangle_{\HH_K} \big|^2 \big]
 =  \E\big[\sum_{i=0}^{\infty}  \lim\limits_{k\to\infty}  \frac{  \psi_{k,k+h-1}}{2\Lambda(i)+ \psi_{k,k+h-1}}\\ \left|\langle f^{\star},e(i)\rangle_{\HH_K} \right|^2\big]=0,\notag
$
which together with (\ref{Bk}) leads to
$\lim\limits_{k\to\infty}\|f_{\lambda,k,h}-f^{\star}\|_{L^2\left(\Omega;\HH_K\right)}=0.$ Similarly, by (\ref{wushishui4}), Condition \ref{con1} and the dominated convergence theorem, we have $\lim\limits_{k\to\infty} \| f_{\lambda,k,h}-f^{\star}\|_{ \HH_K } =0\ \text{a.s.}$ 
\end{proof}

 \vskip 2mm
\noindent
\textbf{Proof of Theorem \ref{wuxian}:}
 Noting that Condition \ref{con1} implies $\sum\limits_{k=0}^{\infty}a_k\lambda_k=\infty$, by (\ref{cnm}), (\ref{cn11cy}) and Lemma III.6 in \cite{Tarres}, we get $\lim_{k\to\infty}\sum_{i=0}^k \|f_{\lambda,i+1} -f_{\lambda,i} \|_{L^2\left(\Omega;\HH_K\right)} \prod_{j=i+1}^k \left(1 -a_j\lambda_j\right)=0 $
and
$
\lim\limits_{k\to\infty}\sum\limits_{i=0}^k \|f_{\lambda,i+1}   -f_{\lambda,i} \|_{\HH_K } \prod_{j=i+1}^k   (1 -a_j\lambda_j )=0,\ \text{a.s.}
$
Then, by  Assumptions \ref{ass2}-\ref{ass1}, Condition \ref{con1}   and Lemma \ref{THMM}, we obtain
\bna\label{jk2}
\lim_{k\to\infty}\left\|f_k-f_{\lambda,k}\right\|_{L^2\left(\Omega;\HH_K\right)}=0,
\ena
and
\bna\label{jk02}
\lim_{k\to\infty}\left\|f_k-f_{\lambda,k}\right\|_{ \HH_K }=0 \ \text{a.s.}
\ena
Noting that Condition \ref{con1} together with (\ref{cnmcy}) leads to
\bna\label{jk33}
\left\|f_{\lambda,k+1}-f_{\lambda,k}\right\|_{L^2\left(\Omega:\HH_K\right)}=o\left(\lambda_k\right),
\ena
and the online data streams $\{(x_k,y_k),k\in\mathbb N\}$   satisfy the RKHS persistence of excitation condition, by (\ref{jk33}),
Assumption \ref{ass2}, Condition \ref{con1} and Lemma \ref{THMM2}, we have
\bna\label{jk4}
\lim_{k\to\infty}\left\|f_{\lambda,k}-f^{\star}\right\|_{L^2\left(\Omega;\HH_K\right)}=0.
\ena
Hence, by (\ref{jk2}) and (\ref{jk4}), we have $\lim\limits_{k\to\infty}\|f_k-f^{\star}\|_{L^2(\Omega;\HH_K)}=0$.
By (\ref{cn11cy}) and Condition  \ref{con1}, we have $ \lim\limits_{k \rightarrow \infty}( \|f_{\lambda,k+j+1}-f_{\lambda, k+j}\|^2 \lambda_k^{-2} )
=  \lim\limits_{k \rightarrow \infty} (\|f_{\lambda, k+j+1} -f_{\lambda, k+j}\|^2   a_{k+j}^{-2}\lambda_{k+j}^{-2})
\lim_{k \rightarrow \infty}\frac{a_{k+j}^2\lambda_{k+j}^2}{\lambda_k^2}=0~\text{a.s.},$
and
\begin{align}
 & \mathbb{E}\left[\sup_ {k\in\mathbb{N}}\frac{\|f_{\lambda,k+j+1}-f_{\lambda,k+j}\|_{\mathscr{H}_K }^2}{\lambda_{k}^2}\right]\notag\\
 \leq & \mathbb{E}\left[\sup_ {k\in\mathbb{N}}\frac{\|f_{\lambda,k+j+1}-f_{\lambda,k+j}\|_{\mathscr{H}_K }^2}
 {\lambda_{k+j}^2}\sup_ {k\in\mathbb{N}}  \frac{\lambda_{k+j}^2}{\lambda_{k}^2}\right]\notag\\
 \leq &
 \left[\left(\sup_ {k\in\mathbb{N}}\frac{\|f_{\lambda,k+j+1}-f_{\lambda,k+j}\|_{\mathscr{H}_K}}
 {\lambda_{k+j}}\right)^2\right]<\infty, \notag
\end{align}
$ j=0,\cdots, h-2,$ where $h$ is given by \textit{RKHS persistence of excitation} condition. This together with L\'{e}vy's upward theorem (Corollary 11.1.4 in \cite{YSChow}) shows that
$
\mathbb{E}\big[ \|f_{\lambda,k+j+1}-f_{\lambda,k+j} \|_{\mathscr{H}_K}^2|\mathcal{F}_{k-1}\big]^{\frac{1}{2}}=o(\lambda_k)~\text{a.s.}, \ j=0,\cdots, h-2,$
which together with Lemma \ref{THMM2} gives
$
\lim_{k \rightarrow \infty}\|f_{\lambda,k}-f^{*}\|_{\mathscr{H}_K}=0 \ \text{a.s.}
$
Then, by (\ref{jk02}) and triangle inequality, we have   $\lim_{k\to\infty}\|f_k-f^{\star}\|_{\HH_K}=0$ a.s.
\hfill $\blacksquare$

\vskip 2mm
\noindent
\textbf{Proof of Corollary \ref{wuxiannsuanfasud}:} By Proposition \ref{mingti100}, Lemmas \ref{LLEEMM6}-\ref{LLEEMM7} and Minkovski inequality, we have
\begin{align}
&\left\|f_k-f_{\lambda,k}\right\|_{L^2\left(\Omega;\HH_K\right)}\notag\\
\leq & \left\|M_k\right\|_{L^2\left(\Omega;\HH_K\right)}+ \left\|D_k\right\|_{L^2\left(\Omega;\HH_K\right)}\notag\\
 \leq & C_6 \left(\frac{\ln^{\frac{3}{2}}(k+1)}{k^{\frac{\tau_1-3\tau_2}{2}}}\right),\ \forall \ k\geq k_0.\label{sudu1}
\end{align}
  By Lemmas \ref{lemma222}-\ref{lemma333} and Minkovski inequality, we have
\begin{align}
&\left\|f_{\lambda,k}-f^*\right\|_{L^2\left(\Omega;\HH_K\right)}\notag\\
\leq & \left\|f_{\lambda,k}-f_{\lambda,k,h}\right\|_{L^2\left(\Omega;\HH_K\right)}+\left\|f_{\lambda,k,h}-f^*\right\|_{L^2\left(\Omega;\HH_K\right)}\notag\\
\leq &  2 \hat{C}_5  \alpha_1 (\kappa+\alpha_{2})\left(\frac{\ln^{\frac{3}{2}}(k+1)}{k^{\frac{\tau_1-3\tau_2}{2}}}\right) \notag\\
&+\left(\hspace{-3pt}\E\left[\sum_{i=0}^{\infty}\frac{  \sum_{j=k }^{k+h-1}\lambda_j}{2\Lambda(i)+\sum_{j=k }^{k+h-1}\lambda_j}\left|\langle f^{\star},e(i)\rangle_{\HH_K}\right|^2 \right]\right)^{\frac{1}{2}},\ \forall \ k\geq k_0. \label{sudu2}
\end{align}
By (\ref{sudu1}), (\ref{sudu2}) and Minkovski inequality, we have
\begin{align}
&\left\|f_k-f^*\right\|_{L^2\left(\Omega;\HH_K\right)}\notag\\
\leq & \left\|f_k-f_{\lambda,k}\right\|_{L^2\left(\Omega;\HH_K\right)}+ \left\| f_{\lambda,k}-f^*\right\|_{L^2\left(\Omega;\HH_K\right)} \notag\\
\leq &   C_7  \left(\frac{\ln^{\frac{3}{2}}(k+1)}{k^{\frac{\tau_1-3\tau_2}{2}}}\right)+ \left(\hspace{-3pt}\E\left[\sum_{i=0}^{\infty}\frac{  \sum_{j=k }^{k+h-1}\lambda_j}{2\Lambda(i)+\sum_{j=k }^{k+h-1}\lambda_j}\left|\langle f^{\star},e(i)\rangle_{\HH_K}\right|^2 \right]\right)^{\frac{1}{2}}, \ \forall \ k\geq k_0,\notag
\end{align}
which shows (\ref{sudumean}).
\hfill $\blacksquare$

\vskip 2mm
\noindent
\textbf{Proof of Corollary \ref{tuijil}:} Note that $  \sum_{i=k}^{k+h-1}\E\left[\left. H_{i}\right|\F_{k-1}\right]
 = \sum_{i=k}^{k+h-1}\E\left[\left.\E\left[\left. H_{i}\right|\F_{i-1}\right]\right| \F_{k-1}\right]
 = \E\big[ \sum_{i=k}^{k+h-1}T_i |\F_{k-1}\big].$
Then by $\sum_{i=k}^{k+h-1}T_i\succeq R$ a.s. and Theorem \ref{wuxian}, we get the conclusion of this corollary.
\hfill $\blacksquare$

\vskip 2mm

\noindent
\textbf{Proof of Proposition  \ref{mixing}:}
Noting that $\lim_{k\to \infty} \phi(k)=0,$ then there exists $M_0>0$, such that $\phi(i)<\frac{m}{2 h_{0} \sup_{x \in \X}K(x,x)},\ \forall \ i\geq M_0$, where  $m=\inf_{\{f\in  \HH_K|\|f\|_{\mathscr{H}_K }=1 \}}\langle  R_{0} f, f\rangle_{\mathscr{H}_K } >0$. By the reproducing  property of RKHS, for any $f\in \HH_K,\ f \neq 0$, $\langle \E\left[ H_{i} \right]f,f\rangle_{\mathscr{H}_K }= \E\left[ f^{2}(x_i)   \right]\geq 0,\langle \E\left[\left. H_{i}\right|\F_{k-1}\right]f,f\rangle_{\mathscr{H}_K }= \E\left[\left. f^{2}(x_i) \right|\F_{k-1} \right]\geq 0, \ \forall \ i\geq k-1, \ k \in \mathbb{N}.$ Let $h=M_0+h_0$. Then, we have, for any  $f\in \HH_K,\ f \neq 0$,
\begin{align}
&\left\langle \sum_{i=k}^{k+h-1}\E\left[\left. H_{i}\right|\F_{k-1}\right]f, f\right\rangle_{\mathscr{H}_K } \notag\\
\geq & \left\langle \sum_{i=k+M_0}^{k+h-1}\E\left[\left. H_{i}\right|\F_{k-1}\right]f, f\right\rangle_{\mathscr{H}_K } \notag\\
=& \left\langle \sum_{i=k+M_0}^{k+h-1}\E\left[\left. H_{i}\right|\F_{k-1}\right]f, f\right\rangle_{\mathscr{H}_K }- \left\langle \sum_{i=k+M_0}^{k+h-1}\E\left[  H_{i} \right]f, f\right\rangle_{\mathscr{H}_K } + \left\langle \sum_{i=k+M_0}^{k+h-1}\E\left[  H_{i} \right]f, f\right\rangle_{\mathscr{H}_K }\notag\\
\geq &\left\langle \sum_{i=k+M_0}^{k+h-1}\E\left[\left. H_{i}\right|\F_{k-1}\right]f, f\right\rangle_{\mathscr{H}_K }- \left\langle \sum_{i=k+M_0}^{k+h-1}\E\left[  H_{i} \right]f, f\right\rangle_{\mathscr{H}_K } +  \left\langle  R_{0} f, f\right\rangle_{\mathscr{H}_K }.\label{mix1}
\end{align}
By Assumption \ref{ass2}, we know that,  $\frac{f^2(x_{i})}{\|f\|_{\mathscr{H}_K }^2\sup_{x \in \X}K(x,x)}\leq \frac{\|f\|_{\mathscr{H}_K }^2\|K_{x_{i}}\|_{\mathscr{H}_K }^2}{\|f\|_{\mathscr{H}_K }^2\sup_{x \in \X}K(x,x)}=\frac{  K(x_{i},x_{i}) }{ \sup_{x \in \X}K(x,x)}\\ \leq 1,\ \forall \ i \in \mathbb{N}$, which together with Theorem 1.3.12 in \cite{Guo2020} shows that
\begin{align}
&\left|\left\langle \sum_{i=k+M_0}^{k+h-1}\E\left[\left. H_{i}\right|\F_{k-1}\right]f, f\right\rangle_{\mathscr{H}_K }- \left\langle \sum_{i=k+M_0}^{k+h-1}\E\left[  H_{i} \right]f, f\right\rangle_{\mathscr{H}_K } \right|\notag\\
= & \Bigg| \|f\|_{\mathscr{H}_K }^2\sup_{x \in \X}K(x,x) \bigg(   \sum_{i=k+M_0}^{k+h-1}\E\left[\frac{f^{2}(x_i)}{ \|f\|_{\mathscr{H}_K }^2\sup_{x \in \X}K(x,x)} \bigg|\F_{k-1}\right] \notag\\
&-  \sum_{i=k+M_0}^{k+h-1}\E\left[  \frac{f^{2}(x_i)}{ \|f\|_{\mathscr{H}_K }^2\sup_{x \in \X}K(x,x)}  \right]\bigg) \Bigg|\notag\\
\leq & \|f\|_{\mathscr{H}_K }^2\sup_{x \in \X}K(x,x)  \sum_{i=k+M_0}^{k+h-1} \phi(i)
 \leq  \frac{m}{2} \|f\|_{\mathscr{H}_K }^2.\notag
\end{align}
This together with (\ref{mix1}) gives
\begin{align}
&\left\langle \sum_{i=k}^{k+h-1}\E\left[\left. H_{i}\right|\F_{k-1}\right]f, f\right\rangle_{\mathscr{H}_K } \notag\\
\geq &- \frac{m}{2} \|f\|_{\mathscr{H}_K }^2 +  \left\langle  R_{0} f, f\right\rangle_{\mathscr{H}_K } \geq \left\langle \frac{ R_{0}}{2} f, f\right\rangle_{\mathscr{H}_K },\  k \in \mathbb{N},  \notag
\end{align}
that is, $\{(x_k,y_k),k\in \mathbb N\}$ satisfies the RKHS persistence of excitation condition.
Furthermore, if  $\{x_k,\ k\in \mathbb N\}$ is   $M$-dependent, then we have $\phi(i)=0,\ \forall \ i>M$.  Similar to the above proof,  we know that $\{(x_k,y_k),k\in \mathbb N\}$ satisfies the RKHS persistence of excitation condition with $h=h_0+M$ and $R=\frac{R_{0}}{2}$.
\hfill $\blacksquare$

%
%

 \section{Proofs in Section V}\label{fulu1ccc}
\setcounter{equation}{0}
\def\theequation{C.\arabic{equation}}
\setcounter{lemma}{0}
\def\thelemma{C.\arabic{lemma}}
\setcounter{remark}{0}
\def\theremark{C.\arabic{remark}}
\setcounter{proposition}{0}
\def\theproposition{C.\arabic{proposition}}
\noindent
\textbf{Proof of Proposition \ref{mingti7}:} Since the online data streams $\{(x_k,y_k),k\in\mathbb N\}$ are independently sampled from the product probability space $\prod_{k=0}^{\infty}(\X\times \mathscr Y, \rho^{(k)})$, then $\sigma(x_k,y_k)$ is independent of $\F_{k-1}$, $\forall ~k\in\mathbb N$. Noting that $H_{k}\in \sigma(x_k,y_k)$, by the definition of the conditional expectation of the random elements with values in the Banach space, we get
\begin{align}
& \int_{A}T_{k}\,\mathrm{d}\mathbb P\cr
 =&\int_{A}H_{k}\,\mathrm{d}\mathbb P\cr
 =&\int_{\Omega}\left(H_{k}\right)\mathbf{1}_A\,\mathrm{d}\mathbb P\cr
 =&\left(\int_{\Omega}H_{k}\,\mathrm{d}\mathbb P\right)\left(\int_{\Omega}\mathbf{1}_A\,\mathrm{d}\mathbb P\right)\cr
 =&\mathbb P(A)\int_{\Omega}H_{k}\,\mathrm{d}\mathbb P\cr
 =&\int_{A}\E\left[H_{k}\right]\,\mathrm{d}\mathbb P~\mathrm{a.s.},~\forall ~A\in \F_{k-1},~\forall~ k\in\mathbb N,\notag
\end{align}
where $\mathbf{1}_{A}$ is the indicator function of the set $A$, from which we know that
\bna\label{m11}
T_{k}=\E\left[H_{k}\right]~\mathrm{a.s.},~\forall~ k\in\mathbb N.
\ena
Noting that $\rho^{(k)}$ is the probability of the observation data $(x_k,y_k)$, by Assumption \ref{ass2} and Fubini theorem, we have
\begin{align}
  \E\left[H_{k}\right]
 =&\int_{\Omega}H_{k}\,\mathrm{d}\mathbb P\cr
 =&\int_{\X\times \mathscr Y}K_x\otimes K_x\,\mathrm{d}\left(\mathbb P\circ (x_k,y_k)^{-1}\right)\cr
 =&\int_{\X\times \mathscr Y}K_x\otimes K_x\,\mathrm{d}\rho^{(k)}\cr
 =&\int_{\X}\left(\int_{\mathscr Y}K_x\otimes K_x\,\mathrm{d}\rho^{(k)}_{\mathscr Y|x}\right)\,\mathrm{d}\rho^{(k)}_{\X}\cr
 =&\int_{\X}K_x\otimes K_x\,\mathrm{d}\rho^{(k)}_{\X},~\forall~ k\in\mathbb N,\notag
\end{align}
where $\rho^{(k)}_{\mathscr Y|x}$ is the conditional probability measure on the sample space $\mathscr Y$ with respect to $x\in\X$. Thus, combining the above and (\ref{m11}) gives
\begin{align}\label{m22}
&\E\Bigg[ \sum_{i=k}^{k+h-1} K_{x_i} \otimes K_{x_i}\bigg|\F_{k-1}\Bigg]
=   \int_{\X} K_x\otimes K_x\,\mathrm{d}\bigg(\sum_{i=k }^{k+h-1}\rho^{(i)}_{\X}\bigg).
\end{align}
On one hand, it follows from (\ref{cedu}) and the reproducing property of RKHS that
\begin{align}
& \left\langle\left[\int_{\X}K_x\otimes K_x\,\mathrm{d}\left(\sum_{i=k}^{k+h-1}\rho^{(i)}_{\X}\right)\right]f,f\right\rangle_{\HH_K} \cr
 =&\int_{\X}\left\langle \left(K_x\otimes K_x\right) f,f\right\rangle_{\HH_K}\,\mathrm{d}\left(\sum_{i=k }^{k+h-1}\rho^{(i)}_{\X}\right)\cr
=&\int_{\X}f(x)\left\langle K_x,f\right\rangle_{\HH_K}\,\mathrm{d}\left(\sum_{i=k }^{k+h-1}\rho^{(i)}_{\X}\right)\cr
=&\int_{\X}f^2(x)\,\mathrm{d}\left(\sum_{i=k }^{k+h-1}\rho^{(i)}_{\X}\right)\cr
 \ge& h\int_{\X}f^2(x)\,\mathrm{d}\gamma\cr
=&h\int_{\X}\left\langle \left(K_x\otimes K_x\right) f,f\right\rangle_{\HH_K} \,\mathrm{d}\gamma\cr
=& h\left\langle\left[\int_{\X}K_x\otimes K_x\,\mathrm{d}\gamma\right]f,f\right\rangle_{\HH_K},~\forall~ f\in\HH_K,~\forall~ k\in\mathbb N,\notag
\end{align}
which leads to
\bna\label{m33}
\int_{\X}K_x\otimes K_x\,\mathrm{d}\bigg(\sum_{i=k+1}^{k+h}\rho^{(i)}_{\X}\bigg)\succeq h\int_{\X}K_x\otimes K_x\,\mathrm{d}\gamma,
\ena
$~\forall~ k\in\mathbb N.$
On the other hand, for any given non-zero element $f\in\HH_K$, there exists $w\in \X$, such that $f^2(w)>0$. If $\int_{\X}f^2(x)\,\mathrm{d}\gamma=0$, then it follows from the measurability of $f$ that $\gamma(\{x\in\X|f^2(x)>0\})=0$. Noting that $\HH_K\subseteq C(\X)$, then there exists a neighborhood $U_w\subseteq \X$ of $w$, such that $f^2(x)>0$, $\forall~x\in U_w$, thus we have $\gamma(U_w)=0$, which is contradictory to the fact that $\gamma$ is a strictly positive measure. Hence, for any given non-zero element $f\in\HH_K$, we have
$$
\int_{\X}f^2(x)\,\mathrm{d}\gamma>0.
$$
Then, for any given non-zero element $f\in\HH_K$, by the reproducing property of RKHS, we get
\ban
&&~~~~\left\langle \left(\int_{\X}K_x\otimes K_x\,\mathrm{d}\gamma\right)f,f\right\rangle_{\HH_K}\cr
&&=\int_{\X}\left\langle \left(K_x\otimes K_x\right) f,f\right\rangle_{\HH_K} \,\mathrm{d}\gamma\cr
&&=\int_{\X}f(x)\left\langle K_x,f\right\rangle_{\HH_K} \,\mathrm{d}\gamma =\int_{\X}f^2(x)\,\mathrm{d}\gamma
 >0.
\ean
Denote $R=h\int_{\X}K_x\otimes K_x\,\mathrm{d}\gamma$. Since $\gamma$ is the strictly positive Borel measure, then $R$ is a compact operator (\cite{Smale3}), which together with the above inequality shows that $R$ is a strictly positive compact operator. Then (\ref{m33}) implies
\ban
\E\left[\sum_{i=k }^{k+h-1}H_{i}\bigg|\F_{k-1}\right]\succeq R,~\forall~ k\in\mathbb N.
\ean
Noting that Assumption \ref{ass2} ensures that $R\in L^2(\Omega;\mathscr L(\HH_K))$, it follows from Definition \ref{pe} that the online data streams satisfy the RKHS persistence of excitation condition.
\hfill $\blacksquare$

\vskip 2mm

\noindent
\textbf{Proof of Corollary \ref{corollary1}:} Since the online data streams $\{(x_k,y_k),k\in\mathbb N\}$ are independently sampled from the product probability space $\prod_{k=0}^{\infty}(\X\times \mathscr Y, \rho^{(k)})$, then $\sigma(x_k,y_k)$ is independent of $\F_{k-1}$, $\forall~ k\in\mathbb N$. Noting that $H_{k}\in \sigma(x_k,y_k)$, by the definition of the conditional expectation of the random elements with values in the Banach space, we get
\begin{align}
& \int_{A}T_{k}\,\mathrm{d}\mathbb P\cr
 =&\int_{A}H_{k}\,\mathrm{d}\mathbb P\cr
=&\int_{\Omega}\left(H_{k}\right)\mathbf{1}_A\,\mathrm{d}\mathbb P\cr
=&\left(\int_{\Omega}H_{k}\,\mathrm{d}\mathbb P\right)\left(\int_{\Omega}\mathbf{1}_A\,\mathrm{d}\mathbb P\right)\cr
=&\mathbb P(A)\int_{\Omega}H_{k}\,\mathrm{d}\mathbb P\cr
=&\int_{A}\E\left[H_{k}\right]\,\mathrm{d}\mathbb P~\mathrm{a.s.},~\forall ~A\in \F_{k-1},~\forall ~k\in\mathbb N,\notag
\end{align}
where $\mathbf{1}_{A}$ is the indicator function of the set $A$, from which we have
\bna\label{mk11}
T_{k}=\E\left[H_{k}\right]~\mathrm{a.s.},~\forall~ k\in\mathbb N.
\ena
Noting that $\rho^{(k)}$ is the probability measure of the observation data $(x_k,y_k)$, by (\ref{mk11}), Assumption \ref{ass3} and Fubini theorem, we obtain
\begin{align}
  T_{k}
 =&\int_{\Omega}H_{k}\,\mathrm{d}\mathbb P\cr
 =&\int_{\X\times \mathscr Y}K_x\otimes K_x\,\mathrm{d}\left(\mathbb P\circ (x_k,y_k)^{-1}\right)\cr
 =&\int_{\X\times \mathscr Y}K_x\otimes K_x\,\mathrm{d}\rho^{(k)}\cr
 =&\int_{\X}\left(\int_{\mathscr Y}K_x\otimes K_x\,\mathrm{d}\rho^{(k)}_{\mathscr Y|x}\right)\,\mathrm{d}\rho^{(k)}_{\X}\cr
 =&\int_{\X}K_x\otimes K_x\,\mathrm{d}\rho^{(k)}_{\X},~\forall~ k\in\mathbb N,\notag
\end{align}
where $\rho^{(k)}_{\mathscr Y|x}$ is the conditional probability measure on the sample space $\mathscr Y$ with respect to $x\in\X$. For any given $f\in \HH_K$, noting that $\int_{\X}K_x\otimes K_x\,\mathrm{d}(\rho^{(k+1)}_{\X}-\rho^{(k)}_{\X})\in \mathscr L(\HH_K)$, by the reproducing property of RKHS, we have
\begin{align}\label{mm1}
&\left\|\left(T_{k+1}-T_{k}\right)f\right\|^2_{\HH_K}\cr
=& \bigg\|\bigg(\int_{\X}K_x\otimes K_x\,\mathrm{d}\rho^{(k+1)}_{\X}-\int_{\X}K_x\otimes K_x\,\mathrm{d}\rho^{(k)}_{\X}\bigg)f\bigg\|^2_{\HH_K}\cr
 =&\int_{\X}f(y)\left(\int_{\X}f(x)K(y,x)\,\mathrm{d}\Delta_k(x)\right) \,\mathrm{d}\Delta_k(y),
\end{align}
$~\forall~ k\in\mathbb N,$
where $\Delta_k=\rho^{(k+1)}_{\X}-\rho^{(k)}_{\X}\in \mathcal M(\X)$. Since $C^s(\X)\subseteq C(\X)$ and $(C(\X))^*=\mathcal M(\X)$, then $\mathcal M(\X)\subseteq (C^s(\X))^*$, from which we have $\Delta_k\in (C^s(\X))^*$. Denote
\bna\label{mm2}
g_k(\cdot)=f(\cdot) \bigg(\int_{\X}f(x)K(\cdot,x)\,\mathrm{d}\Delta_k(x)\bigg),~\forall~ k\in\mathbb N.
\ena
Noting that $\Delta_k\in (C^s(\X))^*$ and by the definition of $(C^s(\X))^*$, we know that
\begin{align}
\label{mm3}
 \hspace{-12pt} \int_{\X}\hspace{-5pt} g_k(y)\,\mathrm{d}\Delta_k(y)\hspace{-4pt}
\leq &  \left\|\Delta_k\right\|_{(C^s(\X))^*}\big(\hspace{-2pt} \left\|g_k
\right\|_{\infty}\hspace{-3pt}
+\left|g_k\right|_{C^s(\X)}\big).\hspace{-6pt}
\end{align}
We now estimate $\|g_k\|_{\infty}$ and $|g_k|_{C^s(\X)}$, respectively.

It follows from Assumption \ref{ass3} that $K\in C^s(\X\times \X)\subseteq C(\X\times \X)$, which shows that there exists a constant $\kappa<\infty$, such that $\kappa=\sup_{x\in\X}\sqrt{K(x,x)}$. It follows from \cite{Smale5} that $\|g\|_{\infty}\leq\kappa\|g\|_{\HH_K}$ and $\|g\|_{C^s(\X)}\leq (\kappa+\tau_s)\|g\|_{\HH_K}$, $\forall ~ g\in\HH_K$.
By Lemma \ref{lemma10086} and the reproducing property of RKHS, we get
\ban
&&~~~\left\|fK_y\right\|_{C^s(\X)}\cr
&&=\left\|fK_y\right\|_{\infty}+\left|fK_y\right|_{C^s(\X)}\cr
&&\leq \left\|f\right\|_{\infty}\left\|K_y\right\|_{\infty}+\left|f\right|_{C^s(\X)}\left\|K_y\right\|_{\infty}+\left\|f\right\|_{\infty}\left|K_y\right|_{C^s(\X)}\cr
&&\leq \kappa\left\|f\right\|_{\infty}\left\|K_y\right\|_{\HH_K}+\kappa\left|f\right|_{C^s(\X)}\left\|K_y\right\|_{\HH_K}+\left\|f\right\|_{C^s(\X)}\left\|K_y\right\|_{C^s(\X)}\cr
&&\leq \kappa\sup_{y\in\X}\sqrt{K(y,y)}\left\|f\right\|_{\infty}+\kappa\sup_{y\in\X}\sqrt{K(y,y)}\left|f\right|_{C^s(\X)}+\left(\kappa+\tau_s\right)\left\|f\right\|_{C^s(\X)}\left\|K_y\right\|_{\HH_K}\cr
&&\leq \kappa^2 \left(\left\|f\right\|_{\infty}+\left|f\right|_{C^s(\X)}\right)+\left(\kappa+\tau_s\right)\sup_{y\in\X}\sqrt{K(y,y)}\left\|f\right\|_{C^s(\X)}\cr
&&=\left(2\kappa^2 +\kappa\tau_s\right)\left\|f\right\|_{C^s(\X)},~\forall~ y\in \X,
\ean
which shows that
\begin{align}\label{mm4}
&\Big|\int_{\X}f(x)K(y,x)\,\mathrm{d}\Delta_k(x)\Big|\cr
\leq & \left\|\Delta_k\right\|_{(C^s(\X))^*}\left\|fK_y\right\|_{C^s(\X)}\cr
 \leq & \left(2\kappa^2  +\kappa\tau_s\right)
\left\|\Delta_k\right\|_{(C^s(\X))^*}
\left\|f\right\|_{C^s(\X)},
\end{align}
$\forall ~y\in \X,~\forall~ k\in\mathbb N.$ Thus, it follows from (\ref{mm2}) and (\ref{mm4}) that
\begin{align}\label{mm5}
\left\|g_k\right\|_{\infty}\leq &  \left\|f\right\|_{\infty}\sup_{y\in\X}
\Big|\int_{\X}f(x)K(y,x)\,\mathrm{d}\Delta_k(x)\Big|\notag\\
\leq & \left(2\kappa^2 +\kappa\tau_s\right)\left\|
f\right\|^2_{C^s(\X)}\left\|\Delta_k\right\|_{(C^s(\X))^*}.
\end{align}
By Lemma \ref{lemma10086} and (\ref{mm4}), we obtain
\begin{align}\label{mm6}
&\left|g_k\right|_{C^s(\X)}\cr
 \leq & \left|f\right|_{C^s(\X)}\Big\|\int_{\X}f(x)K_x\,
 \mathrm{d}\Delta_k(x)\Big\|_{\infty} +\left\|f\right\|_{\infty}\Big|\int_{\X}f(x)K_x\,
 \mathrm{d}\Delta_k(x)\Big|_{C^s(\X)}\cr
 =&\left|f\right|_{C^s(\X)}\sup_{y\in \X}\Big|\int_{\X}f(x)K(y,x)\,\mathrm{d}\Delta_k(x)\Big| +\left\|f\right\|_{\infty}\Big|\int_{\X}f(x)K_x\,\mathrm{d}\Delta_k(x)\Big|_{C^s(\X)}\cr
 \leq & \left(2\kappa^2 +\kappa\tau_s\right)\left\|\Delta_k
 \right\|_{(C^s(\X))^*}\left\|f\right\|^2_{C^s(\X)} +\left\|f\right\|_{C^s(\X)}\Big|\int_{\X}f(x)K_x\,\mathrm{d}\Delta_k(x)\Big|_{C^s(\X)}.
\end{align}
By the definition of $(C^s(\X))^*$, we get
\begin{align}\label{mm7}
 \Big|\int_{\X}f(x)K_x\,\mathrm{d}\Delta_k(x)\Big|_{C^s(\X)}
 \leq & \left\|\Delta_k\right\|_{(C^s(\X))^*}\sup_{z_1\neq z_2\in\X}\Big\|f\frac{K_{z_1}-K_{z_2}}{\left\|z_1-z_2
 \right\|^s}\Big\|_{C^s(\X)},
\end{align}
$~\forall~ k\in\mathbb N.$ By the definition of $\|\cdot\|_{C^s(\X)}$ and Assumption \ref{ass3}, we have
\begin{align}\label{mm8}
& \Big\|f\frac{K_{z_1}-K_{z_2}}{\left\|z_1-z_2\right\|^s}\Big\|_{C^s(\X)}\cr
 \leq & \left\|f\right\|_{\infty}\Big\|\frac{K_{z_1}-K_{z_2}}{\left\|z_1-z_2\right\|^s}\Big\|_{\infty}
 +\Big|f\frac{K_{z_1}-K_{z_2}}{\left\|z_1-z_2\right\|^s}\Big|_{C^s(\X)}\cr
 \leq & \left\|f\right\|_{C^s(\X)}\sup_{(z_1,x)\neq (z_2,x)\in \X\times \X}\frac{\left|K(z_1,x)-K(z_2,x)\right|}
 {\left\|z_1-z_2\right\|^s} +\Big|f\frac{K_{z_1}-K_{z_2}}{\left\|z_1-z_2\right\|^s}\Big|_{C^s(\X)}\cr
 \leq& \left|K\right|_{C^s(\X\times \X)}\left\|f\right\|_{C^s(\X)}+\Big|f\frac{K_{z_1}
 -K_{z_2}}{\left\|z_1-z_2\right\|^s}\Big|_{C^s(\X)},
\end{align}
$\forall~ z_1\neq z_2\in \X.$
It follows from Lemma \ref{lemma10086} and Assumption \ref{ass3} that
\ban
&&~~~\left|f\frac{K_{z_1}-K_{z_2}}{\left\|z_1-z_2\right\|^s}\right|_{C^s(\X)}\cr
&&\leq\left|f\right|_{C^s(\X)}\left\|\frac{K_{z_1}-K_{z_2}}{\left\|z_1-z_2\right\|^s}\right\|_{\infty}+\left\|f\right\|_{\infty}\left|\frac{K_{z_1}-K_{z_2}}{\left\|z_1-z_2\right\|^s}\right|_{C^s(\X)}\cr
&&\leq \left|f\right|_{C^s(\X)}\sup_{(z_1,x)\neq (z_2,x)\in \X\times \X}\frac{\left|K(z_1,x)-K(z_2,x)\right|}{\left\|z_1-z_2\right\|^s}+\left\|f\right\|_{C^s(\X)}\left|\frac{K_{z_1}-K_{z_2}}
{\left\|z_1-z_2\right\|^s}\right|_{C^s(\X)}\cr
&&\leq \left\|f\right\|_{C^s(\X)}\bigg(\left|K\right|_{C^s(\X\times \X)}  +\sup_{w_1\neq w_2\in\X}\frac{\left|K(z_1,w_1)-K(z_2,w_1)-K(z_1,w_2)+K(z_2,w_2)\right|}{\left\|z_1
-z_2\right\|^s\left\|w_1-w_2\right\|^s}\bigg)\cr
&&\leq \left\|f\right\|_{C^s(\X)}\left(\left|K\right|_{C^s(\X\times \X)}+\tau_s\right),~\forall~ z_1\neq z_2\in \X.
\ean
 Thus, by the above and (\ref{mm8}), we get
\begin{align}\label{mm9}
&\Big\|f\frac{K_{z_1}-K_{z_2}}{\left\|z_1-z_2\right\|^s}
\Big\|_{C^s(\X)}
\leq   \left\|f\right\|_{C^s(\X)} (2\left|K\right|_{C^s(\X\times \X)}+\tau_s ).
\end{align}
Putting (\ref{mm8})-(\ref{mm9}) into (\ref{mm7}) leads to
\ban
\left|\int_{\X}f(x)K_x\,\mathrm{d}\Delta_k(x)\right|_{C^s(\X)}\leq \left\|\Delta_k\right\|_{(C^s(\X))^*}\left\|f\right\|_{C^s(\X)}\left(2\left|K\right|_{C^s(\X\times \X)}+\tau_s\right),
\ean
$~\forall~ k\in\mathbb N,$ which together with (\ref{mm6}) gives
\begin{align}\label{mm10}
 \left|g_k\right|_{C^s(\X)}
\leq & \left\|\Delta_k\right\|_{(C^s(\X))^*}\left\|f\right\|^2_{C^s(\X)} (2\kappa^2 +\kappa\tau_s +2\left|K\right|_{C^s(\X\times \X)}+\tau_s ),
\end{align}
$~\forall~ k\in\mathbb N.$ Putting (\ref{mm2})-(\ref{mm3}), (\ref{mm5}) and (\ref{mm10}) into (\ref{mm1}) shows
\begin{align}
& \left\|\left(T_{k+1}-T_{k}\right)f\right\|^2_{\HH_K}\notag\\
 \leq  &
%
\left\|\Delta_k\right\|^2_{(C^s(\X))^*}\left\|f\right\|^2_{K}\left(\kappa+\tau_s\right)^2 (4\kappa^2+2\kappa\tau_s +2\left|K\right|_{C^s(\X\times \X)}+\tau_s ),~\forall~ f\in\HH_K,~\forall~ k\in\mathbb N,\notag
\end{align}
from which we have
\begin{align}
& \left\|T_{k+1}-T_{k}\right\|_{\mathscr L(\HH_K)}\cr
 \leq &
%
\left\|\rho^{(k+1)}_{\X}-\rho^{(k)}_{\X}\right\|_{(C^s(\X))^*}\left(\kappa+\tau_s\right)(4\kappa^2+2\kappa\tau_s +2\left|K\right|_{C^s(\X\times \X)}+\tau_s)^{\frac{1}{2}},~\forall~ k\in\mathbb N.\notag
\end{align}
By the above inequality and (\ref{over1}), we know that there exists a constant $C_1>0$, such that
\bna\label{mm11}
\left\|T_{k+1}-T_{k}\right\|_{\mathscr L(\HH_K)}\leq C_1a_k\lambda^2_k\quad\mathrm{a.s.}
\ena
Noting that $\lim_{x\to 0}\frac{1-(1-x)^a}{x}=a$, $\forall ~a\in\mathbb R$, by Condition \ref{con1}, we obtain
\bna\label{mm12}
\lim_{k\to\infty}(\lambda_k-\lambda_{k+1})(a_k\lambda^2_k)^{-1}
=0.
\ena
It follows from Assumption \ref{ass3} that $K\in C(\X\times \X)$, which shows that Assumption \ref{ass2} holds. Hence, by Lemma \ref{lemma6}, (\ref{mm11})-(\ref{mm12}), we know that there exists a constant $C_2>0$, such that
\bna\label{mm13}
\left\|f_{\lambda,k+1}-f_{\lambda,k}\right\|_{\HH_K}\leq C_2a_k\lambda_k\left\|f_{\lambda,k}-f^{\star}\right\|_{\HH_K}~\mathrm{a.s.}
\ena
It follows from the definition of the random Tikhonov regularization path $f_{\lambda,k}$ of $f^{\star}$ that $\|f_{\lambda,k}\|_{\HH_K}\leq \|f^{\star}\|_{\HH_K}~\mathrm{a.s.}$, then we have $\|f_{\lambda,k}-f^{\star}\|_{\HH_K}\leq 2\|f^{\star}\|_{\HH_K}<\infty~\mathrm{a.s.},$ which together with (\ref{mm13}) gives
$\left\|f_{\lambda,k+1}-f_{\lambda,k}\right\|_{\HH_K}\leq 2C_2\lambda_k a_k\|f^{\star}\|_{\HH_K}~\mathrm{a.s.}$
By Condition \ref{con1} and the above inequality, we have
$
\sup_{k\in \mathbb{N}} \left\|f_{\lambda,k+1}-f_{\lambda,k}\right\|_{\HH_K} \lambda_k^{-1} \leq 2 \alpha_1 C_2 \|f^{\star}\|_{\HH_K} \ \mathrm{a.s.}
$
and
$$
\lim_{k\to\infty} \left\|f_{\lambda,k+1}-f_{\lambda,k}\right\|_{\HH_K} \lambda_k^{-1} =0 \ \mathrm{a.s.}
$$
This   together with the dominated convergence theorem gives
\bna\label{mm14}
\left\|f_{\lambda,k+1}-f_{\lambda,k}\right\|_{L^2\left(\Omega;\HH_K\right)}=o\left(\lambda_k\right).
\ena
It follows from Assumption  \ref{ass1}, Assumption \ref{ass3}, Condition \ref{con1}, (\ref{cedu111}) and Proposition \ref{mingti7} that the online data streams $\{(x_k,y_k),k\in\mathbb N\}$ satisfy the RKHS persistence of excitation condition. Then by (\ref{mm14}) and Lemma \ref{THMM2}, we get
\bna\label{mm15}
\lim_{k\to\infty}\left\|f_{\lambda,k}-f^{\star}\right\|_{L^2\left(\Omega;\HH_K\right)}=0.
\ena
Combining (\ref{mm13}) with (\ref{mm15}) leads to
\bna\label{mm16}
\left\|f_{\lambda,k+1}-f_{\lambda,k}\right\|_{L^2\left(\Omega;\HH_K\right)}=o\left(a_k\lambda_k\right).
\ena
Noting that the online data streams $\{(x_k,y_k),k\in\mathbb N\}$ satisfy the RKHS persistence of excitation condition, by (\ref{mm16}) and Theorem \ref{wuxian}, we have
\begin{align}
\lim_{k\to\infty}\left\|f_k-f^{\star}\right\|_{L^2\left(\Omega;\HH_K\right)}=0.\label{fanshusl}
\end{align}
By Cauchy-Schwartz inequality and the reproducing property of RKHS, we have, for any $x \in \X,$
\begin{align}\label{dddd}
&\mathbb{E}\left[|f_k(x)-f^{\star}(x)|^2\right] \notag\\
=& \mathbb{E}\left[ (\langle f_k-f^{\star}, K_{x} \rangle)^2\right] \notag\\
\leq & \mathbb{E}\left[ \|f_k-f^{\star}\|_{\HH_K}^2 \|K_{x}\|_{\HH_K}^2\right] \notag\\
=& \mathbb{E}\left[ \|f_k-f^{\star}\|_{\HH_K}^2 \langle K_{x}, K_{x} \rangle \right] \notag\\
=&\left\|f_k-f^{\star}\right\|_{L^2\left(\Omega;\HH_K\right)}^2 K(x,x).\notag
\end{align}
This together with (\ref{fanshusl}) gives $\lim_{k\to\infty}\mathbb{E}\left[|f_k(x)-f^{\star}(x)|^2\right]=0, \ \forall \ x\in \X$.
By $$\left\|f_{\lambda,k+1}-f_{\lambda,k}\right\|_{\HH_K}\leq 2C_2\lambda_k a_k\|f^{\star}\|_{\HH_K}~\mathrm{a.s.},$$  we have
\begin{align}
\mathbb{E}\left[ \frac{ \left\|f_{\lambda,k+j+1}-f_{\lambda,k+j}\right\|_{\mathscr{H}_K}^2}{\lambda_{k}^{2}}\bigg| \mathcal{F}_{k-1}\right]\leq \frac{ 4C_2^{2} a_{k+j}^{2}\lambda_{k+j}^{2}}{\lambda_{k}^{2}}   \left\| f^{\star}\right\|_{\mathscr{H}_K}, \ j=0,\cdots, h-2.
\end{align}
This together with Condition  \ref{con1}, Proposition  \ref{mingti7}  and Lemma \ref{THMM2} shows that
$$
\lim_{k\to\infty}\left\|f_{\lambda,k}-f^{\star}\right\|_{ \HH_K }=0 \ \text{a.s.}$$
This together with (\ref{mm13}) shows that
$\left\|f_{\lambda,k+1}-f_{\lambda,k}\right\|_{\HH_K}=o(a_k\lambda_k)$.  Note that $$
\mathbb{E}\left[\sup _{k \in \mathbb{N}} \frac{\left\|f_{\lambda,k+1}-f_{\lambda,k}\right\|_{\mathscr{H}_K }^2}{\lambda_k^2}\right] \leq \left(2 \alpha_1 C_2 \left\|  f^{\star}\right\|_{\mathscr{H}_K}\right)^{2}.$$
Then, by Theorem \ref{wuxian}, we have $
\lim_{k\to\infty}\left\|f_k-f^{\star}\right\|_{ \HH_K }=0 \ \text{a.s.}$  Similarly,   we have $\lim_{k\to\infty} f_k(x)=f^{\star}(x),\ \text{ a.s. } \ \forall \ x\in \X$.
\hfill $\blacksquare$

\vskip 2mm
\noindent
\textbf{Proof of Corollary \ref{zasihenghexianz}:}
By Theorem 4.2.6 in \cite{Berlinet}, we know that  $\hat{\mathscr{H}}_K$
is an RKHS whose reproducing kernel is given by
$
\hat{K} = K|_{\hat{\mathscr{X}} \times \hat{\mathscr{X}}},
$
where $K|_{\hat{\mathscr{X}} \times \hat{\mathscr{X}}}$ is the restriction of $K$ to the subset $\hat{\mathscr{X}} \times \hat{\mathscr{X}}$.
This, together with the compactness of $\hat{\mathscr{X}} $, conditions (\ref{cedu111111})-(\ref{over111}) and Corollary \ref{corollary1}, shows the conclusions. \hfill $\blacksquare$

\section{Theoretical framework of random elements with values in a Banach space and some lemmas and propositions}

\setcounter{equation}{0}
\def\theequation{D.\arabic{equation}}
\setcounter{definition}{0}
\def\thedefinition{D.\arabic{definition}}
\setcounter{remark}{0}
\def\theremark{D.\arabic{remark}}
\setcounter{proposition}{0}
\def\theproposition{D.\arabic{proposition}}
\setcounter{lemma}{0}
\def\thelemma{D.\arabic{lemma}}

Let $(\mathscr V,\|\cdot\|_{\mathscr V})$ be a Banach space.
Let $(S,\mathscr A_1)$ and $(T,\mathscr A_2)$ be measurable spaces. If the map $f:S\to T$ satisfies $f^{-1}(B):=\{x\in S:f(x)\in B\}\in \mathscr A_1,~\forall ~B\in\mathscr A_2$, then $f$ is called $\mathscr A_1/\mathscr A_2$-measurable.
Let $L^p(\Omega;\mathscr V)=\{f\in L^0(\Omega;\mathscr V): \|f\|_{L^p(\Omega;\mathscr V)}<\infty \}$
and $L^p(\Omega) := L^p(\Omega;\mathbb R)$.

\begin{definition}
\rm{Let $(\Omega,\F,\mathbb P)$ be a complete probability space. A mapping $f:\Omega\to\mathscr V$ is   said  to be strongly $\P$-measurable or to be a random element with values in the Banach space $\mathscr V$ if it is $\F/\mathscr B(\mathscr V)$-measurable and almost separable valued with respect to the norm $\|\cdot\|_{\mathscr V}$. }
\end{definition}
\vskip 0.2cm
\begin{remark}
\rm{  Especially,  if $\mathscr V$ is a separable Banach space, then any $\F/\mathscr B(\mathscr V)$-measurable mapping $f:\Omega\to\mathscr V$ is  a random element with values in the Banach space $\mathscr V$ (\cite{LZ}).}
\end{remark}
\vskip 0.2cm
\begin{definition}
\rm{If $f\in L^1(\Omega;\mathscr V)$, then the mathematical expectation of $f$ is defined as the Bochner integral
\[\E[f]=\int_{\Omega}f\,\mathrm{d}\mathbb P.\]}
\end{definition}
\vskip 0.2cm
For any given Bochner integrable random element $f$ with values in a Banach space $\mathscr V$, its conditional expectation $\E[f|\mathscr G]\in L^0(\Omega,\mathscr G;\mathscr V)$ with respect to any sub-$\sigma$-algebra $\mathscr G$ of $\F$ uniquely exists, and $\E[f|\mathscr G]$ is also a random element with values in the Banach space $(\mathscr V,\|\cdot\|_{\mathscr V})$ (\cite{LZ}). We have the following propositions about the conditional expectations of operator-valued random elements.
\vskip 0.2cm
\begin{proposition}[\cite{LZ}]\label{mingti1}
\rm{If $f\in L^1(\Omega;\mathscr L(\mathscr Y,\mathscr Z))$ is a random element with values in Banach space $\mathscr L(\mathscr Y,\mathscr Z)$, then $fy\in L^1(\Omega;\mathscr Z)$ is the random element with values in Banach space $\mathscr Z$, and  $\E[fy]=\E[f]y$, $\forall~ y\in \mathscr Y$.}
\end{proposition}
\vskip 0.2cm
\begin{proposition}[\cite{LZ}]\label{mingti2}
\rm{If $f\in L^2(\Omega;\mathscr L(\mathscr Y,\mathscr Z))$ is a random element with values in Banach space $\mathscr L(\mathscr Y,\mathscr Z)$ and $y\in L^2(\Omega,\mathscr G;\mathscr Y)$ is a random element with values in the Banach space $\mathscr Y$, where $\mathscr G$ is a sub-$\sigma$-algebra of $\F$, then $fy\in L^1(\Omega;\mathscr Z)$ is a random element with values in the Banach space $\mathscr Z$ and $\E[fy|\mathscr G]=\E[f|\mathscr G]y$ a.s.}
\end{proposition}


%
%

At first, we have the following propositions.

\begin{proposition}\label{xingzhi1}
\rm{If Assumption \ref{ass2} holds, then $T_{k}:\HH_K\to \HH_K$, $\forall ~k\in\mathbb N$, is a self-adjoint and compact operator a.s. }
\end{proposition}

\begin{proof}
 Let $\{f_n,n\in\mathbb N\}$ be a bounded sequence in $\HH_K$, i.e. there exists a constant $C>0$, such that $\sup_{n\in\mathbb N}\|f_n\|_{\HH_K}\leq C$. On one hand, for any given $k\in\mathbb N$,  it follows from Assumption \ref{ass2}, Proposition \ref{mingti2}, the reproducing property of RKHS and Cauchy inequality that
\ban
&&~~~\left\|T_{k}f_n\right\|_{\HH_K}\cr
&&=\left\|\E\left[ H_{k}  f_n|\F_{k-1}\right]\right\|_{\HH_K}\cr
&&=\left\|\E\left[f_n(x_k)K_{x_k}|\F_{k-1}\right]\right\|_{\HH_K}\cr
&&=\left\|\E\left[\left\langle f_n,K_{x_k}\right\rangle  K_{x_k}|\F_{k-1}\right]\right\|_{\HH_K}\cr
&&\leq \E\left[\left\|f_n\right\|_{\HH_K}\left\|K_{x_k}\right\|_{\HH_K} \left\|K_{x_k}\right\|_{\HH_K}|\F_{k-1}\right]\cr
&&\leq C\E\left[K(x_k,x_k)|\F_{k-1}\right]\cr
&&\leq C\sup_{x\in\X}K(x,x)<\infty~\mathrm{a.s.},
\ean
thus the sequence $\{T_{k}f_n,n\in\mathbb N\}$ is uniformly bounded a.s. On the other hand, noting that $K(\cdot,\cdot)$ is an uniformly continuous function   on $\X\times\X$, then for any given $\varepsilon>0$, there exists  $\delta(\varepsilon)>0$, such that $|K(x_k,y_1)-K(x_k,y_2)|<\varepsilon$, $\forall~\|y_1-y_2\|<\delta,~y_1,\ y_2\in\X$. By the reproducing property of RKHS and Cauchy inequality, we have
\ban
&&~~~\left|\left(T_{k}f_n\right)(y_1)-\left(T_{k}f_n\right)(y_2)\right|\cr
&&=\left|\E\left[f_n(x_k)(K(x_k,y_1)-K(x_k,y_2))\right|\F_{k-1}]\right|\cr
&&=\left|\E\left[\left\langle f_n,K_{x_k} \right\rangle  (K(x_k,y_1)-K(x_k,y_2))|\F_{k-1}\right]\right|\cr
&&\leq C\sup_{x\in\X}\sqrt{K(x,x)}\E\left[\left|K(x_k,y_1)-K(x_k,y_2)\right||\F_{k-1}\right]\cr
&&\leq C\sup_{x\in\X}\sqrt{K(x,x)}\varepsilon.
\ean
Hence, $\{T_{k}f_n,n\in\mathbb N\}$ is equicontinuous a.s. It follows from Arzela-Ascoli theorem that $\{T_{k}f_n,n\in\mathbb N\}$ has a uniformly convergent subsequence a.s.  Then by the definition of the compact operator in \cite{Lax}, we know that $T_{k}$ is compact a.s. By Assumption \ref{ass2},  Proposition 2.6.31 in \cite{Hytonen}, Proposition \ref{mingti2} and the reproducing property of RKHS, we obtain
\ban
&&~~~\left\langle T_{k} f,g\right\rangle_{\HH_K} \cr
&&=\left\langle \E\left[ H_{k} f|\F_{k-1}\right], g\right\rangle_{\HH_K} \cr
&&=\left\langle \E\left[f(x_k)K_{x_k}|\F_{k-1}\right], g\right\rangle_{\HH_K} \cr
&&=\E\left[\left\langle f(x_k)K_{x_k}, g\right\rangle_{\HH_K} |\F_{k-1}\right]\cr
&&=\E\left[f(x_k)g(x_k)|\F_{k-1}\right]\cr
&&=\E\left[g(x_k)\left\langle K_{x_k},f \right\rangle_{\HH_K} |\F_{k-1}\right]\cr
&&=\E\left[\left\langle  H_{k}g,f \right\rangle_{\HH_K} |\F_{k-1}\right]\cr
&&=\left\langle  f,T_{k}g\right\rangle_{\HH_K} ~\mathrm{a.s.},~\forall~ f,g\in \HH_K,
\ean
thus $T_{k}$ is self-adjoint and compact a.s.
\end{proof}

\begin{proposition}\label{xingzhi2}
\rm{Suppose $\lambda>0$. If Assumption \ref{ass2} holds, then $T_{k}+\lambda I$, $\forall ~k\in\mathbb N$, is invertible a.s.}
\end{proposition}

\begin{proof}
For any given $k\in\mathbb N$, it follows from Proposition \ref{xingzhi1} that $T_{k}$ is compact a.s., the eigensystem of which is denoted by $\{(\Lambda_k(i),e_k(i)), i=1,2,\cdots\}$. Noting that $T_{k}\succeq 0$ a.s., which shows that the eigenvalues of $T_{k}+\lambda I$ satisfy $\Lambda_k(i)+\lambda>0$ a.s., $i=1,2,\cdots$, from which we know that $T_{k}+\lambda I$ is injective a.s. For any $y\in \HH_K$, let
$$u_k=\sum_{i=0}^{\infty}\frac{1}{\Lambda_k(i)+\lambda}\langle y, e_k(i)\rangle_{\HH_K}   e_k(i).$$
Noting that $$\|u_k\|^2_{\HH_K}=\sum_{i=0}^{\infty}\left|\frac{1}{\Lambda_k(i)+\lambda}\langle y, e_k(i)\rangle_{\HH_K} \right|^2\leq \frac{1}{\lambda^2}\sum_{i=0}^{\infty}\left|\langle y,e_k(i)\rangle_{\HH_K} \right|^2=\frac{1}{\lambda^2}\|y\|^2_{\HH_K}<\infty~\mathrm{a.s.},$$
then we have $u_k\in\HH_K$ a.s. Noting that $$\langle u_k,e_k(i)\rangle_{\HH_K} =\frac{1}{\Lambda_k(i)+\lambda}\langle y,e_k(i)\rangle_{\HH_K}  ~\mathrm{a.s.},$$
we obtain
\begin{align*}
   \left(T_{k}+\lambda I\right)u_k
 =&\sum_{i=0}^{\infty}(\Lambda_k(i)+\lambda)\langle u_k, e_k(i)\rangle_{\HH_K}  e_k(i)=\sum_{i=0}^{\infty}\langle y, e_k(i)\rangle_{\HH_K}  e_k(i)=y~\mathrm{a.s.},
\end{align*}
which shows that $T_{k}+\lambda I$ is surjective a.s., and therefore invertible a.s.
\end{proof}

\begin{lemma}\label{lemma1111}
\rm{If Assumption \ref{ass2} and Condition \ref{con1} hold, then there exists an integer $k_0\in\mathbb N$, such that
\ban
\left\|\Phi_{i,j}\right\|_{\mathscr L(\HH_K)}\leq \prod_{k=j}^i\left(1-a_k\lambda_k\right)~\mathrm{a.s.},~\forall~ i,j\ge k_0.
\ean
}
\end{lemma}

\begin{proof}
It follows from Assumption \ref{ass2}, the reproducing property of RKHS and Cauchy inequality that
\ban
\sup_{\|f\|_{\HH_K}=1, f\in\HH_K}a_k\left\langle  H_{k}f,f\right\rangle_{\HH_K}   &=&a_k\sup_{\|f\|_{\HH_K}=1, f\in\HH_K}\left\langle f(x_k)K_{x_k},f \right\rangle_{\HH_K}  \cr
&\leq& a_k\sup_{\|f\|_{\HH_K}=1, f\in\HH_K}\left|f(x_k)\right|\left\|K_{x_k}\right\|_{\HH_K}\cr
&=& a_k\sup_{\|f\|_{\HH_K}=1, f\in\HH_K}\left|\left\langle f,K_{x_k} \right\rangle_{\HH_K} \right|\left\|K_{x_k}\right\|_{\HH_K}\cr
&\leq& a_kK(x_k,x_k)
 \leq  a_k\kappa~\mathrm{a.s.},~\forall~k\in\mathbb N.
\ean
It follows from Condition \ref{con1} that $\lim\limits_{k\to\infty}(a_k\lambda_k+a_k\kappa)=0$. Then there exists an integer $k_0\in\mathbb N$, such that
\begin{align}\label{a}
& 1-a_k\lambda_k-\sup_{\|f\|_{\HH_K}=1, f\in\HH_K}a_k\left\langle  H_{k} f,f\right\rangle_{\HH_K}  \notag\\
\ge & 1-a_k\lambda_k-a_k\kappa>0~\mathrm{a.s.},~\forall~ k\ge k_0.
\end{align}
By the reproducing property of RKHS, we get
$$\left\langle  H_{k}f,f\right\rangle_{\HH_K}  =\left\langle f(x_k)K_{x_k},f \right\rangle_{\HH_K}  =f(x_k)\left\langle K_{x_k},f \right\rangle_{\HH_K}  =f^2(x_k)\ge 0~\mathrm{a.s.},~\forall~ k\in\mathbb N.$$
Thus, it follows from (\ref{a}) that
\ban
&&~~~\left\|I-a_k\left(H_{k}+\lambda_kI\right)\right\|_{\mathscr L(\HH_K)}\cr
&&=\sup_{\|f\|_{\HH_K}=1, f\in\HH_K}\left|\left\langle \left(I-a_k\left(H_{k}+\lambda_kI\right)\right)f,f \right\rangle_{\HH_K}  \right|\cr
&&=\sup_{\|f\|_{\HH_K}=1, f\in\HH_K}\left|1-a_k\lambda_k-a_k\left\langle H_{k}f,f\right\rangle_{\HH_K}  \right|\cr
&&=\sup_{\|f\|_{\HH_K}=1, f\in\HH_K}\left(1-a_k\lambda_k-a_k\left\langle H_{k}f,f\right\rangle_{\HH_K}  \right)\cr
&&=1-a_k\lambda_k-a_k\inf_{\|f\|_{\HH_K}=1, f\in\HH_K}\left\langle  H_{k} f,f\right\rangle_{\HH_K}
 \leq 1-a_k\lambda_k~\mathrm{a.s.},~\forall~ k\ge k_0,
\ean
from which we obtain
\ban
\left\|\Phi_{i,j}\right\|_{\mathscr L(\HH_K)}\leq \prod_{k=j}^i\left\|I-a_k\left(H_{k}+\lambda_kI\right)\right\|_{\mathscr L(\HH_K)} \leq \prod_{k=j}^i(1-a_k\lambda_k)~\mathrm{a.s.},~\forall ~i,j\ge k_0.
\ean
\end{proof}

\begin{lemma}\label{lemma10086}
\rm{If $\X$ is a compact set in $\mathbb R^n$ and $0\leq s\leq 1$, then
\ban
|fg|_{C^s(\X)}\leq |f|_{C^s(\X)}\|g\|_{\infty}+\|f\|_{\infty}|g|_{C^s(\X)},~\forall~ f,g \in C^s(\X).
\ean
}
\end{lemma}

\vskip 0.2cm

\begin{proof}
It follows from the definitions of $|\cdot|_{C^s(\X)}$ and $\|\cdot\|_{C^s(\X)}$ that
\ban
&&~~~|fg|_{C^s(\X)}\cr
&&=\sup_{x\neq y\in \X}\frac{|f(x)g(x)-f(y)g(y)|}{\|x-y\|^s}\cr
&&=\sup_{x\neq y\in \X}\frac{|f(x)g(x)-f(y)g(x)+f(y)g(x)-f(y)g(y)|}{\|x-y\|^s}\cr
&&\leq\sup_{x\neq y\in \X}\frac{|f(x)g(x)-f(y)g(x)|+|f(y)g(x)-f(y)g(y)|}{\|x-y\|^s}\cr
&&\leq\sup_{x\neq y\in \X}\frac{|f(x)g(x)-f(y)g(x)|}{\|x-y\|^s}+\sup_{x\neq y\in \X}\frac{|f(y)g(x)-f(y)g(y)|}{\|x-y\|^s}\cr
&&\leq\left(\sup_{x\neq y\in \X}\frac{|f(x)-f(y)|}{\|x-y\|^s}\right)\left(\sup_{x\in \X}|g(x)|\right)+\left(\sup_{y\in \X}|f(y)|\right)\left(\sup_{x\neq y\in \X}\frac{|g(x)-g(y)|}{\|x-y\|^s}\right)\cr
&&=|f|_{C^s(\X)}\|g\|_{\infty}+\|f\|_{\infty}|g|_{C^s(\X)},~\forall ~f,g \in C^s(\X).
\ean
\end{proof}

\vskip 0.2cm

\begin{lemma}\label{lemma6}
\rm{If Assumption \ref{ass2} holds,  and $\{\lambda_k,k\in\mathbb N\}$ is a sequence of positive real numbers, then
\bna\label{da400}
\left\|f_{\lambda,k+1}-f_{\lambda,k}\right\|_{\HH_K}&\leq& \left(\frac{\left\|T_{k+1}-T_{k}\right\|_{\mathscr L(\HH_K)}}{\lambda_k}\right.\cr
&~~&\left.+\frac{\lambda_k-\lambda_{k+1}}{\lambda_k}\right)\|f_{\lambda,k}-f^{\star}\|_{\HH_K}~\mathrm{a.s.},~\forall~k\in\mathbb N.
\ena
}
\end{lemma}

\vskip 0.2cm
\begin{proof}
It follows from Assumption \ref{ass2} and the definition of $f_{\lambda,k}$ that
\ban
\lambda_kf_{\lambda,k}=T_{k}f^{\star}-T_{k}f_{\lambda,k},~\forall~ k\in\mathbb N,
\ean
from which we have
\ban
&&~~~\left(T_{k+1}+\lambda_{k+1}I\right)\left(f_{\lambda,k+1}-f_{\lambda,k}\right)\cr
&&=T_{k+1}f^{\star}-T_{k+1}f_{\lambda,k}-\lambda_{k+1}f_{\lambda,k}\cr
&&=T_{k+1}f^{\star}-T_{k+1}f_{\lambda,k}-\frac{\lambda_{k+1}\lambda_k}{\lambda_k}f_{\lambda,k}\cr
&&=T_{k+1}f^{\star}-T_{k+1}f_{\lambda,k} -\frac{\lambda_{k+1}}{\lambda_k}\left(T_{k}f^{\star}-T_{k}f_{\lambda,k}\right)\cr
&&=\left(T_{k+1}-\frac{\lambda_{k+1}}{\lambda_k}T_{k}\right)(f^{\star}-f_{\lambda,k})\cr
&&=\frac{1}{\lambda_k}\left(\lambda_kT_{k+1}-\lambda_{k+1}T_{k}\right)(f^{\star}-f_{\lambda,k})\cr
&&=\frac{\lambda_k-\lambda_{k+1}}{\lambda_k}T_{k+1}(f^{\star}-f_{\lambda,k}) +\frac{\lambda_{k+1}}{\lambda_k}\left(T_{k+1}-T_{k}\right)(f^{\star}-f_{\lambda,k})~\mathrm{a.s.},~\forall~k\in\mathbb N.
\ean
By multiplying $(T_{k+1}+\lambda_{k+1}I)^{-1}$ on both sides of the above equality, we get
\bna\label{ppp}
&&f_{\lambda,k+1}-f_{\lambda,k}\cr
&&\hspace{-0.3cm}=\frac{\lambda_k-\lambda_{k+1}}{\lambda_k}\left(T_{k+1}+\lambda_{k+1}I\right)^{-1}T_{k+1}(f^{\star}-f_{\lambda,k})\cr
&&+\frac{\lambda_{k+1}}{\lambda_k}\left(T_{k+1}+\lambda_{k+1}I\right)^{-1} \left(T_{k+1}-T_{k}\right)(f^{\star}-f_{\lambda,k})~\mathrm{a.s.},~\forall~ k\in\mathbb N.
\ena
Noting that
\bna\label{ggg}
\begin{cases}
\displaystyle\left\|\left(T_{k}+\lambda_kI\right)^{-1}T_{k}\right\|_{\mathscr L(\HH_K)}\leq 1~\mathrm{a.s.},\\
\displaystyle\left\|\left(T_{k}+\lambda_kI\right)^{-1}\right\|_{\mathscr L(\HH_K)}\leq \frac{1}{\lambda_k}~\mathrm{a.s.},~\forall~ k\in\mathbb N,
\end{cases}
\ena
then by (\ref{ppp})-(\ref{ggg}), we obtain (\ref{da400}).
\end{proof}

\begin{lemma}\label{lemma111}
\rm{If the sequences $\{a_k,k\in\mathbb N\}$ and $\{\lambda_k,k\in \mathbb N\}$ satisfy \[a_k=\frac{\alpha_{1}}{(k+1)^{\tau_1}},\quad \lambda_k=\frac{\alpha_{2}}{(k+1)^{\tau_2}},~\forall~ k\in \mathbb N,\]
where $\alpha_{1},\ \alpha_{2}$, $\tau_1,\ \tau_{2}>0$, $\tau_1+\tau_2<1,~3\tau_2<\tau_1$, then there exists $k_0=
7+ \lceil
 \big(2(\alpha_1\alpha_2+\alpha_1\kappa) \big)^{\frac{1}{\tau_1}}\rceil+
\lceil\frac{\tau_1-3\tau_2}{\alpha_1\alpha_2}\big)^{\frac{2}{1-\tau_1-\tau_2}}\rceil+
\lceil\big(\frac{32}{\alpha_1\alpha_2(1-\tau_1-\tau_2)}\big)^{\frac{2}{1-\tau_1-\tau_2}}\rceil,$  such that

(i)
 $$ \sum_{i=k_0}^ka^2_i\prod_{j=i+1}^k(1-a_j\lambda_j)
 \sqrt{k-i+1}\leq C_2 (k+1)^{\frac{3\tau_2-\tau_1}{2}}\ln^{\frac{3}{2}}(k+1),  \forall \ k\geq k_0,$$
where $C_2=  \alpha_1^2
+
\alpha_1^2 4^{\tau_1}
\left(
\frac{2}{\alpha_1\alpha_2}+2
\right)^{\frac{3}{2}};$

(ii)
\begin{align}
 \prod_{i=k_0}^k\left(1-a_i\lambda_i\right)
\leq  C_3 (k+1)^{\frac{3\tau_2-\tau_1}{2}}\ln^{\frac{3}{2}}(k+1),\ k\geq k_0.\notag
\end{align}
where $C_3=\exp\left(
\frac{\alpha_1\alpha_2}{1-\tau_1-\tau_2}
(k_0+1)^{1-\tau_1-\tau_2}
\right).$
}
\end{lemma}

\vskip 0.2cm

\begin{proof}
 By the definition of
$k_0$, for all $k\ge k_0$, we have
$(k+1)^{\tau_1+\tau_2}\ge 2\alpha_1\alpha_2$, and hence
$0<1-a_k\lambda_k<1$.
Using $1-x\le e^{-x}$, we obtain, for $k_0\le i\le k$,
\[
\prod_{j=i+1}^k(1-a_j\lambda_j)
\le
\exp\left(
-\frac{\alpha_1\alpha_2}{1-\tau_1-\tau_2}
\left(
(k+1)^{1-\tau_1-\tau_2}
-(i+2)^{1-\tau_1-\tau_2}
\right)
\right),
\]
where we used
\[
\sum_{j=i+1}^k\frac1{(j+1)^{\tau_1+\tau_2}}
\ge
\int_{i+1}^k\frac{dx}{(x+1)^{\tau_1+\tau_2}}.
\]
Define
\[
\epsilon_k=
\left\lceil
\frac{2}{\alpha_1\alpha_2}
(k+1)^{\tau_1+\tau_2}\ln(k+1)
\right\rceil .
\]
Noting that $\ln x\le \frac{2}{1-\tau_1-\tau_2}
x^{\frac{1-\tau_1-\tau_2}{2}}$ for $x\ge1$, and by the choice of
$k_0$, we have
\[
k+1\ge
\left(
\frac{32}{\alpha_1\alpha_2(1-\tau_1-\tau_2)}
\right)^{\frac{2}{1-\tau_1-\tau_2}},
\]
and
\[
\ln(k+1)
\le
\frac{\alpha_1\alpha_2}{16}
(k+1)^{1-\tau_1-\tau_2}.
\]
Hence
\[
2\epsilon_k
\le
\frac{4}{\alpha_1\alpha_2}
(k+1)^{\tau_1+\tau_2}\ln(k+1)+2
\le
\frac{k+1}{4}+2<k,
\]
for all $k\ge k_0$, as $k_0\ge7$.
We consider two cases.

\textbf{Case 1.} If $k-1-\epsilon_k<k_0$, then $k-k_0\le\epsilon_k$.
Moreover, for $i\in\{k_0,\ldots,k\}$, we have
$i+1>k-\epsilon_k>\frac{k}{2}$, and hence
\[
\frac1{(i+1)^{2\tau_1}}
\le
\frac{4^{\tau_1}}{(k+1)^{2\tau_1}}.
\]
Using $\prod_{j=i+1}^k(1-a_j\lambda_j)\le1$ and
$\epsilon_k+1\le
\left(\frac{2}{\alpha_1\alpha_2}+2\right)
(k+1)^{\tau_1+\tau_2}\ln(k+1)$, we get
\begin{align}
\sum_{i=k_0}^k
a_i^2
\prod_{j=i+1}^k(1-a_j\lambda_j)
\sqrt{k-i+1}
&\leq
\alpha_1^2
\frac{4^{\tau_1}}{(k+1)^{2\tau_1}}
(\epsilon_k+1)\sqrt{\epsilon_k+1}  \notag\\
&\le
\alpha_1^2 4^{\tau_1}
\left(
\frac{2}{\alpha_1\alpha_2}+2
\right)^{\frac{3}{2}}
(k+1)^{\frac{3\tau_2-\tau_1}{2}}
\ln^{\frac{3}{2}}(k+1).\label{case1}
\end{align}

\textbf{Case 2.} If $k-1-\epsilon_k\ge k_0$, then
\begin{align}
&\sum_{i=k_0}^k
a_i^2
\prod_{j=i+1}^k(1-a_j\lambda_j)
\sqrt{k-i+1}\notag\\
=&
\sum_{i=k_0}^{k-1-\epsilon_k}
a_i^2\prod_{j=i+1}^k(1-a_j\lambda_j)\sqrt{k-i+1}
+
\sum_{i=k-\epsilon_k}^{k}
a_i^2\prod_{j=i+1}^k(1-a_j\lambda_j)\sqrt{k-i+1}.\notag
\end{align}
For $k_0\le i\le k-1-\epsilon_k$, we have
$i+2\le k+1-\epsilon_k$. By
$(1-x)^{1-\tau_1-\tau_2}\le
1-(1-\tau_1-\tau_2)x$ for $0\le x\le1$, we have
\[
\frac{\alpha_1\alpha_2}{1-\tau_1-\tau_2}
\left(
(k+1)^{1-\tau_1-\tau_2}
-(i+2)^{1-\tau_1-\tau_2}
\right)
\ge 2\ln(k+1).
\]
Consequently,
\[
\prod_{j=i+1}^k(1-a_j\lambda_j)\le \frac1{(k+1)^2},
\qquad k_0\le i\le k-1-\epsilon_k.
\]
Thus
\begin{align}\label{case21}
  \sum_{i=k_0}^{k-1-\epsilon_k}
a_i^2\prod_{j=i+1}^k(1-a_j\lambda_j)\sqrt{k-i+1}
\le
\alpha_1^2(k+1)^{\frac{1}{2}}.
\end{align}
By $0<\tau_1-3\tau_2<1$ and $\ln(k+1)\ge1$, we have
\[
(k+1)^{-\frac{1}{2}}
\le
(k+1)^{\frac{3\tau_2-\tau_1}{2}}
\ln^{\frac{3}{2}}(k+1).
\]
For $k-\epsilon_k\le i\le k$, using $2\epsilon_k<k$, we have
$i+1>\frac{k}{2}$, and hence
\[
\frac1{(i+1)^{2\tau_1}}
\le
\frac{4^{\tau_1}}{(k+1)^{2\tau_1}}.
\]
Therefore,

\begin{align}
&\sum_{i=k-\epsilon_k}^{k}
a_i^2\prod_{j=i+1}^k(1-a_j\lambda_j)\sqrt{k-i+1} \notag\\
&\le
\alpha_1^2 4^{\tau_1}
\left(
\frac{2}{\alpha_1\alpha_2}+2
\right)^{\frac{3}{2}}
(k+1)^{\frac{3\tau_2-\tau_1}{2}}
\ln^{\frac{3}{2}}(k+1).\label{case2}
\end{align}
Combining  (\ref{case21})  and (\ref{case2}) yields
\[
\sum_{i=k-\epsilon_k}^{k}
a_i^2\prod_{j=i+1}^k(1-a_j\lambda_j)\sqrt{k-i+1}
\le
C_2
(k+1)^{\frac{3\tau_2-\tau_1}{2}}
\ln^{\frac{3}{2}}(k+1).
\]
Combining Case 1 and Case 2 gives assertion (i).

It remains to prove assertion (ii). By the same product estimate, we have
\[
\begin{aligned}
\prod_{i=k_0}^k(1-a_i\lambda_i)
&\le
\exp\left(
-\frac{\alpha_1\alpha_2}{1-\tau_1-\tau_2}
\left(
(k+1)^{1-\tau_1-\tau_2}
-(k_0+1)^{1-\tau_1-\tau_2}
\right)
\right)  \\
&\leq
C_3
\exp\left(
-\frac{\alpha_1\alpha_2}{1-\tau_1-\tau_2}
(k+1)^{1-\tau_1-\tau_2}
\right).
\end{aligned}
\]
By the definition of $k_0$, we have
\[
(k+1)^{\frac{1-\tau_1-\tau_2}{2}}
\ge
\frac{\tau_1-3\tau_2}{\alpha_1\alpha_2}.
\]
This together with
$\ln x\le \frac{2}{1-\tau_1-\tau_2}
x^{\frac{1-\tau_1-\tau_2}{2}}$   implies
\[
\frac{\alpha_1\alpha_2}{1-\tau_1-\tau_2}
(k+1)^{1-\tau_1-\tau_2}
\ge
\frac{\tau_1-3\tau_2}{2}\ln(k+1).
\]
Hence
\[
\prod_{i=k_0}^k(1-a_i\lambda_i)
\le
C_3
(k+1)^{\frac{3\tau_2-\tau_1}{2}}.
\]
Noting that $\ln(k+1)\ge1$ for $k\ge7$, we further have
\[
\prod_{i=k_0}^k(1-a_i\lambda_i)
\le
C_3
(k+1)^{\frac{3\tau_2-\tau_1}{2}}
\ln^{\frac{3}{2}}(k+1).
\]
\end{proof}

\vskip 0.2cm
%
%
%
%


\vskip 0.2cm

\end{appendices}

\end{document}